\documentclass[journal]{IEEEtran}
\usepackage{amsmath,amsfonts}
\usepackage{algorithmic}
\usepackage{array}
\usepackage[caption=false,font=normalsize,labelfont=sf,textfont=sf]{subfig}
\usepackage{textcomp}
\usepackage{stfloats}
\usepackage{url}
\usepackage{verbatim}
\usepackage{graphicx}
\usepackage{hyperref}
\usepackage{cite}
\usepackage{mwe}
\usepackage{balance}
\usepackage[ruled,vlined]{algorithm2e}
\usepackage{color}
\usepackage{xcolor}
\usepackage{amssymb}
\usepackage{adjustbox} 
\usepackage{threeparttable} 

\newtheorem{thm}{Theorem}[section]

\newtheorem{rem}{Remark}[section]

\newtheorem{lem}{Lemma}[section]

\newtheorem{assump}{Assumption}[section]

\newenvironment{proof}{\qquad \textit{Proof:}}{\hfill$\square$}

\newcommand{\boxend}{\hfill \ensuremath{\blacksquare}}

\newcommand{\vect}[1]{\boldsymbol{\mathbf{#1}}}
\newcommand{\real}{{\mathbb{R}}}

\NewDocumentCommand{\todo}{o m}{\textcolor{red}{\textbf{TODO\IfNoValueTF{#1}{}{(#1)}:} #2}}
\NewDocumentCommand{\note}{o m}{\textcolor{orange}{\textbf{NOTE\IfNoValueTF{#1}{}{(#1)}:} #2}}

\begin{document}


\title{B-ActiveSEAL: Scalable Uncertainty-Aware Active Exploration with \\ Tightly Coupled Localization-Mapping}





\author{Min-Won Seo, Aamodh Suresh, Carlos Nieto-Granda, and  Solmaz S. Kia,~\IEEEmembership{Senior Member, IEEE}


\thanks{Min-Won Seo, and Solmaz S. Kia are with the Department of Mechanical and Aerospace Engineering,
        University of California, Irvine, CA 92697, USA,
        {\tt\small \{minwons,solmaz\}@uci.edu}}
\thanks{Aamodh Suresh and C. Nieto are with the U.S. DEVCOM Army Research Laboratory (ARL), Adelphi, MD 20783, {\tt\small aamodh@gmail.com}, {\tt\small carlos.p.nieto2.civ@army.mil}}
}

\markboth{Journal of \LaTeX\ Class Files,~Vol.~00, No.~0, Month~Year}%
{Shell \MakeLowercase{\textit{et al.}}: A Sample Article Using IEEEtran.cls for IEEE Journals}


\maketitle

\begin{abstract}
Active robot exploration requires decision-making processes that integrate localization and mapping under \emph{tightly coupled} uncertainty. However, managing these interdependent uncertainties over long-term operations in large-scale environments rapidly becomes computationally intractable. To address this challenge, we propose \textsf{B-ActiveSEAL}, a scalable information-theoretic active exploration framework that explicitly accounts for coupled uncertainties—from perception through mapping—into the decision-making process. Our framework (i) adaptively balances map uncertainty (exploration) and localization uncertainty (exploitation), (ii) accommodates a broad class of generalized entropy measures, enabling flexible and uncertainty-aware active exploration, and (iii) establishes Behavioral entropy (BE) as an effective information measure for active exploration by enabling intuitive and adaptive decision-making under coupled uncertainties. We establish a theoretical foundation for propagating coupled uncertainties and integrating them into general entropy formulations, enabling uncertainty-aware active exploration under tightly coupled localization–mapping.
The effectiveness of the proposed approach is validated through rigorous theoretical analysis and extensive experiments on open-source maps and ROS–Unity simulations across diverse and complex environments. The results demonstrate that \textsf{B-ActiveSEAL} achieves a well-balanced exploration–exploitation trade-off and produces diverse, adaptive exploration behaviors across environments, highlighting clear advantages over representative baselines.
\end{abstract}

\begin{IEEEkeywords}
Active exploration, information theory, uncertainty quantification, Behavioral entropy, localization, dense occupancy mapping, and decision making under uncertainty.
\end{IEEEkeywords}

\section{Introduction} 
\IEEEPARstart{A}{ctive} robot exploration addresses the problem of autonomously selecting control actions to efficiently map unknown environments while simultaneously localizing the robot within it. This approach of autonomously planning control actions distinguishes itself from conventional Simultaneous Localization and Mapping (SLAM), where navigation is typically pre-determined or human-driven (i.e., independent of the mapping process).
Active exploration is a cornerstone capability for critical applications such as search and rescue~\cite{DangAEC2020}, inspection~\cite{walter2008slam, andersen2025TFR}, and environmental monitoring~\cite{dunbabin2012robots}. These applications demand robust navigation and mapping in unknown, GPS-denied environments using only noisy inertial and exteroceptive sensing~\cite{schmid2022scexplorer, kulkarni2022ICRA}. Achieving this requires an \emph{uncertainty-aware} decision-making system capable of managing the \emph{tightly coupled} uncertainties that propagate from perception through mapping to control. Efficiently quantifying and managing these coupled uncertainties remains a central challenge, particularly in large-scale environments. Moreover, real-world environments differ widely in geometry, scale, and sensor visibility, motivating the need for a flexible exploration mechanism whose exploration behavior can adapt to diverse structural and sensing conditions—an ability that generalized entropy measures naturally provide. To overcome these limitations, we develop a novel uncertainty-aware active exploration framework that manages tightly coupled localization–mapping uncertainties throughout the decision-making process using generalized entropy measures, while maintaining computational efficiency and scalability.

Active exploration is often framed within the mathematical framework of partially observable Markov decision processes (POMDPs), which formally model decision-making under uncertainty~\cite{kaelbling1998planning}. Because solving the full POMDP is computationally intractable for this domain, active exploration is commonly decomposed into three subproblems~\cite{fox1998active,makarenko2002experiment}: identifying potential goal points, computing the utility of reaching those points, and selecting actions that maximize utility. The utility function must incorporate both localization (the robot’s state) and map uncertainty, as the robot relies on its position to build a map, and the map in turn supports self-localization. A widely adopted approach is to use \emph{entropy}-based utility functions, which quantify the coupled uncertainty in the joint distribution of the robot’s state $\vect{x}$ and the map $\vect{m}$ in an \emph{information-theoretic} manner. Building on this idea, \cite{bourgault2002information} pioneered the use of entropy-based utility functions for uncertainty-aware exploration, where the utility—commonly referred to as \emph{information gain}—is typically evaluated as
\begin{align}\label{eq::InfoGain}
    &\mathbb{I}_G\bigr[\vect{u}_{t+1:t'}|\mathbf{z}_{t+1:t'}\bigr]  \\ 
    & \triangleq \underbrace{\mathbb{H}\bigl(p(\vect{m}_t, \vect{x}_{0:t}|\vect{l}_{1:t})\bigl)}_{\text{current joint entropy}} \!-\underbrace{\mathbb{H}\bigl(p(\vect{m}_{t'}, \vect{x}_{0:t'}|\vect{l}_{1:t}, \vect{u}_{t+1:t'},\vect{z}_{t+1:t'})\bigl)}_{\text{predicted joint entropy}}\!, \nonumber
\end{align}
where $\vect{x}_{0:t} = \{\vect{x}_0,\dots,\vect{x}_t\}$ denotes the robot’s state trajectory up to time $t$, and $\vect{l}_{1:t} = (\vect{u}_{1:t}, \vect{z}_{1:t})$ represents the history of control actions and exteroceptive sensor measurements, with $\vect{u}_{1:t} = \{\vect{u}_1,\dots,\vect{u}_t\}$ and $\vect{z}_{1:t} = \{\vect{z}_1,\dots,\vect{z}_t\}$. $\vect{z}_{t+1:t'}$ denotes the set of predicted future measurements. The index $t'$ denotes a future time step, with $t' > t$. Maximizing the information gain corresponds to selecting control sequences $\vect{u}_{t+1:t'} \in \mathcal{U}_t$, where $\mathcal{U}_t$ denotes the candidate set, that are predicted to achieve the greatest reduction in the joint entropy, thereby making the estimates of the map and robot pose more certain and accurate. However, realizing this objective in large-scale environments requires addressing a number of fundamental challenges inherent to current information-theoretic exploration approaches, primarily revolving around the computational intractability of coupled uncertainty models, the inflexibility of standard utility functions, and the lack of scalability for long-term operation.

One primary challenge in evaluating the information gain~\eqref{eq::InfoGain} is the computation of the joint entropy of the full system state posterior, defined over the robot's current pose and the evolving map, which is generally analytically intractable. To address this, most works assume either conditional independence~\cite{stachniss2005information} or full independence~\cite{Vallv2014,Suresh2020} between localization and map uncertainty, leading to a \emph{separation} approach. However, this separation introduces a delicate trade-off between exploration (discovering new areas) and exploitation (using mapped areas to improve localization). In practice, this balance often requires \emph{manual tuning} of relative weights, a heuristic process further complicated by \emph{scale mismatches} between the two uncertainty measures~\cite{kim2015active}. Eliminating the need for such heuristic tuning while enabling principled handling of coupled uncertainties remains a central open challenge.

A second challenge lies in selecting the appropriate entropy measure. This selection is critical, as it dictates the trade-off between computational tractability and the flexibility to adjust exploration strategies. Most existing works adopt Shannon entropy (SE)~\cite{shannon1948mathematical} because it admits a \emph{closed-form} expression for Gaussian and binary random variables, making it computationally convenient for localization–mapping frameworks. However, SE offers \emph{limited flexibility} for adjusting exploration strategies. In contrast, generalized entropy (e.g., Tsallis, Rényi) introduces a \emph{tunable} parameter that allows for adaptive exploration strategies. However, their inherent nonlinearity—specifically, being nonlinear functions of the probability distribution (e.g., involving $p^\alpha$)—introduces computational barriers. This nonlinearity makes the expected future entropy required by the utility function~\eqref{eq::InfoGain} analytically intractable, forcing these works to adopt heuristic solutions, such as combining them with SE.

A third challenge lies in maintaining computational efficiency and scalability during \emph{long-term} operation in \emph{large-scale} environments. The computational burden of localization–mapping and multi-step prediction for decision-making grows rapidly. Particle-filter (PF) frameworks, in particular, become computationally prohibitive under such conditions due to the curse of dimensionality. Similarly, Extended Kalman Filter (EKF) frameworks that incorporate the map into the state vector scale poorly as the environment grows. Graph–based frameworks exploit sparsity for efficiency; however, robust performance depends on loop closures, which impose significant computational overhead, and no efficient or principled methods exist for predicting loop–closure hypotheses (i.e., potential virtual nodes)~\cite{Cadena2016}.

To address these fundamental challenges, this paper presents a twofold contribution. First, we develop a \emph{probabilistically principled} framework that efficiently manages the \emph{coupled} uncertainties spanning localization, mapping, and decision-making. Specifically, we introduce a novel formulation where localization and mapping uncertainties are reciprocally embedded. A Gaussian-based filter with \emph{dual} parameter forms integrates map uncertainty into localization, while a weighted marginalized likelihood model for dense occupancy maps incorporates localization uncertainty back into the map. This two-way coupling is founded on a novel likelihood model for beam measurements that captures both dense-map and sensor uncertainties within a unified Gaussian formulation. This framework integrates the resulting coupled uncertainty directly into the decision-making process, allowing map-only entropy to implicitly account for localization uncertainty without requiring explicit joint-entropy computation in~\eqref{eq::InfoGain}. As a result, the proposed framework (i) eliminates reliance on the separation scheme and heuristic methods for computing joint entropy, ensuring a principled balance between exploration and exploitation; (ii) applies to a broad class of generalized entropy measures, enabling adaptable exploration strategies; and (iii) renders the coupled-uncertainty prediction step for decision-making computationally tractable, supporting long-term, large-scale operations. Collectively, these capabilities lead to improved exploration and mapping performance.

\begin{figure}[t]
    \centering
    \begin{minipage}[t]{0.24\textwidth}
        \centering
        \includegraphics[width=\textwidth]{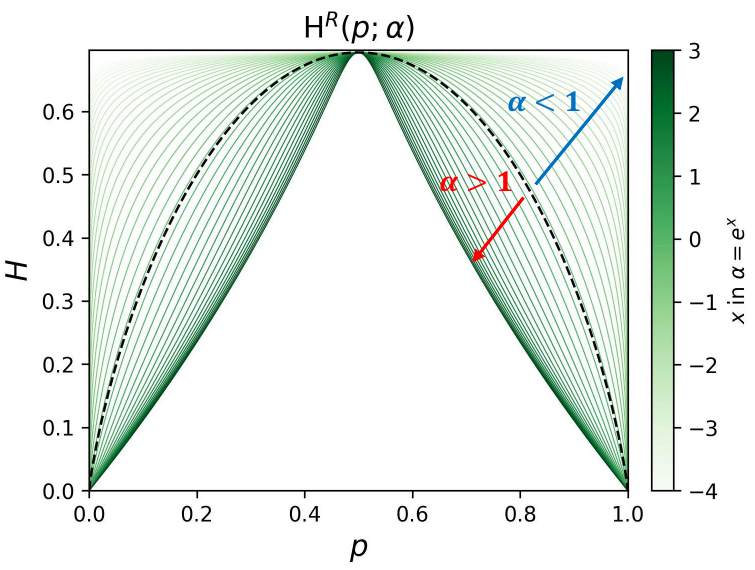}
        {{\scriptsize (a) Rényi Entropy}}
    \end{minipage}
    \vspace{1pt}
    \begin{minipage}[t]{0.22\textwidth}
        \centering
        \includegraphics[width=\textwidth]{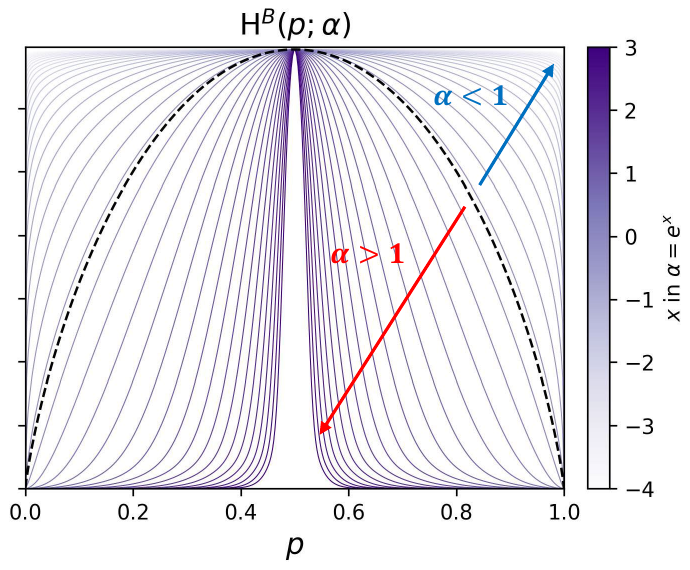}
        {{\scriptsize (b) Behavioral Entropy}}
    \end{minipage}
\caption{{\small Comparison between (a) Rényi entropy, $\mathbb{H}^R_{\alpha}(p)=\frac{1}{1-\alpha} \log\sum_ip_i^{\alpha}$, and (b) Behavioral entropy  (see~\eqref{eq::BE}) as functions of probability $p$ and parameter $\alpha \!> \!0$ (with fixed $\beta\! >\! 0$) for a binomial distribution. Both entropies vary with $\alpha$ and reduce to Shannon entropy (black dashed line) at $\alpha \rightarrow 1$. The plots show that Behavioral entropy offers a more expressive representation of uncertainty than Rényi entropy, covering a broader entropy range as~$\alpha$.}}
    \label{fig::Generalized_Entropy}
\end{figure}

Second, to accomplish mission-aware exploration, we integrate Behavioral entropy (BE)~\cite{suresh2024robotic} into the active exploration framework. BE uses a mathematical model derived from behavioral economics~\cite{dhami2016foundations} to represent human-like decision-making under uncertainty~\cite{tversky1992advances}, which often deviates from traditional rational models. It uses Prelec's weighting function and introduces parameters ($\alpha$ and $\beta$) to model both \emph{uncertainty averse} and \emph{uncertainty ignorant} behaviors by altering how probabilities are weighted and perceived in the final entropy calculation. As shown in Fig.~\ref{fig::Generalized_Entropy}, while Rényi entropy provides some degree of adaptive behavior, BE offers a more comprehensive and intuitive framework for encoding mission goals. BE's expressiveness allows for covering different missions through adaptive online behavior, going beyond just reacting to uncertainty. By adjusting BE's parameters, we can prioritize different exploration strategies: for example, an \emph{exploitation}-driven, uncertainty-averse robot for a hazardous search-and-rescue mission, or an \emph{exploration}-driven, uncertainty-ignorant robot for rapid warehouse mapping.

In summary, the main contributions of this work are: 
\begin{itemize} \item A Probabilistically Principled Active Exploration Framework: A novel and computationally tractable framework for managing coupled localization and mapping uncertainty. In particular, we introduce \textsf{T-BayesMap}, which enables map-only entropy to implicitly account for localization uncertainty without requiring explicit joint-entropy computation. This yields a principled balance between exploration and exploitation, eliminates reliance on heuristic tuning, and supports a broad class of generalized entropy measures for adjustable exploration strategies. \item Integration of Behavioral entropy (BE): We introduce \textsf{B-ActiveSEAL}, which incorporates BE into the proposed framework to enable mission-aware exploration. BE induces intuitive, uncertainty-aware behaviors, such as uncertainty-averse or uncertainty-ignorant decision-making. \end{itemize}

We validated our proposed framework through rigorous theoretical analysis and extensive experiments on open-source maps and ROS–Unity 3D simulations in complex~environments.

\section{Related Work}

Most prior work in active exploration assumes \emph{nearly perfect} localization and therefore focuses decision-making solely on mapping objectives, e.g.,~\cite{Koga2021Iterative,cao2021tare,ho2024mapex,chen2025hierarchical}. In contrast, active exploration under coupled localization–mapping uncertainty—often termed active SLAM—requires jointly reasoning about both components and explicitly managing the exploration–exploitation trade-off~\cite{placed2023survey}. In the remainder of this section, we review work that addresses this coupled setting and refer to it simply as \emph{active exploration}.

Advanced SLAM techniques have been incorporated into active exploration, particularly graph-based approaches in which active loop closure enhances exploration robustness~\cite{bai2024graph,gao2024active,yin2024probabilistic}. In parallel, filtering-based methods have been widely adopted for their computational efficiency~\cite{carlone2014active,an2016ceiling,schlotfeldt2018anytime}. Additional mechanisms such as sub-mapping, sparse updates, and hierarchical map representations further improve robustness and scalability in large environments~\cite{xu2021invariant,ossenkopf2019long}. However, while these localization–mapping components strengthen estimation, they generally operate independently of the decision-making layer.

Given localization-mapping frameworks, the decision-making layer must balance exploration and exploitation by predicting how future actions will affect localization–mapping uncertainty and the resulting information gain. Information-theoretic approaches remain the dominant strategy, evaluating expected information gain via entropy and typically leveraging discrete map representations that enable probabilistic reasoning. Feature-based methods~\cite{chen2020active} provide compactness but lack sufficient spatial richness, motivating the use of dense representations such as occupancy grids~\cite{li2019deep} and OctoMap-based approaches~\cite{batinovic2021multi}. However, dense maps introduce substantial computational load~\cite{wang2020autonomous}, leading most approaches to rely on \emph{decoupling} localization and mapping and on careful \emph{parameter tuning} to maintain tractability in decision-making. More recent work seeks to reduce such heuristic dependence by introducing more interpretable decision-making mechanisms~\cite{valencia2018active,placed2022explorb,sansoni2025optimizing}, although a fully principled treatment of coupled uncertainty remains an open direction.

A key differentiating factor among information-theoretic approaches is the choice of entropy measure. The majority of work uses Shannon entropy (SE), whose mathematical properties align well with Gaussian and binary models common in exploration~\cite{ahmed2023active}. To obtain more flexible or mission-aware behavior, generalized entropies have been explored. Rényi entropy (RE) enables tuning sensitivity to uncertainty but often requires heuristic handling to avoid computational burdens~\cite{carrillo2018autonomous,popovic2020informative}. More recently, Behavioral entropy (BE) has been introduced in robotic exploration tasks~\cite{ghimire2025beasst,mandal2025behaviorally}, offering intuitive modulation of risk-sensitive behaviors. However, BE has not yet been integrated into a fully \emph{coupled} active exploration framework in which localization and mapping uncertainties jointly influence action selection. In this work, we explicitly address this gap by incorporating BE into a decision-making framework that accounts for tightly coupled localization–mapping uncertainty, enabling adaptive and uncertainty-aware action selection.

\begin{figure}[t]
    \centering
    \includegraphics[width=0.48\textwidth]{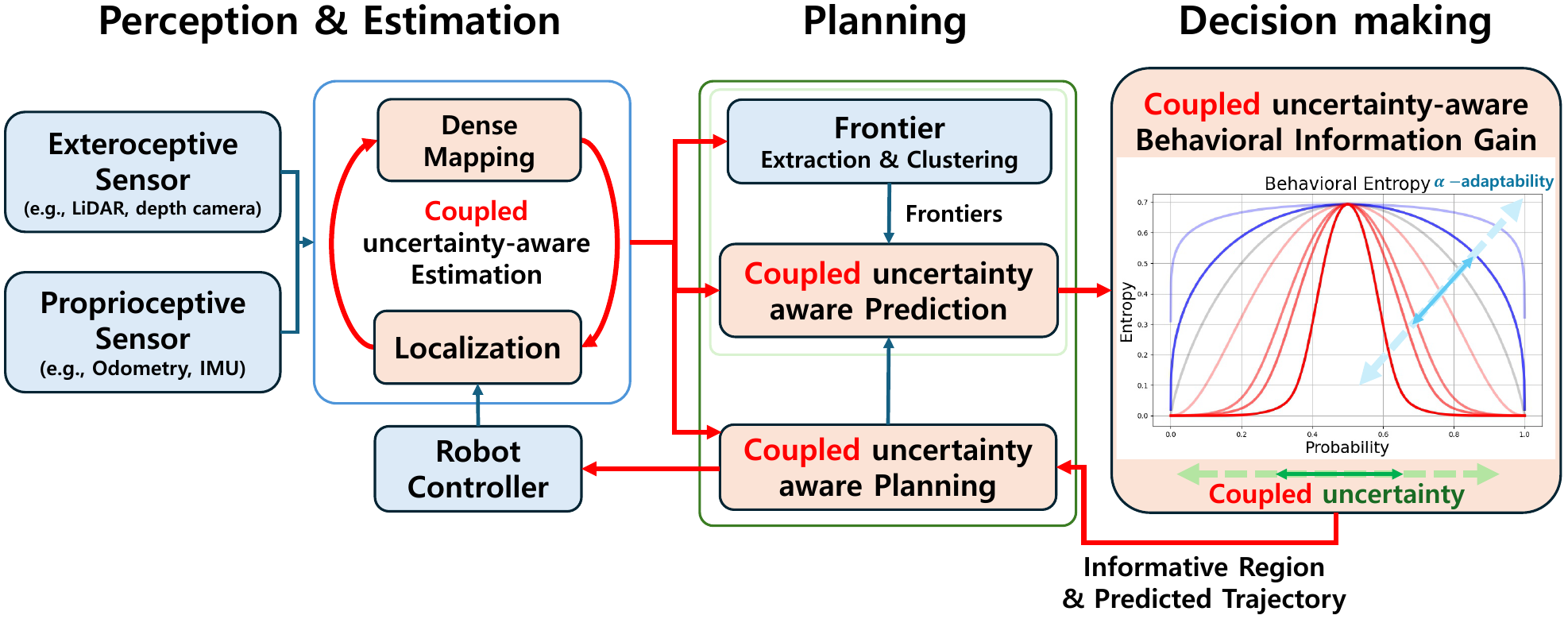}
    \caption{{\small Proposed uncertainty-aware active exploration framework that explicitly accounts for \emph{tightly} coupled uncertainty throughout the localization–mapping–decision pipeline. In perception and estimation, localization and dense mapping are jointly updated under coupled uncertainty in a computationally efficient manner, yielding an informative map. Based on this map, frontiers—defined as the boundaries between known and unknown regions—are extracted and clustered to generate multiple goal candidates. For each candidate, the predicted coupled uncertainty is propagated over a sequence of control actions. At each goal, the uncertainty-aware behavioral information gain is evaluated by considering both the coupled uncertainties and the parameter~$\alpha$. The exploration module then selects the goal that achieves a principled balance between exploration and exploitation, forwarding it to the navigation manager, which subsequently communicates it to the controller.}}
    \label{fig:framework}
\end{figure}

\section{Problem Formulation and System Model}
We address the problem of \emph{active robot exploration} in an initially \emph{unknown, static} environment using a single mobile platform. The robot is equipped with proprioceptive sensors (e.g., odometry or IMU) and a multi-beam exteroceptive sensor (e.g., LiDAR or depth camera) for localization and mapping. The goal is to select control actions that maximize the information gain in the environment (see Fig.~\ref{fig:framework}).

\subsection{System Modeling}

We model the environment as a bounded 2D or 3D dense occupancy grid. The occupancy grid at time $t$ is represented by the random variables $\mathbf{m}_t = \{m^i_t, \dots, m^M_t\}$, where $M$ is the total number of discretized cells, and $m^i_t \in \{0, 1\}$ is the binary random variable indicating the occupancy of the $i$-th cell. Specifically, $m^i_t = 0$ indicates an \emph{unoccupied} cell, while $m^i_t = 1$ indicates an \emph{occupied} cell. The dense occupancy map relies on the following standard assumption.
\smallskip
\begin{assump}\label{assum::indep_cell} \emph{(Independence of cells) Each occupancy cell is independent of the others. Therefore, the map can be represented as a factorized probability mass function, i.e.,}
\begin{align}
\label{eq::Factor_Map}
    p(\mathbf{m}_t) = \prod\nolimits_{i=1}^M p(m^i_t).
\end{align}
\end{assump}

The robot's state transition model and measurement model are defined as:
\begin{subequations}
\label{eq::NSSM}
\begin{align}
    &\mathbf{x}_t = f_{t|t-1}(\mathbf{x}_{t-1}, \mathbf{u}_t) + \mathbf{q}_{t-1}, \label{eq::motion_control} \\
    &\mathbf{z}_t = h_t(\mathbf{x}_t, \mathbf{m}_t) + \mathbf{r}_t, \label{eq::ext_sensor}
\end{align}
\end{subequations}
where $\mathbf{q}_{t-1}\sim\mathcal{N}(\mathbf{0}, \mathbf{Q}_{t-1})$ and $\mathbf{r}_t\sim\mathcal{N}(\mathbf{0}, \mathbf{R}_t)$\footnote{$\mathcal{N}(\mu, \Sigma)$ denotes a multivariate normal distribution with mean $\mu \in \real^n$ and covariance $\Sigma \in \mathcal{S}_{++}^n$, the set of $n \times n$ symmetric positive definite matrices.}. Here, $f_{t|t-1}(\mathbf{x}_{t-1}, \mathbf{u}_t)$ denotes the nonlinear motion model, where $\mathbf{x}_t \in \real^{n_x}$ represents the robot pose and $\mathbf{u}_t \in \real^{n_u}$ denotes the control input. Also, $h_t(\mathbf{x}_t, \mathbf{m}_t)$ denotes the nonlinear exteroceptive sensor model. The measurement $\mathbf{z}_t$ for a $N_b$-beam LiDAR is a vector of $N_b$ instantaneous distance readings from the sensor to the first obstacle along each of its beams. If a beam $k$ detects no obstacle, the reading $z^k_t$ defaults to the sensor's maximum detectable range, $z_{\text{max}}$. For simplicity of notation, we will omit $t$ and $t-1$ in the system parameters $(f_{t|t-1}, h_t, \mathbf{Q}_{t-1}, \mathbf{R}_t)$ throughout our discussion.

\subsection{Active Exploration Objective}

Following the formulation in~\cite{bourgault2002information}, we pose the information-theoretic active exploration problem as selecting an action sequence over a time horizon of $t'$ (starting from time $t+1$) that maximizes the information gain defined in \eqref{eq::InfoGain}, i.e.,
\begin{align} \label{eq::cost_original}
  \mathbf{u}_{t+1:t'}^{\star} &= \arg\!\!\!\!\!\!\max_{\mathbf{u}_{t+1:t'} \in \mathcal{U}_t} \mathbb{I}_G\bigr[\vect{u}_{t+1:t'}|\mathbf{z}_{t+1:t'}\bigr],
\end{align}
where $\mathcal{U}_t$ denotes the set of candidate action sequences at time $t$. Our objective is to develop a general framework for computing information gain with any generalized entropy measure, while balancing exploration and exploitation in a principled manner. We then incorporate BE into the formulation to enable adaptive decision-making under uncertainty. A detailed explanation of how to solve the problem is provided in Section~\ref{sec::Active_EXplore}.

\smallskip
\begin{rem} \label{remark_1}
\emph{(Challenge). Maximizing the information gain in \eqref{eq::cost_original} is challenging due to the inherent nonlinearity and non-separability\footnote{For random variables $a$ and $b$, the joint entropy $\mathbb{H}(a,b)$ is separable if $\mathbb{H}(a,b) = \mathbb{H}(a) + \mathbb{H}(b|a)$, where $\mathbb{H}(b|a)$ denotes the conditional entropy of $b$ given $a$. This property holds \emph{if and only if} $\mathbb{H}$ is a scaled form of SE~\cite{shannon1948mathematical}.} of generalized entropy measures. Consequently, the joint entropy of localization and mapping in \eqref{eq::InfoGain} cannot be decoupled, unlike in SE-based frameworks~\cite{stachniss2005information}.}
\end{rem}

\medskip
The following sections present a localization–mapping framework that resolves coupled uncertainty with scalability, making it possible to apply \eqref{eq::cost_original} using any generalized entropy measure.

\section{Coupled Uncertainty-aware Estimation}
This section presents scalable and efficient estimation rules that generate the predictive, coupled uncertainty metrics required for principled active exploration. While conventional SLAM focuses solely on robust state estimation, active exploration requires the estimation module to provide spatially informative uncertainty measures that can be accurately predicted and integrated into an information gain utility function. Current approaches for dense mapping either decouple localization and mapping—which fundamentally breaks the link needed for accurate uncertainty prediction and active decision-making—or become computationally intractable due to including the full map state in the optimization. Therefore, to achieve principled and scalable uncertainty management for active exploration, a novel, coupled estimation rule is mandatory.

To address this need, we begin by introducing a novel likelihood model for beam measurements. This model is the foundation that enables the reciprocal coupling, as it incorporates dense occupancy map uncertainty and sensor uncertainty within a unified Gaussian formulation. Building on this core model, we then propose:
\begin{enumerate}
    \item A Gaussian-based filtering approach with \emph{dual} parameter forms for localization, $p(\mathbf{x}_t \mid \mathbf{u}_{1:t}, \mathbf{z}_{1:t})$, designed to embed map uncertainty.
    \item An occupancy update formulation for dense mapping, $p(\mathbf{m}_t \mid \mathbf{u}_{1:t}, \mathbf{z}_{1:t})$, designed to account for localization uncertainty and enhance robustness.
\end{enumerate}
By achieving analytical uncertainty \emph{coupling} in a computationally efficient manner, the proposed framework provides the robust and scalable foundation required for active exploration.

\begin{figure}[t]
\centering
\begin{minipage}[t]{0.24\textwidth}
\centering
 \includegraphics[width=\textwidth]{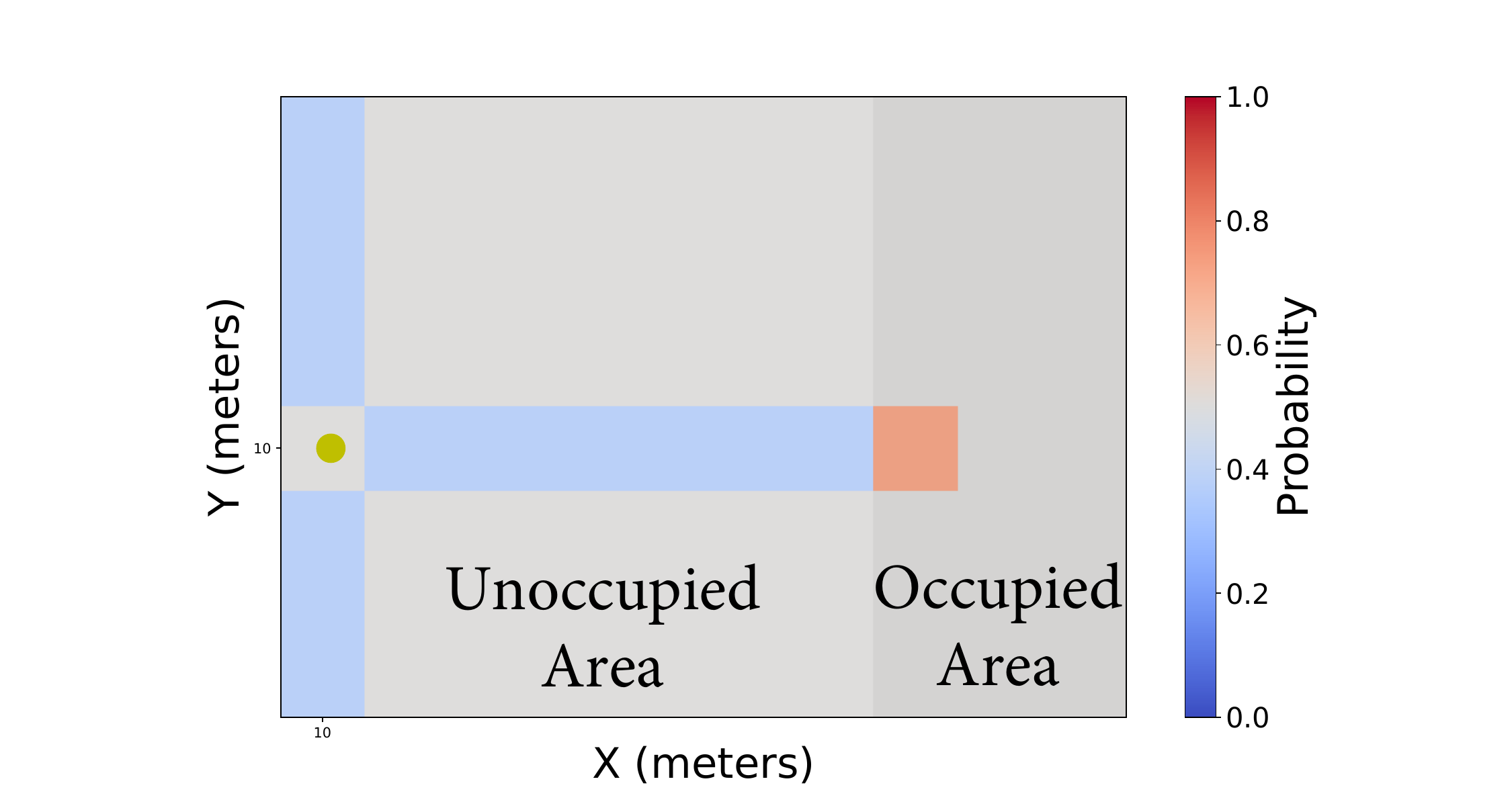}
{{\scriptsize (a) Map updates based on inverse model}}
 \end{minipage}
\vspace{1pt}
 \begin{minipage}[t]{0.24\textwidth}
\centering
\includegraphics[width=\textwidth]{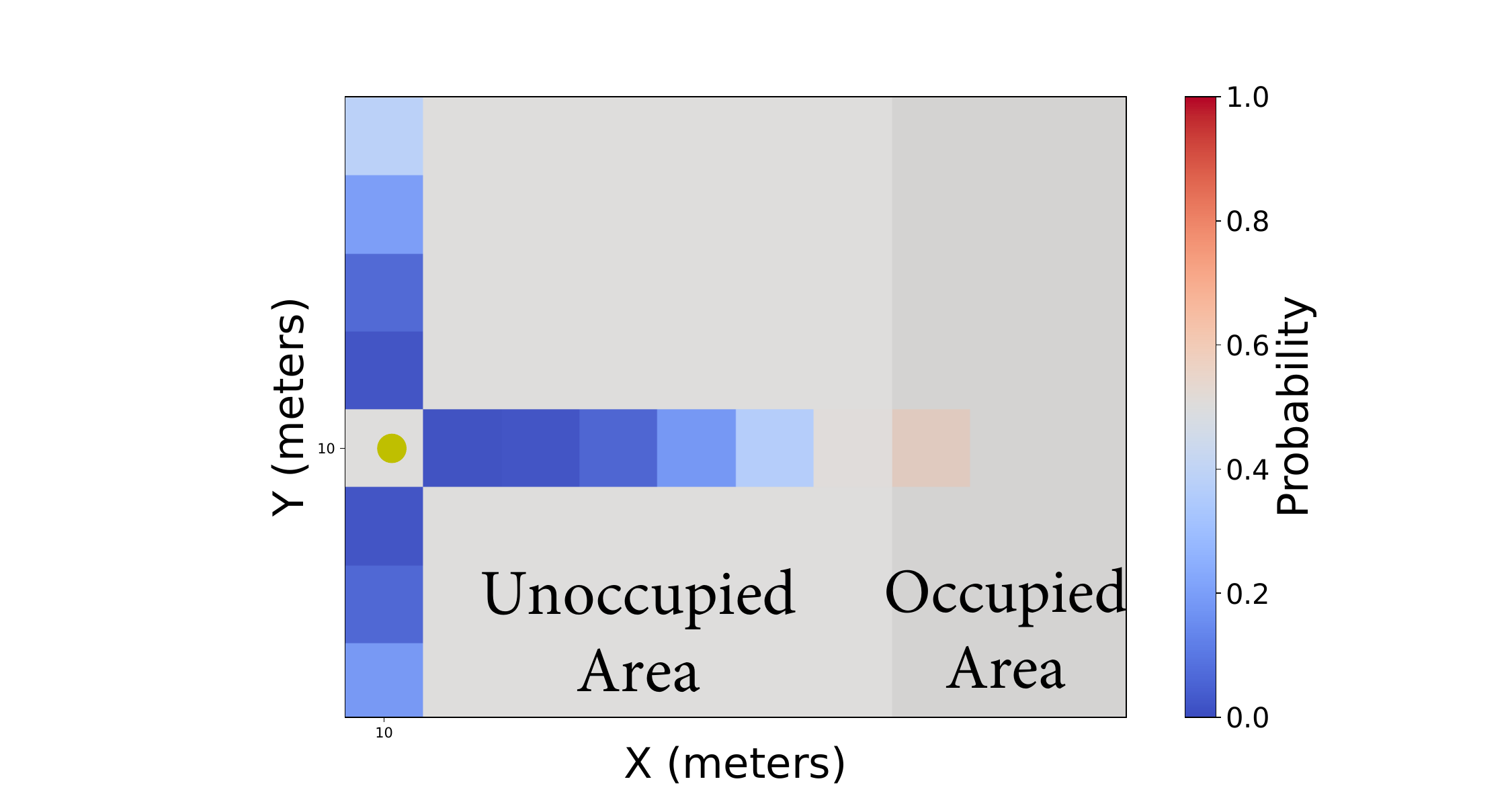}
 {{\scriptsize (b) Map updates based on our model}}
\end{minipage}
\caption{{\small Comparison of map updates between (a) the inverse model and (b) the proposed model. The inverse sensor model simplifies the update process by assigning uniform probabilities, whereas the proposed model accounts for beam-wise uncertainty, resulting in more informative map estimates.}}
\label{fig::4_Likelihood_comparison}
\end{figure}

\subsection{The Foundational Likelihood Model for Coupled Estimation}

In the context of dense occupancy maps, existing methods for handling sensor measurements typically employ the \emph{inverse sensor model}~\cite{thrun2002probabilistic}. Rather than using a principled approach that combines a detailed measurement likelihood model with the propagated system model, the inverse sensor model simplifies the occupancy map updates. It treats the endpoint of a beam measurement as a \emph{hit (occupied)} and the traversed cells as \emph{free (unoccupied)}, assigning predefined, uniform probability ratios (often represented in log-odds form)—which serves as the evidence term for recursive updates—along the beam for recursive Bayesian updates. This simplification inherently neglects the spatial and beam-wise characteristics of sensor noise and uncertainty (see Fig.~\ref{fig::4_Likelihood_comparison}(a)). This uniform treatment creates two major shortcomings that hinder effective active exploration:
\begin{enumerate}
\item Inaccurate Map Uncertainty: By simplifying the sensor model, the resulting occupancy probabilities often fail to capture the true uncertainty distribution, leading to less informative map estimates.
 \item Inability to Couple Uncertainties: The model does not naturally lend itself to a closed-form marginalization that can embed map uncertainty into localization or vice versa, thereby preventing the principled management of coupled uncertainty essential for decision-making.
\end{enumerate}


To overcome these limitations and provide the necessary predictive coupled uncertainty metrics required for principled active exploration, we introduce a novel likelihood model that incorporates dense occupancy maps and sensor uncertainties under a unified Gaussian formulation.

In our representation, dense occupancy maps consist of discrete grid cells, $\mathbf{m}_t$, each encoding the occupancy probability. At time $t$, the beam measurements consist of $N_b$ beams, denoted as $\mathbf{z}_t = \{z^k_t\}_{k=1}^{N_b}$. For the map cell $m^i_t$, the $k$-th beam $z^k_t \in \mathbb{R}_{\geq 0}$ interacts with two sets of cells: those it passes through (free), $i \in \mathcal{I}^{p_k}_t$, and the cell it hits, $i \in \mathcal{I}^{h_k}_t$. Thus, the set of all cells associated with the $k$-th beam is given by $\mathcal{I}^k_t = \mathcal{I}^{p_k}_t \cup \mathcal{I}^{h_k}_t$. Analogously, for the full set of beams $\mathbf{z}_t$, we define $\mathcal{I}^p_t$, $\mathcal{I}^h_t$, and $\mathcal{I}_t$ as the sets of all traversed, hit, and related cells, respectively. The beam measurement model relies on the following assumption:
\begin{assump}[Conditional independence of $\mathbf{z}_t$] \label{assum::indep_obs}\emph{Given the robot's state $\mathbf{x}_t$ and the map $m^i_t$, the likelihood model for beam measurements factorized as:}
\begin{align}
p(\mathbf{z}_t| \mathbf{x}_t, \mathbf{m}_t) = \prod\nolimits_{k = 1}^{N_b}\prod\nolimits_{i \in \mathcal{I}^k_t} p(z^k_t| \mathbf{x}_t, m^i_t),
\end{align}
\emph{where the outer product runs over each $k$-th beam, and the inner product runs over all map cells $i$ that interact with the $k$-th beam.}
\end{assump}
This \emph{conditional independence assumption} is a necessary simplification to maintain \emph{computational tractability} for high-dimensional dense maps, following standard practice in efficient estimation frameworks.

\medskip
Under Assumption~\ref{assum::indep_obs}, the likelihood model for the $k$-th beam $z^k_t$ is conditioned on the robot's localization $\mathbf{x}_t$ and the map cell $m^i_t$. We define this likelihood as a mixture of two Gaussian distributions and refer to it as DVL (Differential Variance Likelihood) model :
\begin{align} \label{eq::likelihood_beam}
p(z^k_t|\mathbf{x}_t, m^i_t) = \left\{
\begin{array}{ccc}
\!\! \mathcal{N}\bigl(z^k_t; h(\mathbf{x}_t, m^i_t), \, R_o\bigl), & m^i_t = 1\\
\! \!\mathcal{N}\bigl(z^k_t; h(\mathbf{x}_t, m^i_t), \, R_u\bigl), &m^i_t = 0\end{array} \right.\!\!,
\end{align}
where $h(\mathbf{x}_t, m^i_t) = \| \mathbf{x}_t - \mathbf{\chi}(m^i_t)\|_2 \in \mathbb{R}_{\geq 0}$ is the Euclidean distance.

\medskip

\noindent The premise of this model is to quantify how likely the observed reading $z^k_t$ is if cell $i$ is truly occupied, versus if it is truly unoccupied. The key mechanism is the residual between the measurement and the expected distance, $z^k_t - h(\mathbf{x}_t, m^i_t)$. A small residual signifies high consistency between the sensor reading and the geometric prediction.

\medskip

The variances $R_o$ (occupied) and $R_u$ (unoccupied) are set such that $R_u \gg R_o$ ($5R_o \leq R_u \leq 15R_o$). This design models the \emph{inherently higher uncertainty} associated with a beam traversing unoccupied space ($R_u$) compared to a precise obstacle hit ($R_o$). The ratio of these two likelihoods directly controls the filter's sensitivity and confidence.

\begin{figure}[t]
\centering
\begin{minipage}[t]{0.35\textwidth}
\centering
\includegraphics[width=\textwidth]{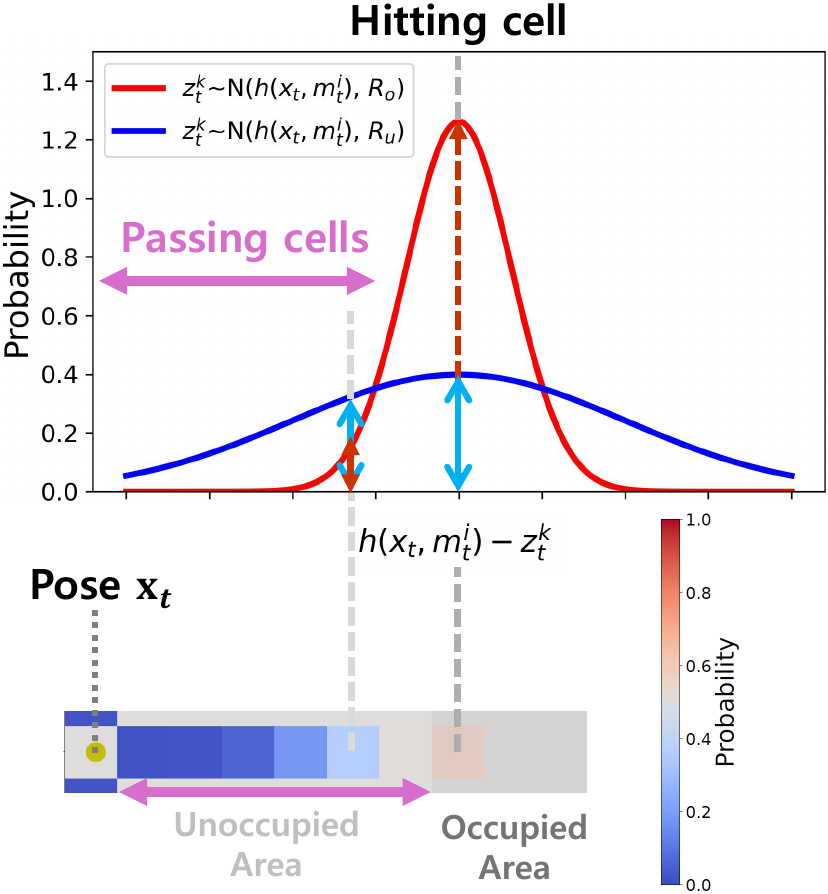}
\end{minipage}
\caption{{\small \textbf{(Top)} The likelihood of the $k$-th beam measurement~\eqref{eq::likelihood_beam} depends on the beam residual $h(\mathbf{x}_t, m^i_t) - z^k_t$ and on whether the corresponding cell is occupied or unoccupied (red vs. blue curves). When the residual is near zero, the occupied-cell likelihood (red) dominates; as the residual grows, the unoccupied-cell likelihood (blue) becomes more significant. \textbf{(Bottom)} The occupancy map is updated based on the probability ratio between the two beam likelihoods (arrows). Additional details are provided in Section~\ref{sec::mapping}.}}
\label{fig::3_Likelihood}
\end{figure}

This formulation offers two critical advantages over the simplified Inverse Sensor Model:

\begin{enumerate}
\item Direct Analytical Coupling of Uncertainty: The choice of the Gaussian form ($\mathcal{N}$) is an analytical approximation that enables efficient analytical marginalization in the estimation framework. This is the foundation for the tight coupling of localization ($\mathbf{x}_t$) and map ($\mathbf{m}_t$) uncertainties, which is crucial for decision-making systems that predict information gain (Active Exploration).

\item Informative, Spatially Heterogeneous Map Updates: The model's reliance on the beam residual, $z^k_t - h(\mathbf{x}_t, m^i_t)$, captures beam-wise uncertainty that the uniform Inverse Model neglects. Cells traversed near the robot, where the geometric distance is inherently more robust to localization error, receive a higher confidence `free' update. Since the occupancy map is updated based on the probability ratio between the two beam likelihoods (occupied vs. unoccupied), our likelihood model produces map estimates that are more informative and spatially heterogeneous, accurately reflecting confidence in cleared space; see Fig.~\ref{fig::3_Likelihood}.
\end{enumerate}

These properties establish the model as an ideal foundation for robust, coupled, and scalable $\mathcal{O}(n_x^3)$ estimation rules, directly supporting advanced active exploration objectives.

\subsection{Localization under Coupled Map Uncertainty} 

The core difference between conventional filtering and our approach centers on the computation of the Likelihood term $p(\mathbf{z}_t|\mathbf{x}_t)$ within the Bayesian update. Classical filtering often adopts a decoupled formulation in which the localization update is simplified by treating the map $\mathbf{m}_t$ as certain. In practice, this assumes an occupied cell ($m^i_t=1$) once its occupancy probability exceeds a predefined threshold, leading to the approximation $p(\mathbf{z}_t|\mathbf{x}_t; m^i_t=1)$. This simplification ignores map uncertainty, resulting in a measurement covariance that only accounts for fixed sensor noise ($R_{\text{sensor}}$). Consequently, the total localization covariance ($\Sigma_t$) is underestimated, critically compromising the accuracy required for predicting information gain during active exploration. To overcome this, our method rigorously achieves analytical coupling: the Gaussian structure of our DVL Likelihood Model enables the closed-form marginalization of the predicted map state $p(m^i_t|\mathbf{z}_{t-1})$ out of the joint likelihood. This process yields a direct, coupled measurement likelihood $p(\mathbf{z}_t|\mathbf{x}_t)$ where the resulting equivalent measurement covariance $R^i_{\mathbf{x}} = (p^i_t)^2 R_o + (1 - p^i_t)^2 R_u$, given in~\eqref{eq::R_x}, is dynamically weighted by the map's current confidence $p(m^i_t)$. This analytical embedding ensures that the localization covariance $\Sigma_t$ is rigorously and accurately influenced by the predicted map uncertainty (Lemma~\ref{lem::4_1}), thus breaking the traditional separation and providing the necessary rigorous covariance tracking for principled decision-making.

In the prediction step, we follow the conventional Bayesian filtering approach to compute the predicted distribution $p(\mathbf{x}_t|\mathbf{u}_{1:t},\mathbf{z}_{1:t-1})$ by marginalizing the motion model with the previous posterior distribution. This operation is performed in the moment form (mean and covariance) and is mathematically represented as a convolution of the motion model and the previous state:
\begin{align} \label{eq::pred_motion}
    &p(\mathbf{x}_t|\mathbf{u}_{1:t},\mathbf{z}_{1:t-1}) \nonumber \\
    &= \int p(\mathbf{x}_t|\mathbf{x}_{t-1}, \mathbf{u}_t)p(\mathbf{x}_{t-1}|\mathbf{u}_{1:t-1},\mathbf{z}_{1:t-1}) \text{d}\mathbf{x}_{t-1} \nonumber \\
    &\approx \mathcal{N}(\mathbf{x}_t; \bar{\mu}_t, \bar{\Sigma}_t).
\end{align}
Using $p(\mathbf{x}_t|\mathbf{x}_{t-1}, \mathbf{u}_t) = \mathcal{N}(\mathbf{x}_t; f(\mathbf{x}_{t-1}, \mathbf{u}_t), \mathbf{Q})$ from the state transition model and $p(\mathbf{x}_{t-1}|\mathbf{u}_{1:t-1},\mathbf{z}_{1:t-1}) = \mathcal{N}(\mathbf{x}_{t-1}; \mu_{t-1}, \Sigma_{t-1})$ from the previous posterior, the resulting predicted mean is $\bar{\mu}_t = f(\mu_{t-1},\mathbf{u}_t)$ and the predicted covariance is $\bar{\Sigma}_t = \mathbf{F}_{\mathbf{x}} \Sigma_{t-1} \mathbf{F}_{\mathbf{x}}^{\top} + \mathbf{Q}$, where $\mathbf{F}_{\mathbf{x}} = \nabla_{\mathbf{x}} f(\mu_{t-1}, \mathbf{u}_t)$ is the Jacobian of the motion function. The approximation in~\eqref{eq::pred_motion} arises from the necessary linearization of the non-linear motion model $f(\cdot)$ via the Jacobian matrix $F_{\mathbf{x}}$ for covariance propagation (i.e., the Extended Kalman Filter approach).


In our framework, the update step is where the localization state is rigorously coupled with the predicted map, defining our primary innovation. Unlike standard filtering, which only fuses sensor data, our update incorporates map uncertainty by analytically marginalizing over the predicted map state $p(\mathbf{m}_t|\mathbf{u}_{1:t-1}, \mathbf{z}_{1:t-1})$. This process transforms the joint likelihood $p(\mathbf{z}_t|\mathbf{x}_t, \mathbf{m}_t)$ into the direct, coupled measurement likelihood $p(\mathbf{z}_t|\mathbf{x}_t)$ that inherently carries map uncertainty within its covariance.

We begin by applying Bayes' rule to represent the posterior of localization:
\begin{align}
&p(\mathbf{x}_t|\mathbf{u}_{1:t},\mathbf{z}_{1:t}) \propto p(\mathbf{z}_t|\mathbf{x}_t) p(\mathbf{x}_t|\mathbf{u}_{1:t},\mathbf{z}_{1:t-1}). \label{eq::post_start}
\end{align}
The crucial step is to expand the coupled likelihood term $p(\mathbf{z}_t|\mathbf{x}_t)= \prod\nolimits_{\{ k|\mathcal{I}^{h_k}_t\neq \emptyset\}} p(z^k_t|\mathbf{x}_t)^{\frac{\gamma}{N^h_t}}$ by marginalizing the beam measurement $z^k_t$ over the binary occupancy state $m^i_t$ of the hit cells $i\in\mathcal{I}^{h_k}_t$, i.e.,
$$p(z^k_t|\mathbf{x}_t)=\sum\nolimits_{m^i_t \in \{0, 1\}}  p(z^k_t| \mathbf{x}_t, m^i_t) p(m^i_t|\mathbf{u}_{1:t-1},\mathbf{z}_{1:t-1}),$$
where $p(m^i_t|\mathbf{u}_{1:t-1},\mathbf{z}_{1:t-1})$ is the predicted occupancy probability (map uncertainty).

 Following Assumption~\ref{assum::indep_obs}, and substituting the marginalization the expression becomes:
\begin{align}
 &p(\mathbf{x}_t|\mathbf{u}_{1:t},\mathbf{z}_{1:t}) \nonumber \\
 &\propto \prod\nolimits_{\{ k|\mathcal{I}^{h_k}_t\neq \emptyset\}} p(z^k_t|\mathbf{x}_t)^{\frac{\gamma}{N^h_t}} p(\mathbf{x}_t|\mathbf{u}_{1:t},\mathbf{z}_{1:t-1}) \nonumber \\
 &\!=\! \!\!\!\!\!\!\! \prod_{\{ k|\mathcal{I}^{h_k}_t\neq \emptyset\}} \!\!\!\!\!\!\!\! \prod\nolimits_{i \in \mathcal{I}^{h_k}_t} \!\! \biggl(\sum_{m^i_t \in \{0, 1\}} \!\!\!\!\! p(z^k_t| \mathbf{x}_t, m^i_t) p(m^i_{t}|\mathbf{u}_{1:t-1},\mathbf{z}_{1:t-1})\! \biggl)^{\frac{\gamma}{N^h_t}} \nonumber \\
&\quad~ \times \underbrace{p(\mathbf{x}_t|\mathbf{u}_{1:t},\mathbf{z}_{1:t-1})}_{\text{$\mathcal{N}(\mathbf{x}_t;\bar{\mu}_t, \bar{\Sigma}_t)$}}. \label{eq::marginalization_full}
\end{align}
The sum over $m^i_t$ in $\sum_{m^i_t \in \{0, 1\}}  p(z^k_t| \mathbf{x}_t, m^i_t) p(m^i_{t}|\mathbf{u}_{1:t-1},\mathbf{z}_{1:t-1})$ represents a Mixture of Two Gaussians (from~\eqref{eq::likelihood_beam}).
We employ a Gaussian approximation for this mixture term, which yields a single, equivalent Gaussian $\mathcal{N}(z^k_t; h(\mathbf{x}_t, m^i_t), \, R^i_{\mathbf{x}})^{\frac{\gamma}{N^h_t}}$. The resulting equivalent measurement covariance $R^i_{\mathbf{x}}$ is the key to coupling:
\begin{equation}\label{eq::R_x}
    R^i_{\mathbf{x}} = (p^i_t)^2 R_o + (1 - p^i_t)^2 R_u,
\end{equation}
where $p^i_t\equiv p(m^i_{t-1}=1|\mathbf{u}_{1:t-1},\mathbf{z}_{1:t-1})$. Note that due to the static environment assumption, we have $$p^i_t=p(m^i_t=1|\mathbf{u}_{1:t-1},\mathbf{z}_{1:t-1}) = p(m^i_{t-1}=1|\mathbf{u}_{1:t-1},\mathbf{z}_{1:t-1}).$$ 
This structure ensures that the localization update is directly modulated by the confidence $p^i_t$ in the map cell used for the measurement. By linearizing this Gaussian approximation in the information form and combining it with the predicted pose distribution $\mathcal{N}(\mathbf{x}_t;\bar{\mu}_t, \bar{\Sigma}_t)$, we arrive at the closed-form update rule summarized in Lemma~\ref{lem::4_1}.

\begin{lem}[Update for localization] \label{lem::4_1}
\emph{Let Assumption \ref{assum::indep_cell} and \ref{assum::indep_obs} hold. Given $p(\mathbf{m}_{t-1}|\mathbf{u}_{1:t-1}, \mathbf{z}_{1:t-1})$,~\eqref{eq::likelihood_beam}, and~\eqref{eq::pred_motion}, the posterior $p(\mathbf{x}_t|\mathbf{u}_{1:t}, \mathbf{z}_{1:t})$ can be approximated by a Gaussian distribution $\mathcal{N}(\mathbf{x}_t; \mu_t, \Sigma_t)$, where the mean $\mu_t$ and covariance $\Sigma_t$ are:}
\begin{subequations} \label{eq::update_localization}
\begin{align}
\Sigma_t &= \biggl(\bar{\Sigma}_t^{-1} + \frac{\gamma}{N^h_t} \!\!\!\!\! \sum_{\{ k \mid \mathcal{I}^{h_k}_t \neq \emptyset \}} \!\!\!\!\!\! \sum\nolimits_{i \in \mathcal{I}^{h_k}_t} (\mathbf{H}_{\mathbf{x}}^i)^{\top} (R^i_{\mathbf{x}})^{-1} \mathbf{H}_{\mathbf{x}}^i\biggl)^{-1} \!\!\!\!\!\!\!,\!\!\! \\
\mu_t &= \Sigma_t \biggl( \bar{\Sigma}_t^{-1} \bar{\mu}_t + \frac{\gamma}{N^h_t} \!\!\!\!\! \sum_{\{ k|\mathcal{I}^{h_k}_t\neq \emptyset\}} \!\!\!\!\!\! \sum\nolimits_{i \in \mathcal{I}^{h_k}_t} (\mathbf{H}_{\mathbf{x}}^i)^{\top} (R^i_{\mathbf{x}})^{-1} \nonumber \\
&\qquad\qquad\qquad\quad\times\left[ z^k_t - h(\mathbf{x}_t, m^i_t) + \mathbf{H}_{\mathbf{x}}^i \bar{\mu}_t \right] \biggl),
\end{align}
\end{subequations}
with $R^i_{\mathbf{x}}$ defined in \eqref{eq::R_x}, where $p^i_t = p(m^i_t = 1 | \mathbf{u}_{1:t-1}, \mathbf{z}_{1:t-1})$ and $\mathbf{H}_{\mathbf{x}}^i = \nabla_{\mathbf{x}} h(\bar{\mu}_t, m^i_t)$. Here, $N^h_t=\sum_{\{ k \mid \mathcal{I}^{h_k}_t \neq \emptyset \}} \sum_{i \in \mathcal{I}^{h_k}_t} 1$ represents the total number of hitting cells at time $t$ and $\gamma$ is an information-weighting coefficient. \boxend
\end{lem}
\medskip
In the update step, we use only the hit cells $(N^h_t \leq |\mathcal{I}_t|)$, where $|\cdot|$ is the cardinality, since hit cells provide more precise information than pass-through cells.

\begin{figure}[t]
\centering
\begin{minipage}[t]{0.24\textwidth}
\centering
\includegraphics[width=\textwidth]{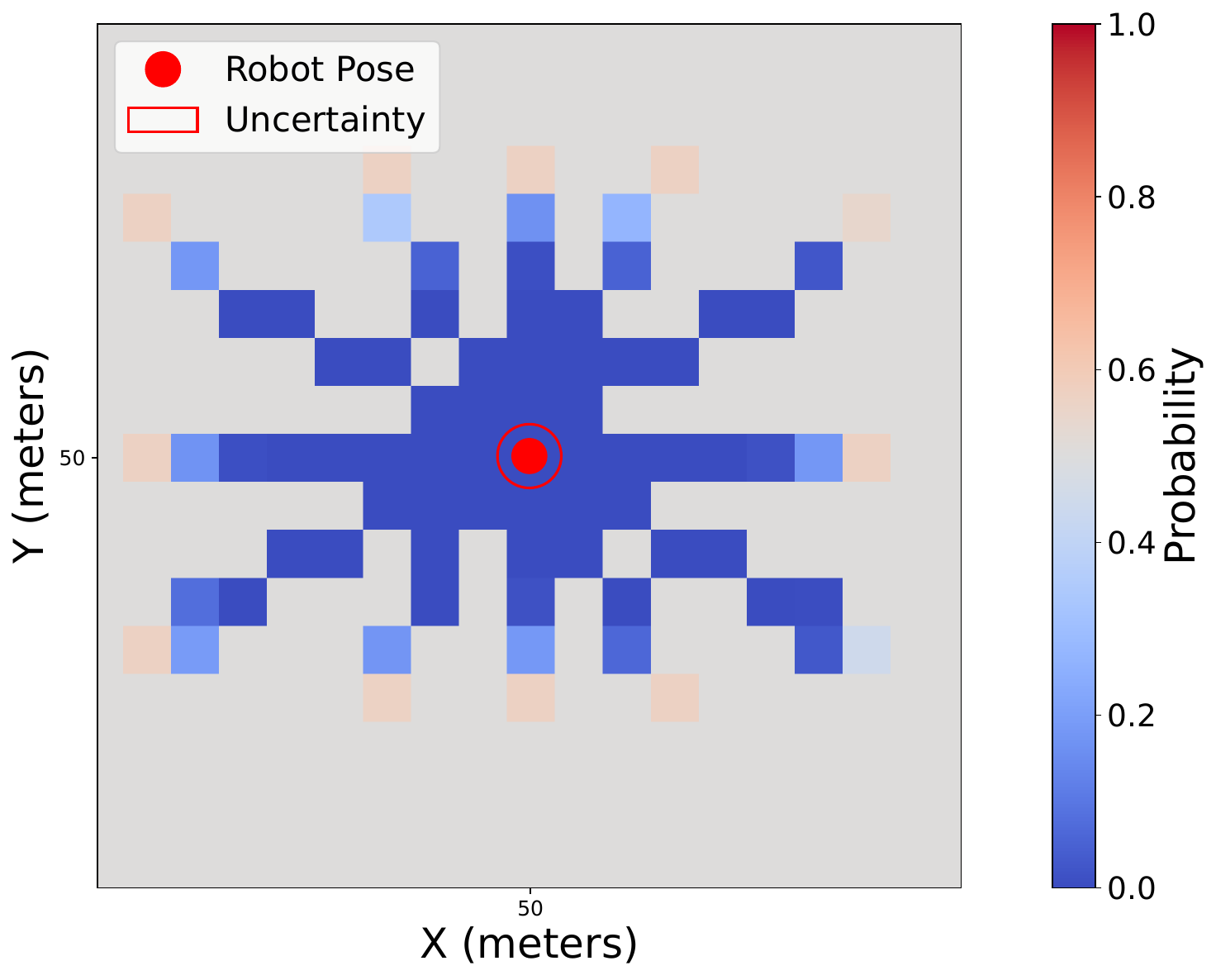}
{{\scriptsize (a) Low localization uncertainty}}
\end{minipage}
\vspace{1pt}
\begin{minipage}[t]{0.24\textwidth}
\centering
\includegraphics[width=\textwidth]{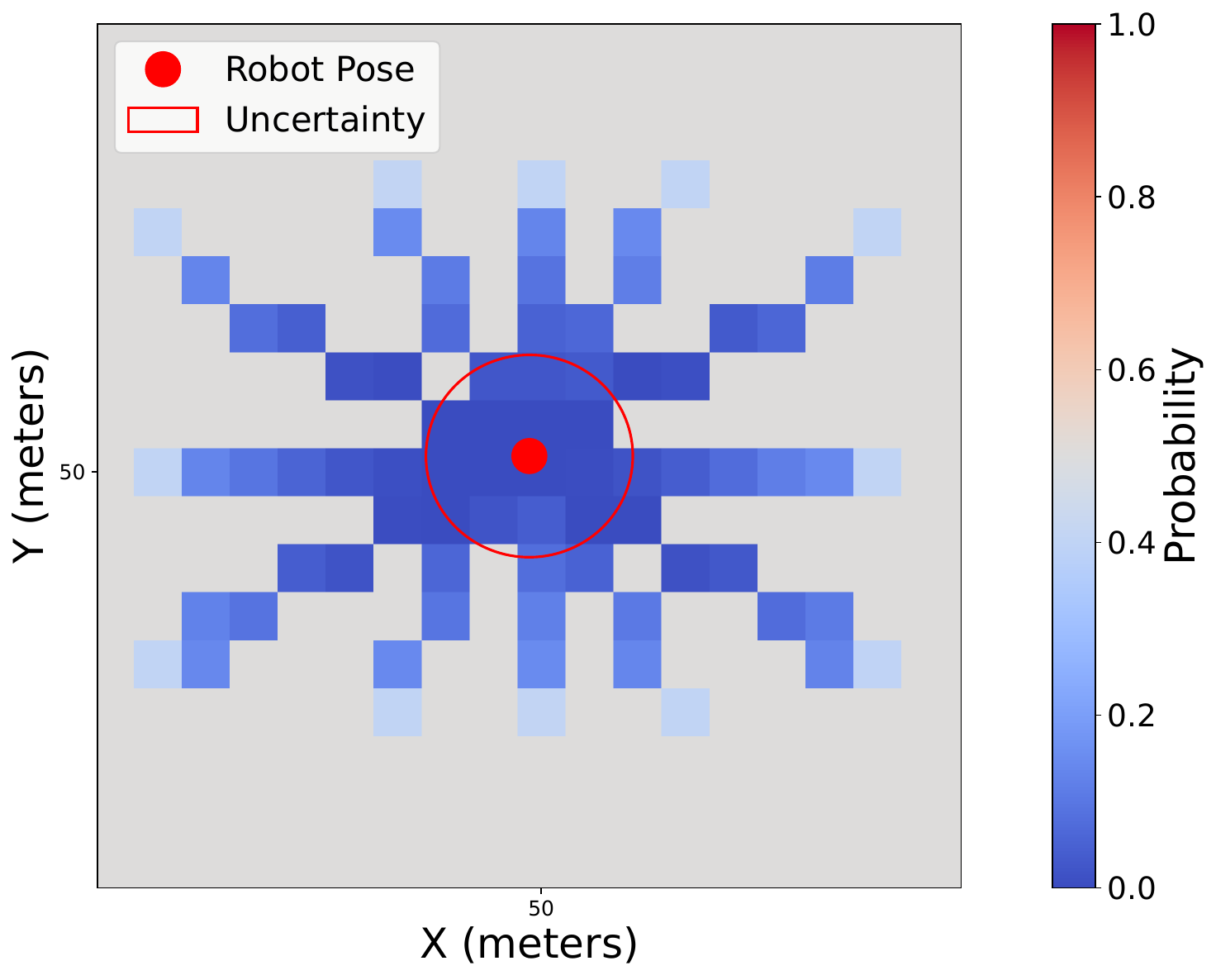}
{{\scriptsize (b) High localization uncertainty}}
\end{minipage}
\caption{{\small Map updates in \textsf{T-BayesMap} depend on localization uncertainty, represented by the size of the uncertainty ellipsoid. (a) When localization uncertainty is low, both occupied and unoccupied cells are strongly updated, meaning their probabilities change significantly. (b) In contrast, when localization uncertainty is high, updates to both occupied and unoccupied cells are marginal, and their probabilities change only slightly (i.e., conservative updates). Even the update in the probability of occupied cells remains below $0.5$ due to the high localization uncertainty.}}
\label{fig:7_Map_loc_uncertainty}
\end{figure}

The update rule for localization exhibits the following properties:
\begin{enumerate}
\item \emph{Localization with coupled map uncertainty:} The predicted map uncertainty of the $i$-th hit cell is incorporated into the localization update as a weighted combination of the occupied and unoccupied uncertainties : $R^i_{\mathbf{x}} = (p^i_t)^2 R_o + (1 - p^i_t)^2 R_u$, where $p^i_t = p(m^i_t = 1 | \mathbf{u}_{1:t-1}, \mathbf{z}_{1:t-1})$. As a result, localization uncertainty is inherently coupled with map uncertainty.

\item \emph{Computation efficiency and scalability:} By embedding map uncertainty, the computational complexity of localization becomes $\mathcal{O}(n_x^3) + \mathcal{O}(n_x^2 N^h_t)$, where the first term arises from the matrix inversion required for conversion between the information and moment forms, and the second term from the beam measurement update. The complexity depends only on the sensor specification (number of beams $N_b$) and the occupancy grid resolution, which are fixed and determine $N^h_t \leq |\mathcal{I}_t|$, but not on the environment size ($M$); therefore, it remains \emph{constant} with respect to the environment size.

\item \emph{Robustness:} The likelihood is tempered by the factors $N_t^h$ and $\gamma$, yielding $\mathcal{N}(z^k_t; h(\mathbf{x}_t, m^i_t), \, R^i_{\mathbf{x}})^{\frac{\gamma}{N^h_t}}$, which acts as a weighted likelihood that balances multi beam contributions with the predicted motion in~\eqref{eq::pred_motion}. This tempered update reduces the influence of inconsistent or outlier exteroceptive measurements, thereby improving robustness during the correction step.\footnote{Tempered likelihoods are widely used in generalized Bayesian inference to reduce the influence of unreliable observations; see~\cite{bissiri2016general}.}

\item \emph{Estimation vs. prediction for decision-making:} During estimation, a point-to-map registration can be applied in the update form—commonly used in filtering-based LiDAR odometry approaches (e.g.,~\cite{xu2021fast})—in which orientation is corrected based on the predicted map uncertainty and $R^i_{\mathbf{x}}$ in~\eqref{eq::R_x}. In contrast, during prediction for decision-making, no orientation correction is applied; instead, ray-based updates under the predicted map uncertainty are used to compute information gain.

\end{enumerate}
\subsection{\textsf{T-BayesMap}: Mapping under Coupled Localization Uncertainty} \label{sec::mapping}

The mapping update closes the uncertainty loop by rigorously embedding the robot's current localization uncertainty ($\Sigma_t$) into the map state update. This ensures that the map is updated conservatively in areas of high pose uncertainty, a key differentiator from conventional systems where mapping is often treated deterministically or relies on simple, decoupled uncertainty models.

We propose \textsf{T-BayesMap}, a binary Bayesian update that uses tempered, weight-modulated marginalized likelihoods to achieve more informative and robust map updates under coupled localization uncertainty.


The posterior probability of the $i$-th map cell, $p(m^i_t|\mathbf{u}_{1:t},\mathbf{z}_{1:t})$, is derived using Bayes' theorem conditioned on the accumulated measurements. Crucially, we marginalize over the updated localization pose $p(\mathbf{x}_t|\mathbf{u}_{1:t},\mathbf{z}_{1:t})$ to rigorously incorporate localization uncertainty into the map likelihood:
\begin{align}
 &p(m^i_t|\mathbf{u}_{1:t},\mathbf{z}_{1:t}) \nonumber \\
&\quad=\eta_{m^i_t} p(\mathbf{z}_t|m^i_t) p(m^i_t|\mathbf{u}_{1:t-1},\mathbf{z}_{1:t-1}) \nonumber \\
&\quad\propto \biggr(\int_{\mathbf{x}_t} p(z^k_t| \mathbf{x}_t, m^i_t)~p(\mathbf{x}_t|\mathbf{u}_{1:t},\mathbf{z}_{1:t}) \text{d}\mathbf{x}_t\biggr)^{w^i_t} \nonumber \\
 &\quad~~~~~\times p(m^i_t|\mathbf{u}_{1:t-1},\mathbf{z}_{1:t-1}), \label{eq::marginalization_map_integral}
\end{align}
where $\eta_{m^i_t}$ is the normalization term and $w^i_t$ is a tempered weight related to the accumulated number of beam interactions, as explained further below. The term inside the integral is the product of two Gaussian distributions: $p(z^k_t| \mathbf{x}_t, m^i_t)$ (the beam likelihood~\eqref{eq::likelihood_beam}) and $p(\mathbf{x}_t|\mathbf{u}_{1:t},\mathbf{z}_{1:t})$ (the updated pose posterior~\eqref{eq::update_localization}).

Since both distributions are Gaussian and the measurement model is locally linearized, the integral admits the following closed-form Gaussian approximation for the marginalized measurement likelihood:
\begin{align}
\int_{\mathbf{x}_t}\!\!\!\!\! \underbrace{p(z^k_t| \mathbf{x}_t, m^i_t)}_{\text{$\mathcal{N}(z^k_t; h(\mathbf{x}_t, m^i_t), \, R_{\{o, u\}})$}} \!\!\! \underbrace{p(\mathbf{x}_t|\mathbf{u}_{1:t},\mathbf{z}_{1:t})}_{\mathcal{N}(\mathbf{x}_t; \mu_t, \Sigma_t)} \text{d}\mathbf{x}_t \approx \mathcal{N}(z^k_t; \hat{\mu}_t, \hat{\Sigma}^i_t). \label{eq::map_marginalization_result}
\end{align}
This result analytically embeds the localization uncertainty $\Sigma_t$ (from the pose posterior) into the effective measurement covariance $\hat{\Sigma}^i_t$ for the map update. Specifically, by linearizing the measurement function $h(\mathbf{x}_t, m^i_t)$ around the updated mean pose $\mu_t$, the resulting mean is $\hat{\mu}_t = h(\mu_t, m^i_t)$ and the covariance becomes:
\begin{align}
\hat{\Sigma}^i_t = \left\{ 
\begin{array}{ccc}
\mathbf{H}^i_{\mathbf{x}} \Sigma_t (\mathbf{H}^i_{\mathbf{x}})^{\top} + R_o, & m^i_t = 1\\
\mathbf{H}^i_{\mathbf{x}} \Sigma_t (\mathbf{H}^i_{\mathbf{x}})^{\top} + R_u, &m^i_t = 0
 \end{array} \right., \nonumber
\end{align}
where $\mathbf{H}_{\mathbf{x}}^i = \nabla_{\mathbf{x}} h(\mu_t, m^i_t)$. The term $\mathbf{H}^i_{\mathbf{x}} \Sigma_t (\mathbf{H}^i_{\mathbf{x}})^{\top}$ is the projected localization uncertainty onto the measurement space, which directly couples the pose uncertainty $\Sigma_t$ into the map update likelihood.

To enhance robustness against early or noisy observations, we introduce a tempered weight $w^i_t$ defined using the accumulated number of beam interactions $N^i_{a,t}$ up to time $t$:
\begin{align} \label{eq::weight_map}
 w^i_t = \left\{ 
\begin{array}{ccc}
\frac{N_{max}-N^i_{a,t}}{N_{max}}, & m^i_t = 1\\
\frac{N^i_{a,t}}{N_{max}}, &m^i_t = 0
\end{array} \right.,
\end{align}
where $N_{max}$ is a predefined hyperparameter. This weighting scheme dampens the influence of the current measurement on the map posterior when prior evidence ($N^i_{a,t}$) is strong. In contrast to conventional approaches that directly interpret hit counts as occupancy probabilities~\cite{thrun2002probabilistic}, we employ them as tempering weights within the Bayesian update.

Combining the analytical marginalized likelihood with the weighting scheme and Bayes' theorem provides the final closed-form map update rule, \textsf{T-BayesMap}, which we summarize in the formal statement below.

\begin{lem}[Update for Map]
\label{lem::4_2}
\emph{Let Assumptions~\ref{assum::indep_cell} and \ref{assum::indep_obs} hold. Given the beam likelihood~\eqref{eq::likelihood_beam}, the posterior of localization in~\eqref{eq::update_localization}, and the weighting scheme~\eqref{eq::weight_map}, for all cells $i \in \mathcal{I}^k_t$ associated with the $k$-th beam, the map posterior $p(m^i_t|\mathbf{u}_{1:t},\mathbf{z}_{1:t})$ is computed as:
\begin{align} \label{eq::update_map}
 &p(m^i_t|\mathbf{u}_{1:t}, \mathbf{z}_{1:t}) \\
 &~=\frac{\mathcal{N}(z^k_t; \hat{\mu}_t, \hat{\Sigma}^i_t)^{w^i_t}p(m^i_t|\mathbf{u}_{1:t-1}, \mathbf{z}_{1:t-1})}{\sum_{m^i_t \in \{0, 1\}} \mathcal{N}(z^k_t; \hat{\mu}_t, \hat{\Sigma}^i_t)^{w^i_t}p(m^i_t|\mathbf{u}_{1:t-1}, \mathbf{z}_{1:t-1})}, \nonumber
\end{align}
with $\hat{\mu}_t = h(\mu_t, m^i_t)$ and $\hat{\Sigma}^i_t = \mathbf{H}^i_{\mathbf{x}} \Sigma_t (\mathbf{H}^i_{\mathbf{x}})^{\top} + R_{\{o, u\}}$, where $R_{\{o, u\}} = R_o$ for $m^i_t = 1$ and $R_u$ for $m^i_t = 0$. $\mathbf{H}_{\mathbf{x}}^i = \nabla_{\mathbf{x}} h(\mu_t, m^i_t)$.}\boxend
\end{lem}

\medskip
The mapping update rule in \textsf{T-BayesMap} exhibits the following key properties:
\begin{enumerate}
\item \emph{Mapping with coupled localization uncertainty:} The localization uncertainty from~\eqref{eq::update_localization} is incorporated into the map update through the $\mathbf{H}^i_{\mathbf{x}} \Sigma_t (\mathbf{H}^i_{\mathbf{x}})^{\top}$ term within $\hat{\Sigma}^i_t$. Consequently, the posterior of the map intrinsically reflects localization uncertainty, as illustrated in Fig.~\ref{fig:7_Map_loc_uncertainty}, thereby eliminating the need for heuristic parameter tuning and avoiding scale mismatches in the decision-making process.

\item \emph{Computation efficiency and scalability:} Under Assumptions~\ref{assum::indep_cell} and~\ref{assum::indep_obs}, the map update in~\eqref{eq::update_map} for all $i \in \mathcal{I}_t$ can be computed \emph{analytically} and \emph{in parallel}, resulting in $\mathcal{O}(1)$ complexity per cell, excluding the ray-tracing step. In practice, the updates of $\mathcal{I}^k_t$ for each beam $z^k_t$ are executed in parallel on parallel computing hardware.

\item \emph{Robustness:} The likelihood is tempered by the weight $w^i_t$, yielding $\mathcal{N}(z^k_t; \hat{\mu}_t, \hat{\Sigma}^i_t)^{w^i_t}$. By embedding the accumulated hit count $N^i_{a,t}$, this scheme moderates the effect of new measurements when prior evidence is strong, improving robustness against noise and outliers.

\item \emph{Extensibility:} The \textsf{T-BayesMap} update can act as a drop-in replacement for standard occupancy update rules across both graph-based frameworks and Gaussian assumed filtering approaches.

\end{enumerate}

\begin{figure}[t]
 \centering
\begin{minipage}[t]{0.485\textwidth}
 \centering
\includegraphics[width=\textwidth]{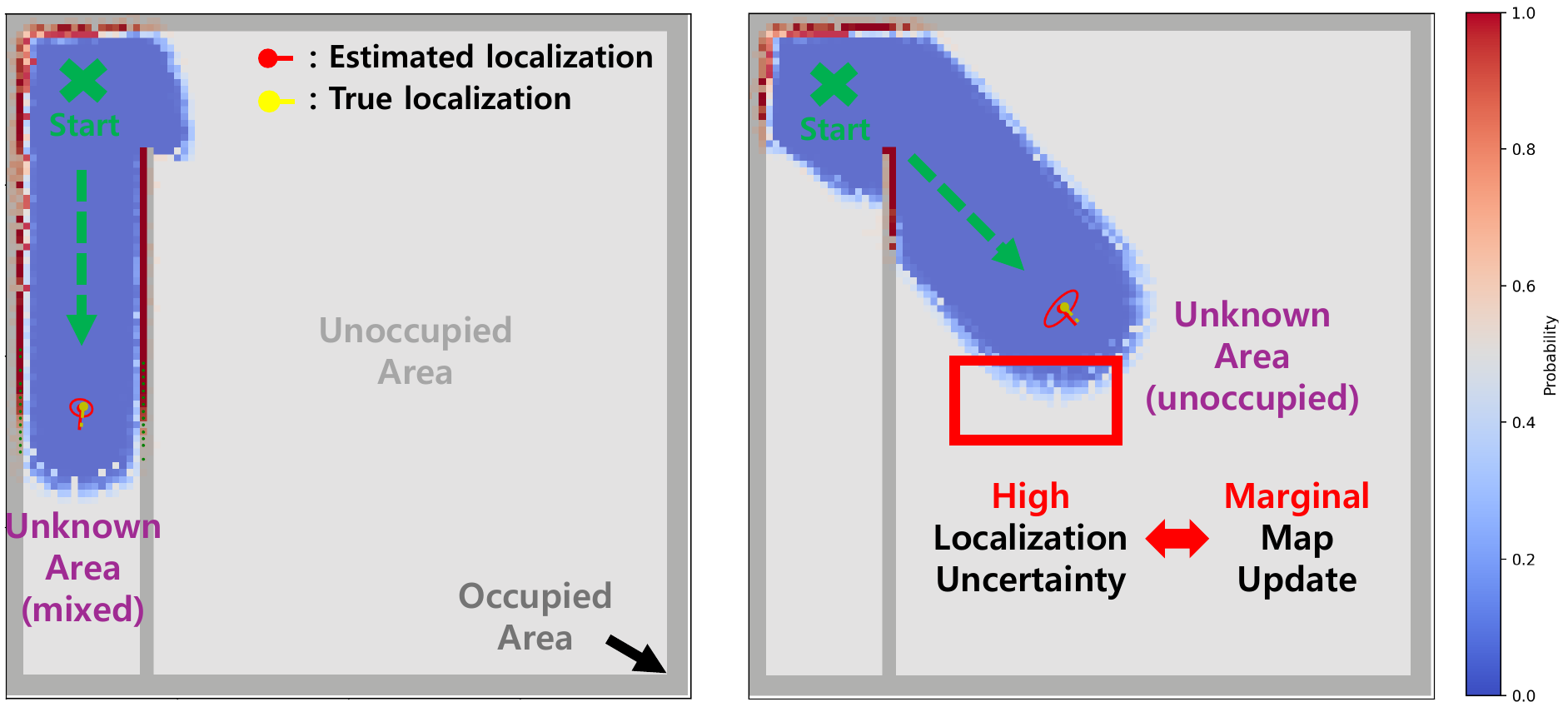}
 \end{minipage}
\caption{{\small Toy examples illustrating the effect of coupled uncertainty in the proposed localization–mapping framework. \textbf{(Left)} The robot explores an unknown region containing both occupied and unoccupied cells. Because the mixed environment and some occupied cells are insufficiently updated (as $R_u$ influences uncertainty), the localization uncertainty slightly increases, resulting in $\mathrm{tr}(\Sigma_t) = 3\times\mathrm{tr}(\Sigma_{\text{init}})$. \textbf{(Right)} The robot explores an unknown region consisting only of unoccupied cells. In this case, the localization uncertainty increases substantially, with $\mathrm{tr}(\Sigma_t) = 10\times\mathrm{tr}(\Sigma_{\text{init}})$, resulting in marginal map updates. These examples demonstrate the mutual influence between localization and mapping, which, in turn, affects the decision-making process discussed in Fig.~\ref{fig::9_Decision_making} and in Section~\ref{sec::Active_EXplore}.}}
 \label{fig::8_SLAM_framework}
\end{figure}

In summary, we propose a localization–mapping framework that embeds map uncertainty into localization, and vice versa, through marginalization—thereby \emph{decoupling} the localization and map states while preserving their \emph{coupled} uncertainty. Consequently, the framework achieves robust and scalable uncertainty-aware estimation and, \emph{beyond} estimation, facilitates \emph{uncertainty-aware} decision-making. Fig.~\ref{fig::8_SLAM_framework} illustrates how the coupled uncertainty evolves during robot exploration in illustrative examples.

In the following sections, we detail how this localization–mapping framework enables solving for $\mathbf{u}_{t+1:t'}^{\star}$ without direct computation of the joint entropy in the information gain.

\section{Coupled Uncertainty-aware Active Exploration}\label{sec::Active_EXplore}

In this section, we introduce our active exploration methodology built on the coupled localization–mapping framework. Because solving~\eqref{eq::cost_original} with~\eqref{eq::InfoGain} is mathematically intractable, the proposed \textsf{T-BayesMap} enables an indirect treatment of coupled uncertainty that removes the need for the classical separation (decoupling) scheme, yielding a principled balance between exploration and exploitation. Furthermore, it accommodates a broad class of generalized entropy measures, enabling adaptable exploration strategies. Finally, by incorporating BE into the formulation, the framework supports intuitive decision-making under coupled uncertainty.

\subsection{Generalized Entropy-based Active Exploration}

Generalized entropy measures, such as RE, enable diverse exploration strategies and, under specific conditions, reduce to SE (see Fig.~\ref{fig::Generalized_Entropy}). However, as noted in Remark~\ref{remark_1}, computing joint entropy with generalized entropy measures poses significant mathematical challenges. 

To overcome this, our \emph{key idea} is to compute only the map entropy—which inherently incorporates localization uncertainty of $\mathbf{x}_{t+1:t'}$—rather than the full joint entropy in~\eqref{eq::cost_original}. This reformulation indirectly overcomes the difficulties of joint entropy computation caused by the nonlinearity and non-separability of generalized entropy measures. Consequently, the optimization problem is reformulated in terms of coupled map entropy under localization uncertainty, allowing any generalized entropy measure to be applied:
\begin{align}
\label{eq::generalized_entropy_cost}
    &\mathbb{I}^{g}_{G}\bigr[\vect{u}_{t+1:t'}|\mathbf{z}_{t+1:t'}\bigr]  \\
    &~~~\triangleq \!\!\!\!\!\! \underbrace{\mathbb{H}\bigl(p(\mathbf{m}_t|\vect{l}_{1:t})\bigl)}_{\text{current coupled map entropy}} \!\!\!\!\!\!-~\underbrace{\mathbb{H}\bigl(p(\mathbf{m}_{t'}|\vect{l}_{1:t},\mathbf{u}_{t+1:t'},\mathbf{z}_{t+1:t'})\bigl)}_{\text{predicted coupled map entropy}}, \nonumber
\end{align}
where the entropy is computed over the map state, a binary random variable, and can therefore be obtained in \emph{closed form} without approximation\footnote{Unlike the binary random variable, the generalized entropy of Gaussian random variables does not admit a closed-form expression.}. In the set of predicted future measurements $\mathbf{z}_{t+1:t'}$ are typically estimated at the predicted poses $\mathbf{x}_{t+1:t'}$ using approximate ray-casting techniques combined with a plausible sensor model~\cite{burgard2005coordinated, stachniss2005information}.

Importantly, this formulation directly embeds localization uncertainty $\Sigma_t$ into each occupancy cell $m^i_t$, which removes the need for \emph{parameter tuning} and prevents \emph{scale mismatches} in the exploration–exploitation dilemma. Building on this, the next subsection introduces \emph{Behavioral entropy}, a recently proposed generalized entropy measure, into the decision-making~process.

\subsection{\textsf{B-ActiveSEAL}: Behavioral Active Scalable Exploration And Localization}

In robot exploration tasks, to provide adaptive action behavior, we draw upon the concept of \emph{Behavioral entropy}, $\mathbb{H}^B$, as introduced by~\cite{dhami2016foundations, suresh2024robotic}. BE leverages uncertainty models that adapt more flexibly to the complex and often unpredictable environments encountered in robotics. Unlike traditional entropy measures, BE provides a more expressive assessment of uncertainty by incorporating weighted probabilities. This enhanced measure enables robots to make intuitive and adaptive decisions during exploration.

The BE integrates the principles of Boltzmann-Gibbs-Shannon (BGS) entropy with Prelec's weighting function~\cite{prelec1998probability}. Consider a probability vector $p=(p_1,\dots,p_L) \in \mathcal{P}_L$. BE is defined as:
\begin{align}\label{eq::BE}
    \mathbb{H}^B(p_1,\dots,p_L) = -\sum\nolimits_{i=1}^L \Tilde{w}\bigl(p_i\bigl) \ln \Tilde{w}\bigl(p_i\bigl),
\end{align}
where the weighting function $\Tilde{w}\bigl(p_i\bigr) = e^{-\beta (-\ln{p_i})^{\alpha}}$ modifies the raw probabilities $p_i$ to better reflect perceived uncertainties. The parameter $\alpha>0$ controls the sensitivity to changes in probability; lower values of $\alpha$ make the entropy measure more responsive to rare events. Meanwhile, $\beta>0$ scales the overall weighting applied, amplifying the impact of perceived uncertainty. Notably, BE is a generalization of the BGS entropy and can recover SE by appropriately setting the parameters.

By applying BE from~\eqref{eq::BE} to~\eqref{eq::generalized_entropy_cost}, we obtain the \emph{Behavioral Information Gain} (BIG), defined as follows:
\begin{align}
\label{eq::behavioral_entropy_cost}
    &\mathbb{I}^B_{\alpha}\bigr[\vect{u}_{t+1:t'}|\mathbf{z}_{t+1:t'}\bigr]  \\
    &~~~\triangleq \mathbb{H}^B_{\alpha}\bigl(p(\mathbf{m}_t|\vect{l}_{1:t})\bigl) -\mathbb{H}^B_{\alpha}\bigl(p(\mathbf{m}_{t'}|\vect{l}_{1:t},\mathbf{u}_{t+1:t'},\mathbf{z}_{t+1:t'})\bigl), \nonumber
\end{align}
with $\beta = e^{(1-\alpha)\ln(\ln L)}$, where the number of outcomes is $L=2$ (occupied or unoccupied), as derived from Theorem 1 in~\cite{suresh2024robotic}. Therefore, $\alpha$ is the only free parameter and provides intuitive entropy results, as shown in Fig.~\ref{fig::Generalized_Entropy} and detailed in~\cite{suresh2024robotic}. Note that when $\alpha = 1$, the BE coincides with SE, and all the properties of SE apply to BE.

Algorithm~\ref{alg:ActBex} outlines the procedure for determining the optimal action sequence $\mathbf{u}_{t+1:t'}^{\star}$ under tightly coupled localization–mapping. In the prediction phase, candidate action sequences are evaluated while jointly updating localization and mapping. The algorithm leverages~\eqref{eq::pred_motion} and~\eqref{eq::update_localization} from Lemma~\ref{lem::4_1}, together with~\eqref{eq::update_map} from Lemma~\ref{lem::4_2}, and computes the BIG from~\eqref{eq::behavioral_entropy_cost}. The sequence with maximum BIG is selected as $\mathbf{u}_{t+1:t'}^{\star}$ and executed, with localization and mapping updated accordingly. The following results highlight the key properties of the BE-based active exploration approach.

\begin{figure}[t]
    \centering
    \begin{minipage}[t]{0.485\textwidth}
        \centering
        \includegraphics[width=\textwidth]{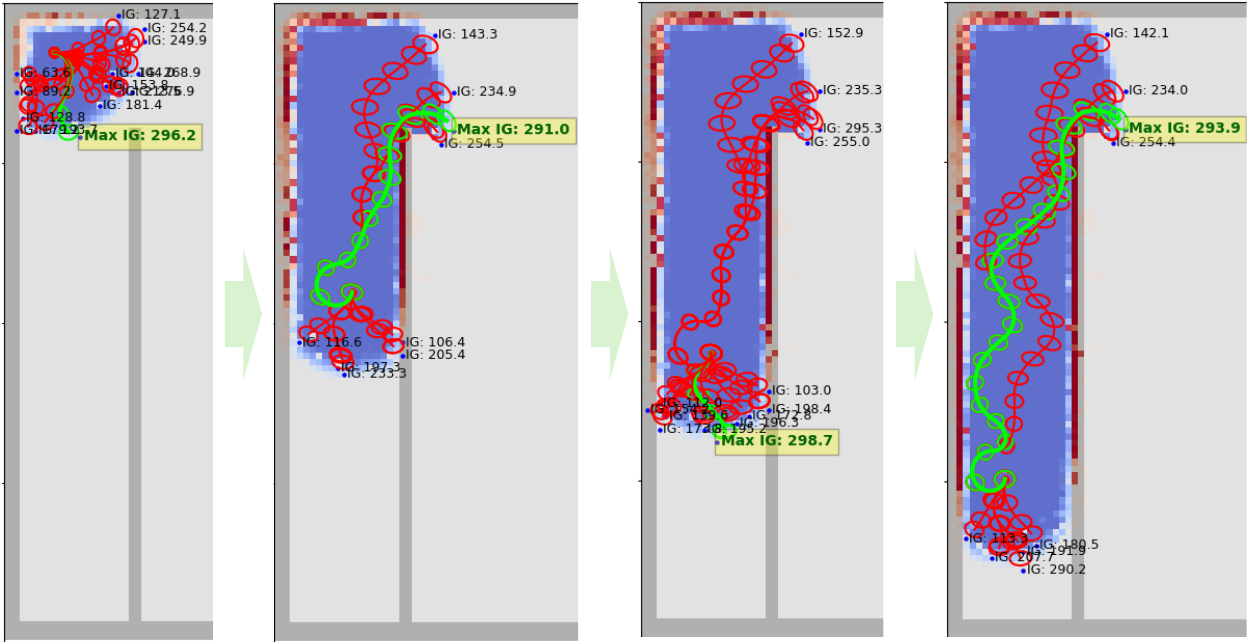}
    \end{minipage}
    \caption{{\small Toy example illustrating the proposed decision-making process. The robot moves downward, and at a specific step, candidate trajectories are generated according to $\mathbf{u}_{t+1:t'} \in \mathcal{U}_t$. The Behavioral Information Gain (BIG) is computed for each trajectory, with the trajectory yielding the maximum BIG shown in green and the others in red. The ellipsoids represent localization uncertainty along each trajectory. As shown in the second and fourth panels, continuously moving through unknown areas increases localization uncertainty and results in marginal map updates, as discussed in Fig.~\ref{fig::8_SLAM_framework}. Accordingly, the proposed decision-making process favors trajectories that mitigate these effects by transitioning from known regions into unknown areas.}}
    \label{fig::9_Decision_making}
\end{figure}

\begin{algorithm}[t]
\caption{\textsf{B-ActiveSEAL}}
\label{alg:ActBex}
\SetAlgoLined
\LinesNumbered
\DontPrintSemicolon
\SetKwInOut{Input}{Input}\SetKwInOut{Output}{Output}
\Input{$\mu_t$, $\Sigma_t$, $\mathbf{m}_t$, $\mathbf{u}_{t+1:t'} \in \mathcal{U}_t$, $\alpha$, $\gamma$, $N_{max}$}
\Output{$\mu_{t'}$, $\Sigma_{t'}$, $\mathbf{m}_{t'}$}
\BlankLine
\tcp{Prediction \& Decision-Making}
\For{$\mathbf{u}_{t+1:t'} \in \mathcal{U}_t$ in parallel}{
    \For{$t \leftarrow t+1$ \textbf{to} $t'$}{
    {propagate $\bar{\mu}_t, \bar{\Sigma}_t$ from~\eqref{eq::pred_motion} given $\mathbf{u}_{t}, \mu_{t-1}, \Sigma_{t-1}$} \\
    \tcp{At LiDAR update rate}
    {predict $\mathbf{z}_t$ given $\bar{\mu}_t$} \\
    {update $\mu_t, \Sigma_t$ from~\eqref{eq::update_localization} given $\mathbf{m}_{t-1}, \bar{\Sigma}^{-1}_t, \bar{\mu}_t, \mathbf{z}_t\!$}  \\
    \For{$\forall i \in \mathcal{I}_t$ in parallel}
    {\tcp{T-BayesMap}
    if $i \in \mathcal{I}^h_t$, update $\bar{N}^i_{a,t}$ given $\mathbf{z}_t$\\
    update $m^i_t$ from~\eqref{eq::update_map} given $\mu_t, \Sigma_t, N_{max}, \bar{N}^i_{a,t}\!\!\!\!\!$}
    }
    {compute $\mathbb{I}^{B}_{\alpha}\bigr[\vect{u}_{t+1:t'}|\mathbf{z}_{t+1:t'}\bigr]$ from~\eqref{eq::behavioral_entropy_cost} \\ 
    given $\mathbf{m}_t, \mathbf{m}_{t'},\alpha$}}
    $\mathbf{u}_{t+1:t'}^{\star} = \arg\max_{\mathbf{u}_{t+1:t'} \in \mathcal{U}_t} \mathbb{I}^{B}_{\alpha}\bigr[\vect{u}_{t+1:t'}|\mathbf{Z}_{t+1:t'}\bigr]$\;
    \tcp{Executing \& Updating}
    \For{$t \leftarrow t+1$ \textbf{to} $t'$}{
    {compute $\bar{\mathbf{u}}^{\star}_{t} \leftarrow$ \texttt{Planner}($\mathbf{u}_{t+1:t'}^{\star},\mathbf{m}_t,\mu_{t-1}, \Sigma_{t-1}$)} \\
    {propagate $\bar{\mu}_t, \bar{\Sigma}_t$ from~\eqref{eq::pred_motion} given $\bar{\mathbf{u}}^{\star}_{t}, \mu_{t-1}, \Sigma_{t-1}$} \\
    \tcp{Executed only if $\mathbf{z}_t$ is available}
    {update $\mu_t, \Sigma_t$ from~\eqref{eq::update_localization} given $\mathbf{m}_{t-1}, \bar{\Sigma}^{-1}_t, \bar{\mu}_t, \mathbf{z}_t\!$} \\
    \For{ $\forall i \in \mathcal{I}_t$ in parallel}
    {\tcp{T-BayesMap}
    if $i \in \mathcal{I}^h_t$, update $N^i_{a,t}$ given $\mathbf{z}_t$ \\
    update $m^i_t$ from~\eqref{eq::update_map} given $\mu_t, \Sigma_t, N_{max}, N^i_{a,t}$}
    }
\end{algorithm}

First, note that the robot’s localization uncertainty is inherently embedded in the map-state estimate. The following result shows that lower localization uncertainty leads to a larger reduction in the predicted BE, which corresponds to a higher BIG. Fig.~\ref{fig::9_Decision_making} further illustrates how map entropy, under coupled localization uncertainty, directly shapes the decision-making process and, consequently, the robot’s exploration behavior.

\smallskip
\begin{thm}[BIG Increases with Reduced Localization Uncertainty] \label{thm::5_1} 
\emph{Let Assumption~\ref{assum::indep_cell} and~\ref{assum::indep_obs} hold. Consider two scenarios in which the collected measurements $\mathbf{z}_{t+1}$, and the current map are the same, i.e., $\mathbf{m}_t=\mathbf{\bar{m}}_t$. Let $\Sigma_{t+1}$ and $\bar{\Sigma}_{t+1}$ denote the covariance of the robot's localization uncertainty under these two different scenarios. For any $\alpha>0$, if $\Sigma_{t+1} \leq \bar{\Sigma}_{t+1}$, then the predicted coupled map BE satisfies
\begin{align}\label{eq::theorem_BE}
    \!\!\!\mathbb{H}^B_{\alpha}\bigl(p(\mathbf{m}_{t+1}| \mathbf{u}_{t+1},\mathbf{z}_{t+1})\!\bigl)~\leq~\mathbb{H}^B_{\alpha}\bigl(p(\mathbf{\bar{m}}_{t+1}|\mathbf{u}_{t+1},\mathbf{z}_{t+1})\!\bigl),\!\!\!
\end{align}
and consequently,
\begin{align}\label{eq::theorem_BIG}
    \mathbb{I}^B_{\alpha}\bigr[\vect{u}_{t+1}|\mathbf{z}_{t+1}\bigr] \geq \bar{\mathbb{I}}^B_{\alpha}\bigr[\vect{u}_{t+1}|\mathbf{z}_{t+1}\bigr],
\end{align}
where $\bar{\mathbb{I}}^B_{\alpha}$ denotes the corresponding BIG evaluated from $\mathbf{\bar{m}}_{t+1}$ under $\bar{\Sigma}_{t+1}$.
}\end{thm}
\begin{proof}
    See Appendix~\ref{sec::appen_C}
\end{proof}

\begin{figure}[t]
    \centering
    \begin{minipage}[t]{0.41\textwidth}
        \centering
        \includegraphics[width=\textwidth]{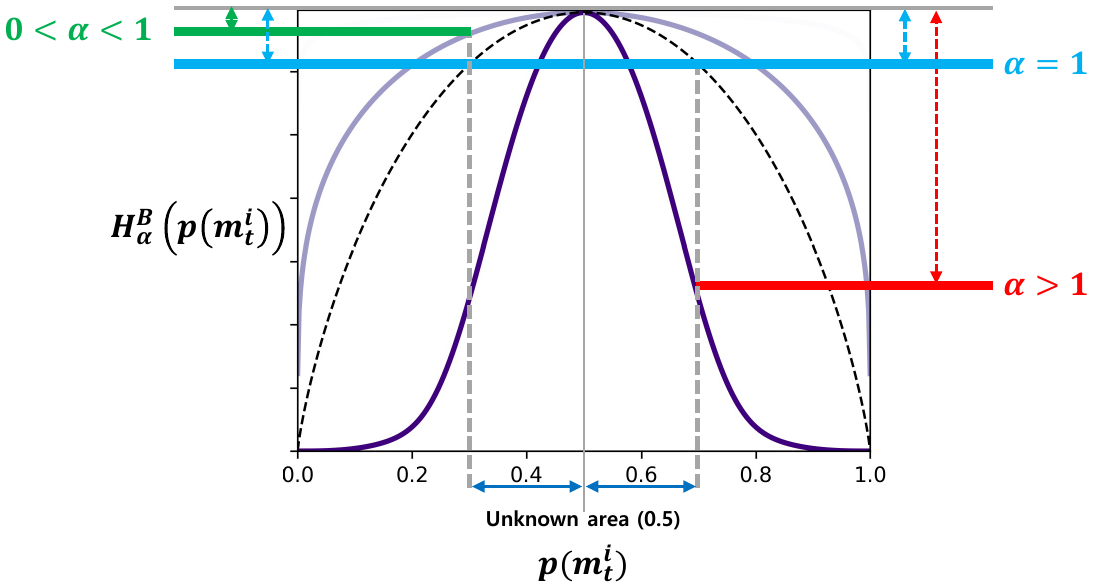}
    \end{minipage}
    \caption{{\small Sensitivity of Behavioral Entropy $H^B_{\alpha}(p(m^i_t))$ to localization uncertainty as a function of $\alpha$. When the robot explores an unknown area ($p(m^i_t)=0.5$), reductions in $H^B_{\alpha}(p(m^i_t))$ correspond to changes in $p(m^i_t)$, which in turn reflect changes in localization uncertainty. For $\alpha > 1$, BE becomes more sensitive to these changes (steeper slope), producing a larger increase in BIG. In contrast, for $0 < \alpha < 1$, BE becomes less sensitive to localization-uncertainty changes (flatter slope). Thus, $\alpha$ serves as a tunable parameter for controlling the exploration–exploitation trade-off.}}
    \label{fig::8_Theorem5_2}
\end{figure}

\medskip
Recall that one of the advantages of using BE is the adjustable parameter $\alpha$, which can be tuned to influence exploration behavior. While Theorem~\ref{thm::5_1} shows that BIG increases as localization uncertainty decreases, $\alpha$ controls the sensitivity of BE (and thus BIG) to localization uncertainty. As formalized in Theorem~\ref{thm::5_2} and shown in Fig.~\ref{fig::8_Theorem5_2}, when the robot explores an unknown area, choosing $\alpha > 1$ produces a larger increase in BIG than SE $(\alpha = 1)$, and the resulting BIG becomes more sensitive to variations in localization uncertainty. This makes the action policy more strongly favor exploration when localization uncertainty is low. In contrast, when $0 < \alpha < 1$, the policy becomes more conservative: exploration is reduced, and the policy responds less to decreases in localization uncertainty (i.e., it is less sensitive to exploitation) compared to the SE-guided case. These properties naturally provide a tunable balance between exploration and exploitation. In practice, $\alpha$ can be selected based on mission objectives, environmental structure, and sensor characteristics (e.g., maximum LiDAR range).

\smallskip
\begin{thm}[Sensitivity of BE to Localization Uncertainty as a Function of $\alpha$]
\label{thm::5_2}
\emph{ 
Let $\Sigma_t$ denote the covariance of the robot's pose uncertainty. When the robot explores unknown areas, the sensitivity of BE to localization uncertainty satisfies the following properties:
}
\begin{enumerate}
    \item For $\alpha > 1$, BE exhibits greater sensitivity to localization uncertainty compared to Shannon entropy ($\alpha = 1$):
    \begin{equation}
        \left| \frac{\partial \mathbb{H}^B_{\alpha > 1}}{\partial \Sigma_t} \right| > \left| \frac{\partial \mathbb{H}^B_{\alpha = 1}}{\partial \Sigma_t} \right|, \quad \text{for} \quad p(m^i_t) \neq 0.5. \nonumber
    \end{equation}
    
    \item For $0 < \alpha < 1$, BE exhibits lower sensitivity to localization uncertainty than Shannon entropy:
    \begin{equation}
        \left| \frac{\partial \mathbb{H}^B_{0<\alpha < 1}}{\partial \Sigma_t} \right| < \left| \frac{\partial \mathbb{H}^B_{\alpha = 1}}{\partial \Sigma_t} \right|, \quad \text{for} \quad p(m^i_t) \neq 0.5. \nonumber
    \end{equation}
\end{enumerate}
\end{thm}
\begin{proof} 
    See Appendix~\ref{sec::appen_D}
\end{proof}

\subsection{Coupled Uncertainty-aware Planning}

The proposed coupled uncertainty-aware estimation framework produces more informative map uncertainty, enabling global planners (e.g., $A^{\star}$, $RRT^{\star}$) to generate more accurate and informative planning results.
Moreover, the Gaussian-based localization, which incorporates both sensor and map uncertainties, is compatible with existing optimization- or sampling-based local planners (e.g., iLQR, MPPI).
While these planning results can further enhance exploration performance, a detailed analysis of such integration is beyond the scope of this paper and will be explored in future work. 

\begin{figure}[!t]
    \centering
    \begin{minipage}[t]{0.232\textwidth}
        \centering
        \includegraphics[width=\textwidth]{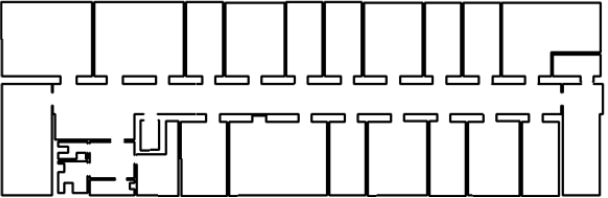}
        {{\scriptsize (a) Small map (55m × 160m)}}
    \end{minipage}
    \par\vspace*{2pt}
    \begin{minipage}[t]{0.232\textwidth}
        \centering
        \includegraphics[width=\textwidth]{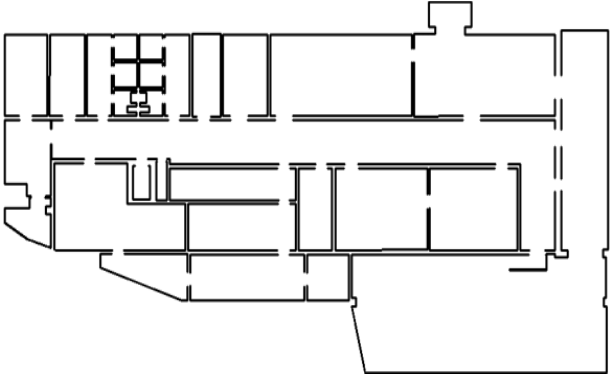}
        {{\scriptsize (b) Medium map (90m × 170m)}}
    \end{minipage}
    \begin{minipage}[t]{0.232\textwidth}
        \centering
        \includegraphics[width=\textwidth]{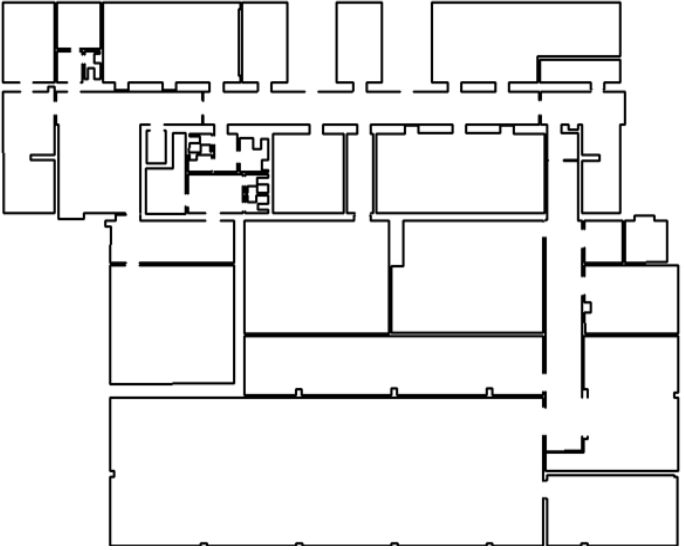}
        {{\scriptsize (c) Large map (130m × 170m)}}
    \end{minipage}
    \caption{{\small The three indoor maps used for evaluation. Small, medium, and large environments are shown, each with distinct topologies.}}
    \label{fig:10_open_source_map}
\end{figure}

\section{Experiments}\label{sec::experiments}

In this section, we evaluate both the quantitative and qualitative performance of the proposed active exploration framework.
We first define the metrics used for quantitative evaluation.
Next, we conduct two ablation studies on open-source maps to analyze how \emph{coupled} uncertainty in both the localization–mapping framework and the decision-making process affects exploration. Finally, we examine the $\alpha$-adjustable uncertainty-aware behavior of the proposed framework in real-world scenarios using ROS–Unity 3D simulations. Although additional terms (e.g., Euclidean distance) could be incorporated into the decision-making utility to further accelerate exploration, this study focuses exclusively on information-theoretic measures.

\subsection{Metrics}
We evaluate performance using the following metrics:
\begin{enumerate}
    \item \textbf{Localization error} is quantified using the root mean squared error (RMSE) and mean absolute error (MAE) of the robot’s translational error—arising from both orientation and map error—averaged over all time steps along the trajectory:
    \begin{align}
    \text{RMSE} &= \sqrt{\frac{1}{T} \sum\nolimits_{t=1}^{T} 
    \left\| \hat{\mathbf{x}}_t - \mathbf{x}^g_t \right\|^2 }, \nonumber \\
    \text{MAE} &= \frac{1}{T} \sum\nolimits_{t=1}^{T} 
    \left\| \hat{\mathbf{x}}_t - \mathbf{x}^g_t \right\|, \nonumber
    \end{align}
    where $\hat{\mathbf{x}}_t$ and $\mathbf{x}^g_t$ denote the estimated and ground-truth localization at time step $t$, respectively. $T$ is the total number of time steps, and $\parallel\cdot\parallel$ is the Euclidean norm.
    \item \textbf{Map error} is quantified using the RMSE between the estimated occupancy probabilities and the ground-truth map. For each cell $i$, the squared error is $(1 - p(m^i_t))^2$ for occupied cells ($m^i_t = 1$) and $(p(m^i_t))^2$ for unoccupied cells ($m^i_t = 0$). The RMSE is obtained by averaging over all occupied ($N_{\text{occ}}$) and unoccupied ($N_{\text{unocc}}$) cells except for unknown cells ($p(m^i_t)=0.5$), taking the square root, and multiplying by 100:
    \begin{equation}
         \Biggl(\!\sqrt{\frac{1}{N_{occ}}\!\!\sum_i^{N_{occ}}(1 \!-\! p(m^i_t))^2 + \frac{1}{N_{unocc}}\!\!\!\sum_i^{N_{unocc}}\! p(m^i_t)^2} \Biggl) \!\times 100. \nonumber
    \end{equation} 
    \item \textbf{Map uncertainty} is quantified as the average per-step reduction in SE over the map:
    \begin{equation}
        \frac{1}{T}\sum\nolimits_{t=1}^{T} \Big( \mathbb{H}^s\bigl(p(\mathbf{m}_{t-\Delta t})\bigl) - \mathbb{H}^s\bigl(p(\mathbf{m}_t)\bigl) \Big), \nonumber
    \end{equation}
    where $\mathbb{H}^s(\cdot)$ denotes SE.
\end{enumerate}
We further assess the reduction of map uncertainty (i.e., exploration progress) by analyzing the temporal evolution of $\mathbb{H}^s\bigl(p(\mathbf{m}_t)\bigl)$.

\begin{figure*}[!t]
    \centering
    \begin{minipage}[t]{0.25\textwidth}
        \centering
        \includegraphics[width=\textwidth]{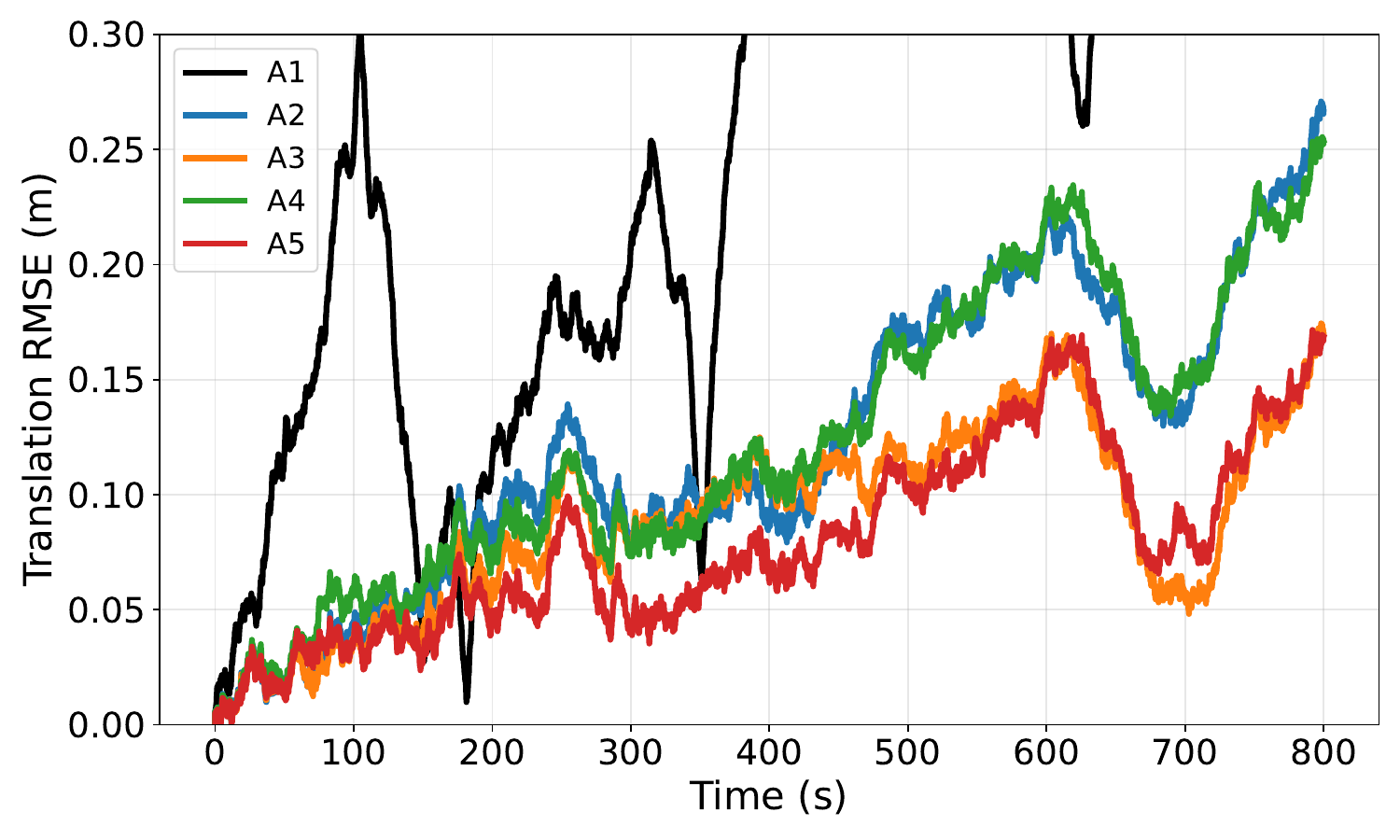}
    \end{minipage}
    \begin{minipage}[t]{0.25\textwidth}
        \centering
        \includegraphics[width=\textwidth]{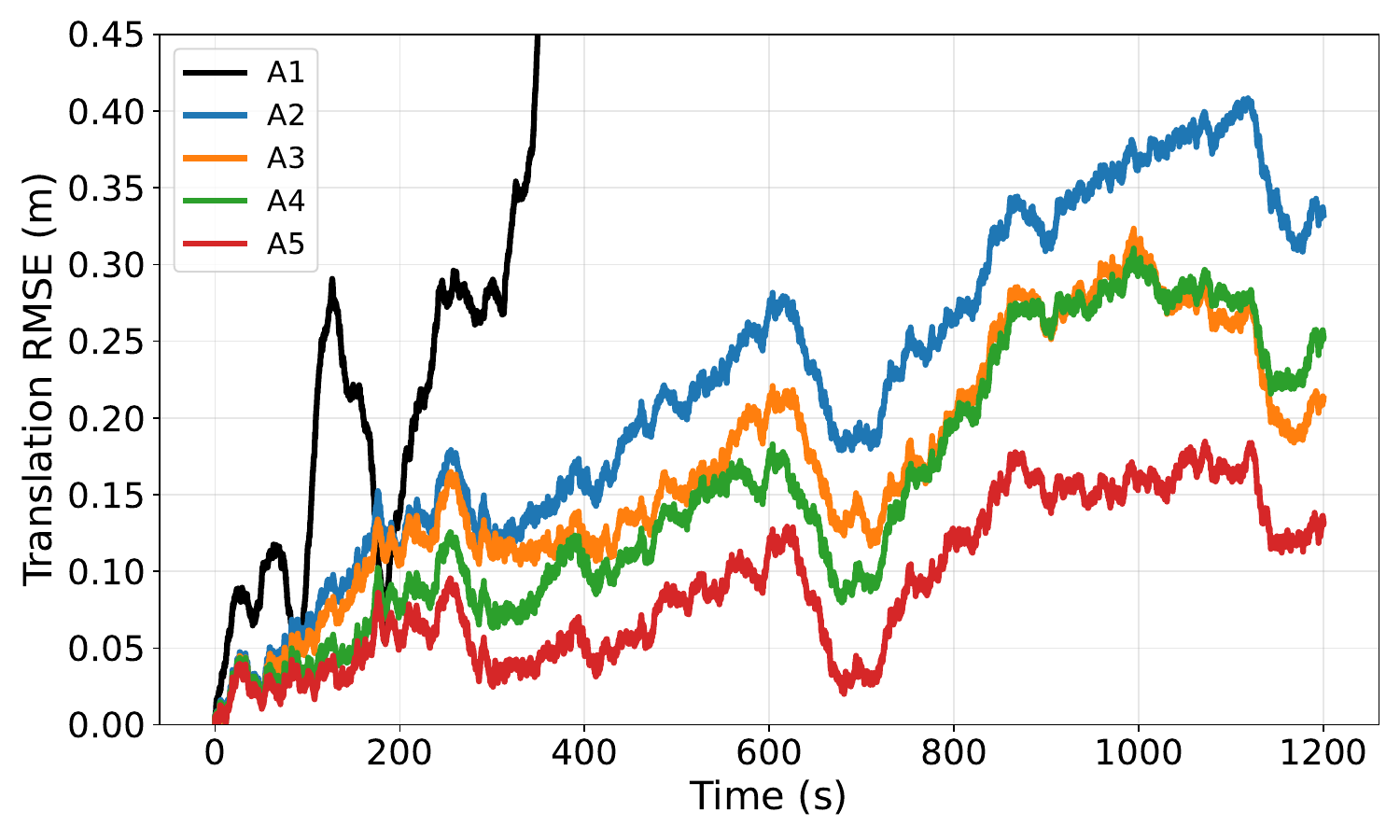}
    \end{minipage}
    \begin{minipage}[t]{0.25\textwidth}
        \centering
        \includegraphics[width=\textwidth]{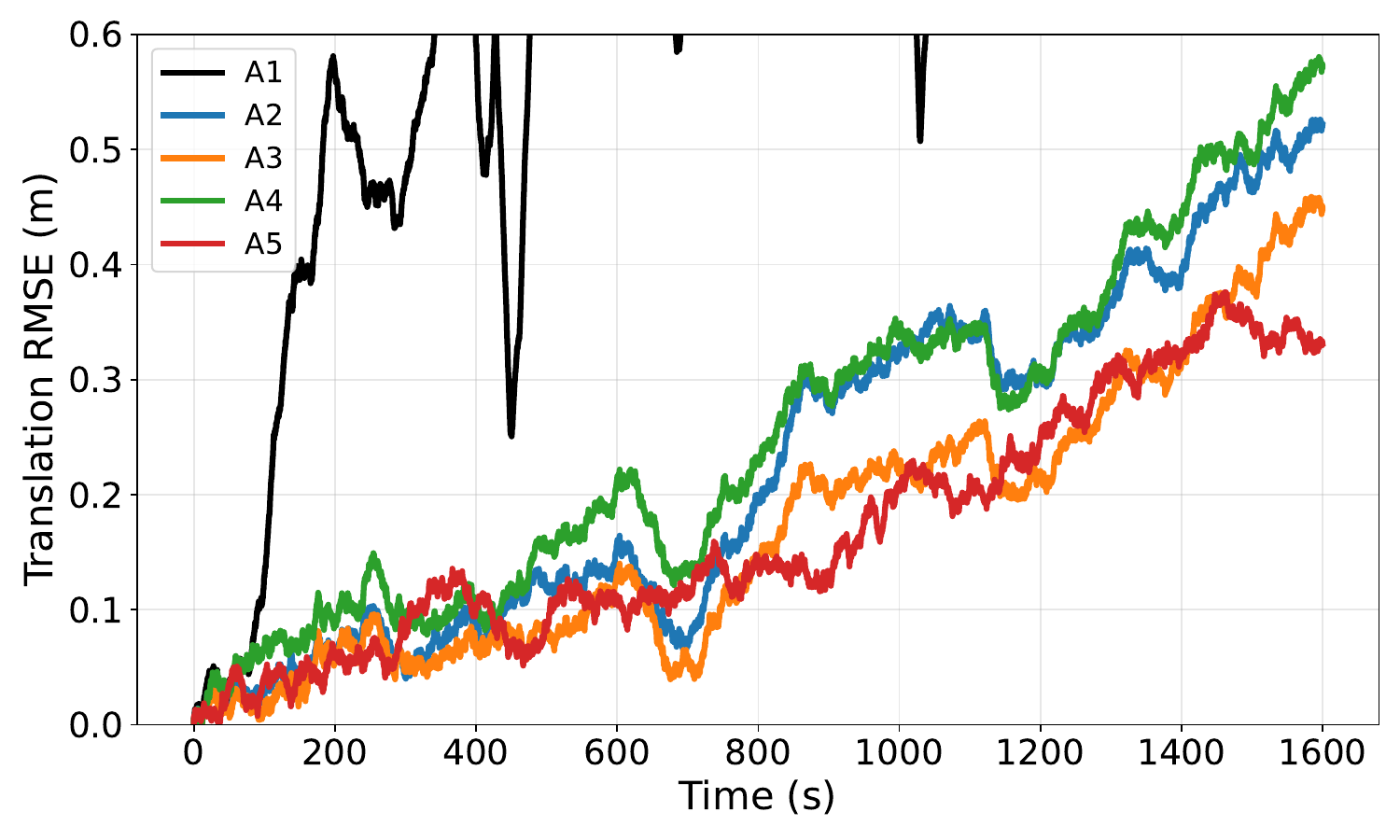}
    \end{minipage}
    \begin{minipage}[t]{0.25\textwidth}
        \centering
        \includegraphics[width=\textwidth]{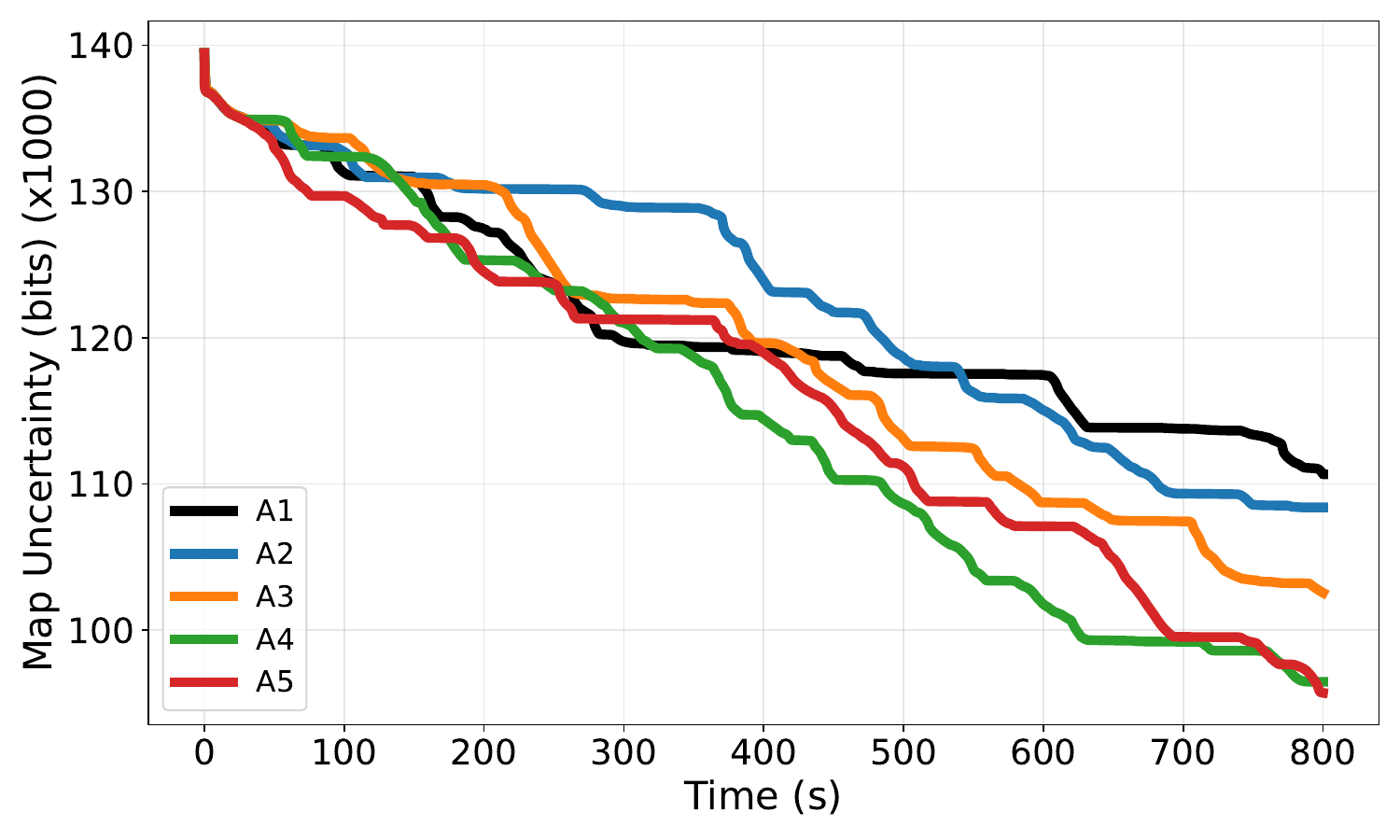}
    \end{minipage}
    \begin{minipage}[t]{0.25\textwidth}
        \centering
        \includegraphics[width=\textwidth]{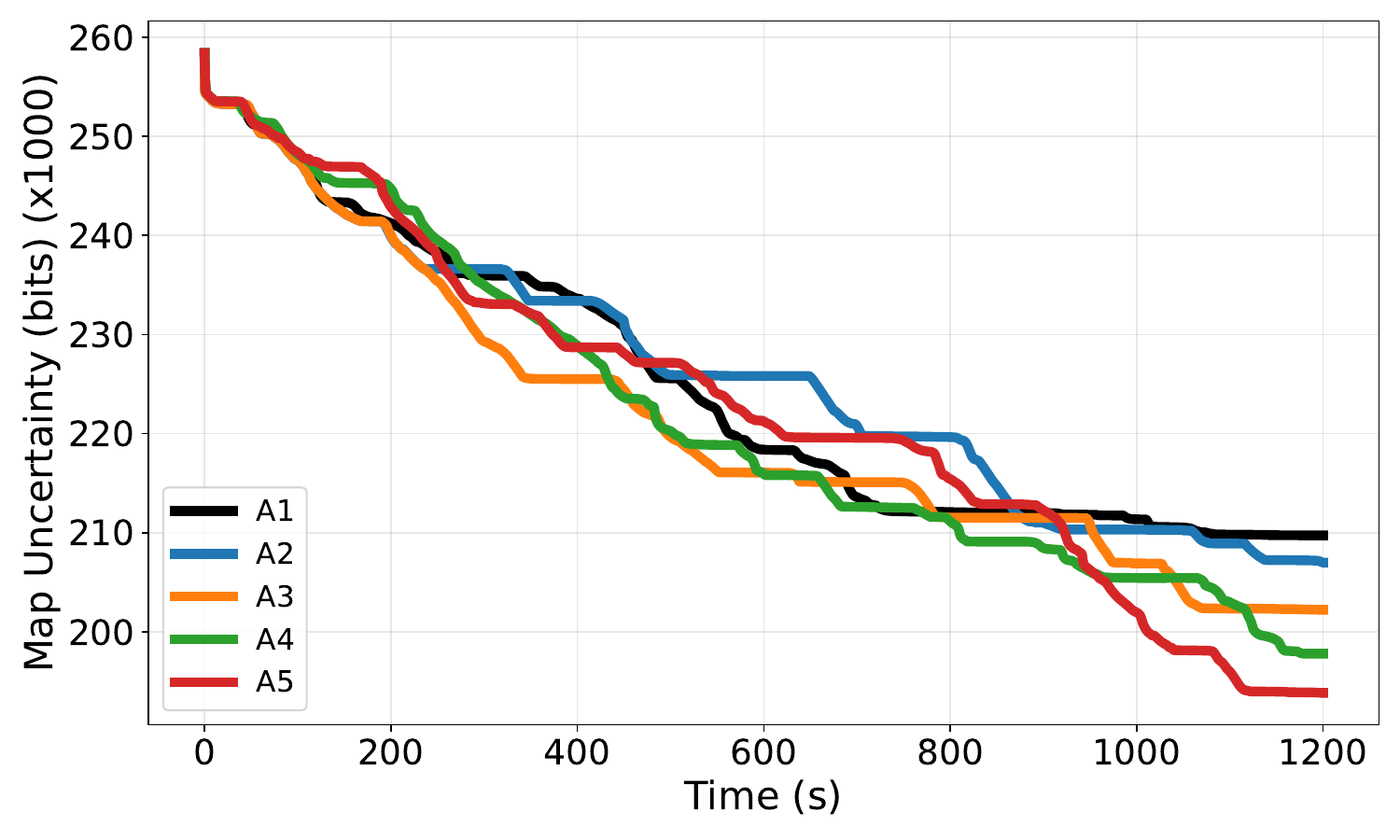}
    \end{minipage}
    \begin{minipage}[t]{0.25\textwidth}
        \centering
        \includegraphics[width=\textwidth]{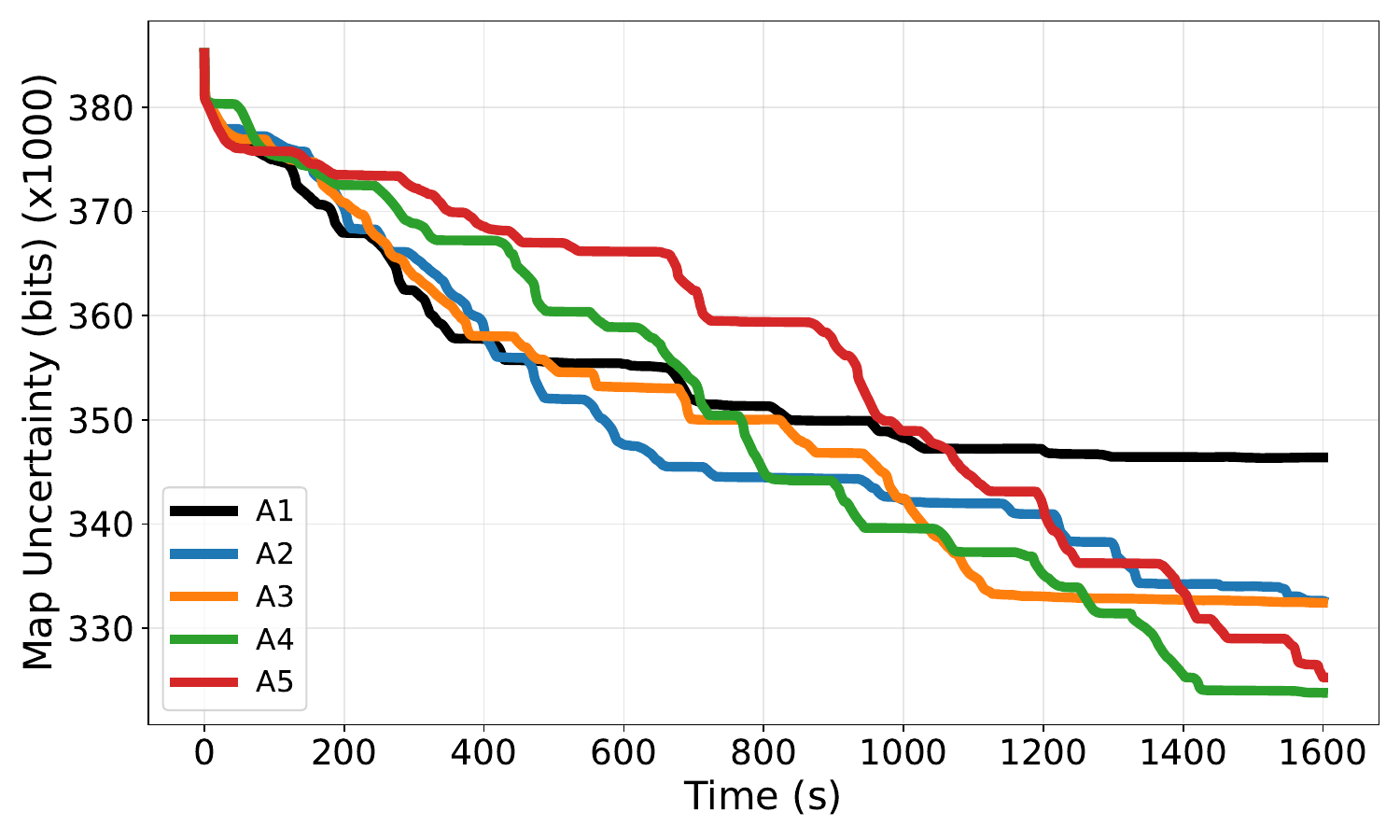}
    \end{minipage}
    \begin{minipage}[t]{0.251\textwidth}
        \centering
        \includegraphics[width=\textwidth]{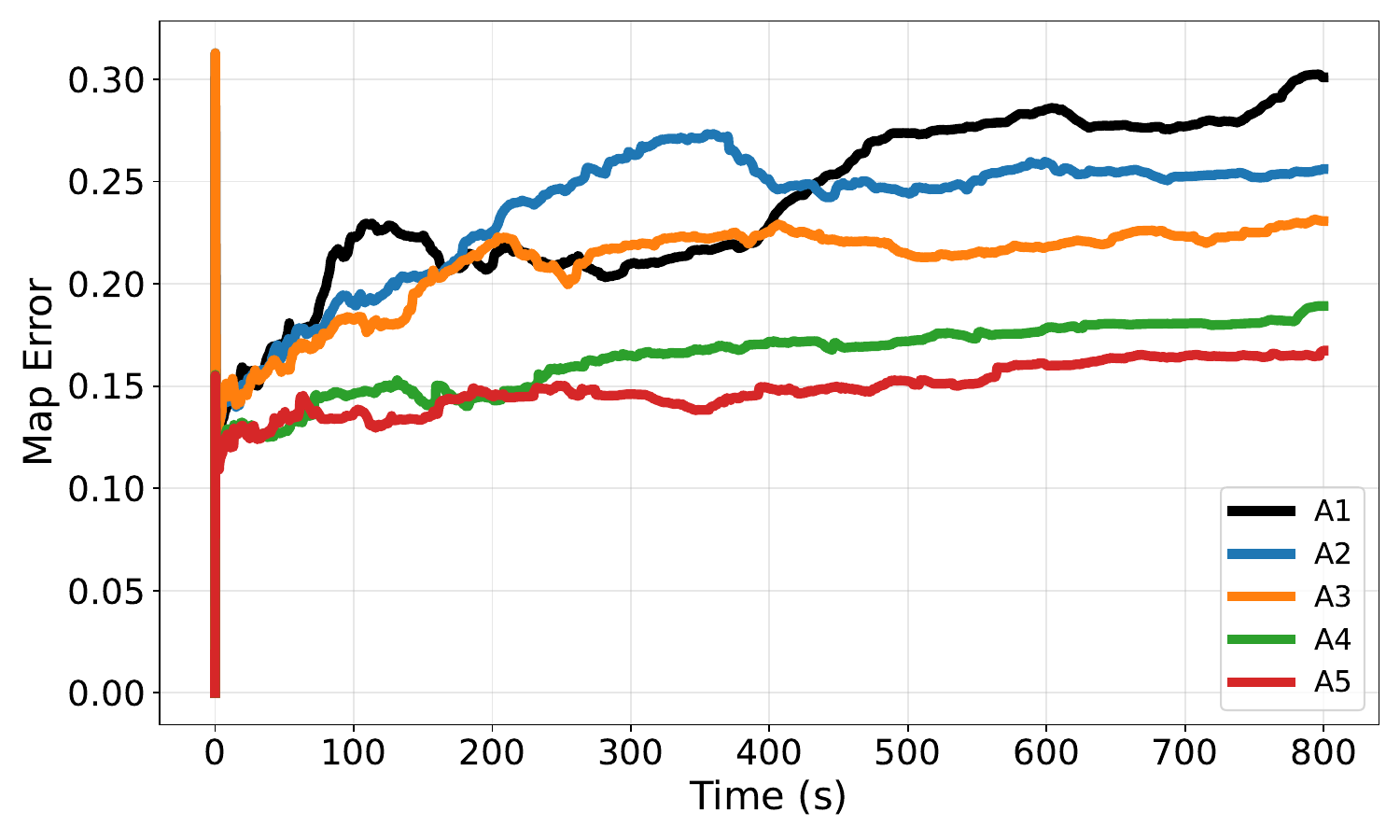}
        {{\scriptsize (a) Small Map (55m × 160m, 800s)}}
    \end{minipage}
    \begin{minipage}[t]{0.251\textwidth}
        \centering
        \includegraphics[width=\textwidth]{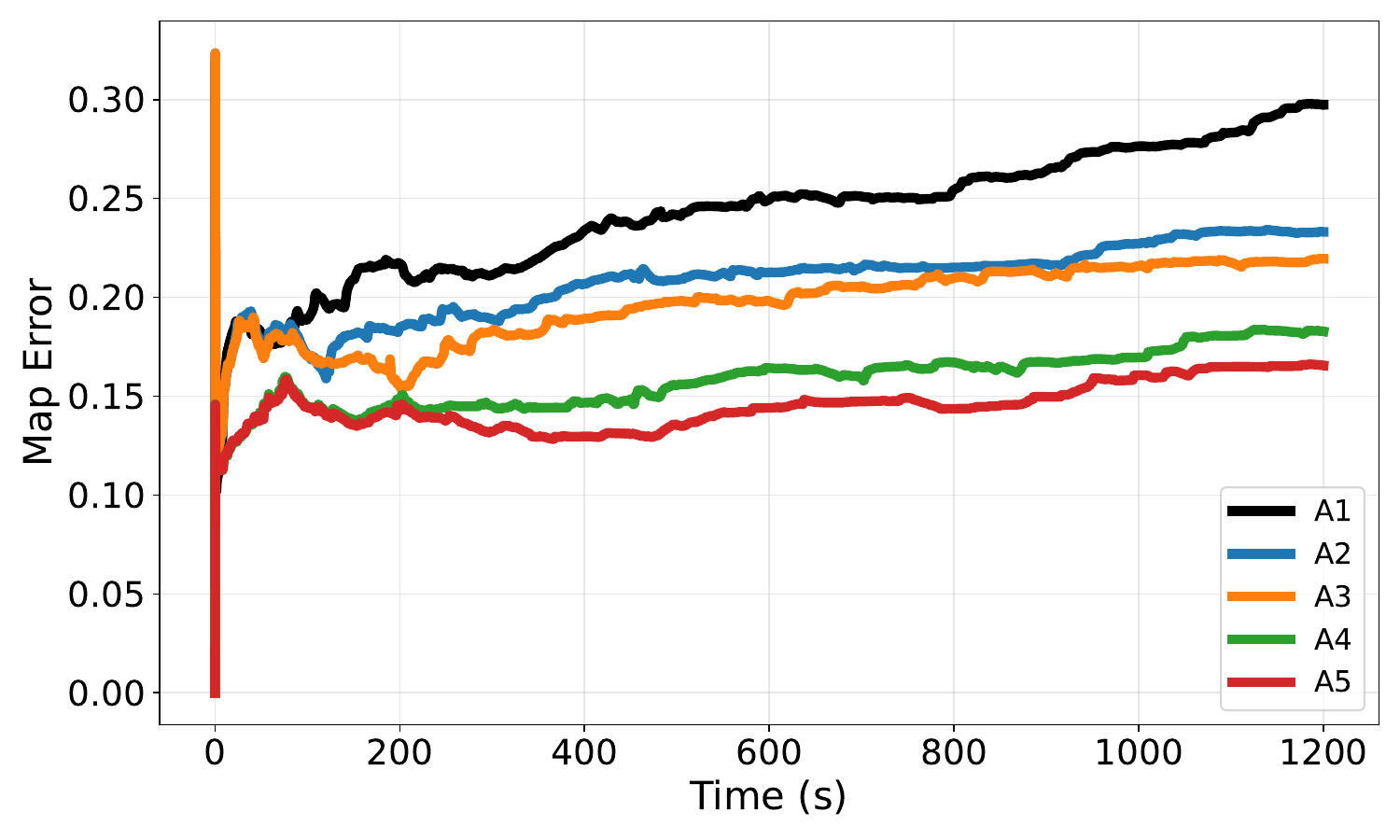}
        {{\scriptsize (b) Medium Map (90m × 170m, 1200s)}}
    \end{minipage}
    \begin{minipage}[t]{0.251\textwidth}
        \centering
        \includegraphics[width=\textwidth]{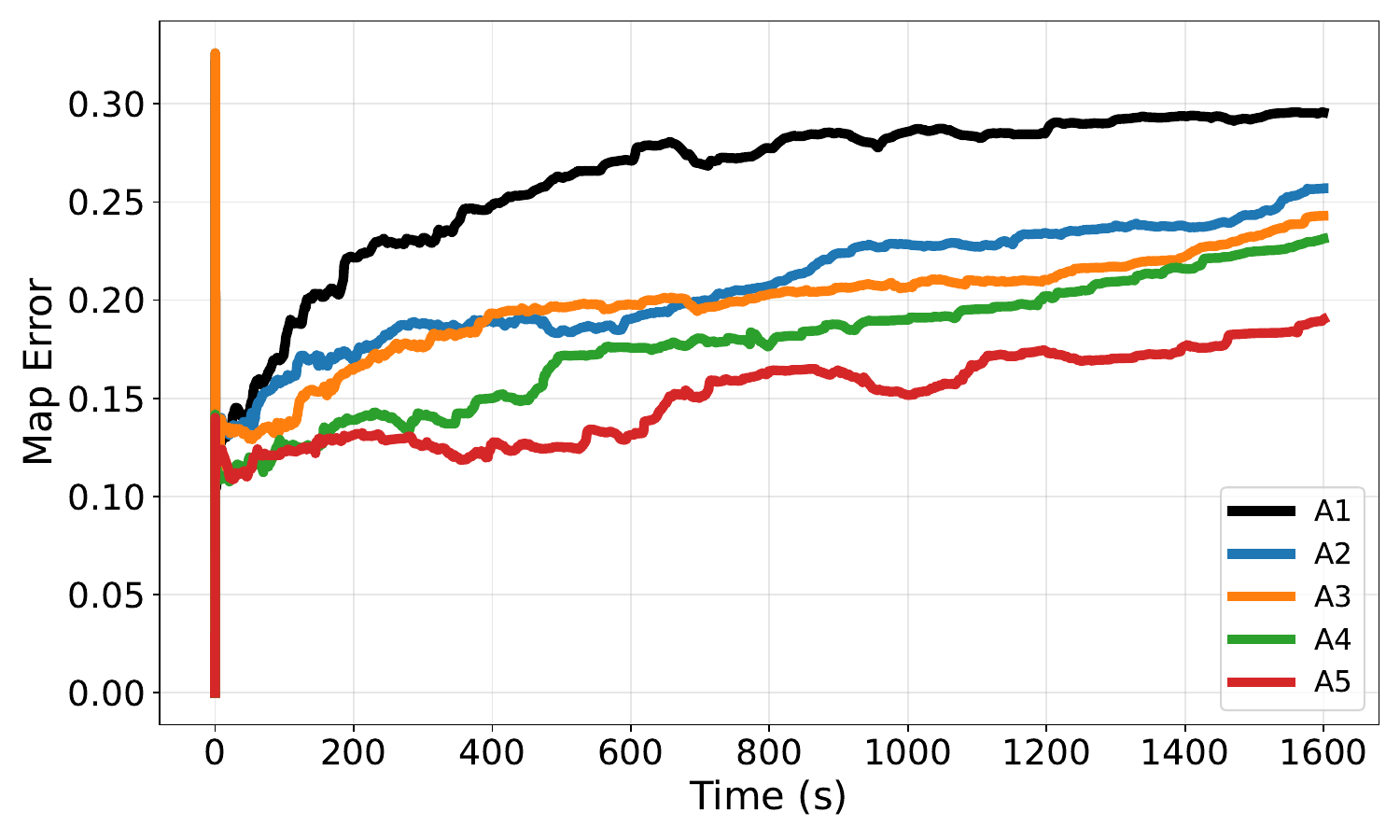}
        {{\scriptsize (c) Large Map (130m × 170m, 1600s)}}
    \end{minipage}
    \caption{{\small Comparisons of the mean results of the localization–mapping framework over time using 10 independent random initializations from a fixed initial location. A1 consistently exhibits poor performance across all map sizes. A4 and A5, which employ \textsf{T-BayesMap}, achieve lower map error and map uncertainty than A2 and A3, which use decoupled mapping with the inverse model. Between A3 and A4, coupled localization in A3 yields lower translational error than A4. A5, which integrates both coupled localization and \textsf{T-BayesMap}, achieves the lowest translational and map errors while maintaining effective exploration (low map uncertainty).}}
    \label{fig:11_result}
\end{figure*}

\begin{table*}[t]
\begin{center}
\caption{Monte Carlo Ablation Study of the Localization–Mapping Framework with 10 Distinct Initial Locations}
\label{Table_Ex1}
\renewcommand{\arraystretch}{1.25}
\large
\resizebox{0.98\textwidth}{!}{
    \begin{tabular}{c|c c c|c c c|c c c} 
    \hline
    & \multicolumn{9}{c}{\Large Map Size} \\
    \cline{2-10}
    & \multicolumn{3}{c}{\Large Small Map (55m × 160m, 800s)} & \multicolumn{3}{c}{\Large Medium Map (90m × 170m, 1200s)} & \multicolumn{3}{c}{\Large Large Map (130m × 170m, 1600s)} \\
    \cline{2-10}
    {\LARGE Methods} & Translation Error $\downarrow$ & Map Error $\downarrow$ & Map Uncertainty $\uparrow$ & Translation Error $\downarrow$ & Map Error $\downarrow$ & Map Uncertainty $\uparrow$ & Translation Error $\downarrow$ & Map Error $\downarrow$ & Map Uncertainty $\uparrow$ \\[-2pt]
    & RMSE ($m$), MAE ($m$) & RMSE & (bits/step) & RMSE ($m$), MAE ($m$) & RMSE & (bits/step) & RMSE ($m$), MAE ($m$) & RMSE & (bits/step) \\[3pt]
    \hline 
    \textbf{\Large A1} & {\Large $0.8846$},~~{\Large $0.5903$} & {\Large $0.3009$} & {\Large $0.7203$} & {\Large $1.3537$},~~{\Large $1.1016$} & {\Large $0.2974$}& {\Large $0.7102$} & {\Large $1.8692$},~~{\Large $1.2415$} & {\Large $0.2956$} & {\Large $0.4854$} \\
    \textbf{\Large A2} & {\Large$0.1379$,~~$0.1220$} & {\Large $0.2562$} & {\Large $0.7767$} & {\Large $0.2484$},~~{\Large $0.1843$} & {\Large $0.2331$} & {\Large $0.8113$} & {\Large $0.2709$},~~{\Large $0.2239$} & {\Large $0.2570$} & {\Large $0.6573$} \\
    \textbf{\Large A3} & {\Large \textcolor{blue}{\boldmath$0.0974$}},~~{\Large \textcolor{blue}{\boldmath$0.0758$}} & {\Large $0.2307$} & {\Large $0.9227$} & {\Large \textcolor{blue}{\boldmath$0.1771$}},~~{\Large \textcolor{blue}{\boldmath$0.1391$}} & {\Large $0.2195$} & {\Large $0.8572$} & {\Large \textcolor{blue}{\boldmath$0.2094$}},~~{\Large \textcolor{blue}{\boldmath$0.1641$}} & {\Large $0.2430$} & {\Large $0.6598$}\\
    \textbf{\Large A4} & {\Large $0.1381$},~~{\Large $0.1040$} & {\Large \textcolor{blue}{\boldmath$0.1892$}} & {\Large \textcolor{blue}{\boldmath$1.0749$}} & {\Large $0.1871$},~~{\Large $0.1443$} & {\Large \textcolor{blue}{\boldmath$0.1828$}} & {\Large \textcolor{red}{\boldmath$1.0760$}} & {\Large $0.2914$},~~{\Large $0.2381$} & {\Large \textcolor{blue}{\boldmath$0.2313$}} & {\Large \textcolor{blue}{\boldmath$0.7493$}}\\
    \textbf{\Large A5} & {\Large \textcolor{red}{\boldmath$0.0742$}},~~{\Large \textcolor{red}{\boldmath$0.0588$}} & {\Large \textcolor{red}{\boldmath$0.1648$}} & {\Large \textcolor{red}{\boldmath$1.0937$}} & {\Large \textcolor{red}{\boldmath$0.1032$}},~~{\Large \textcolor{red}{\boldmath$0.0844$}} & {\Large \textcolor{red}{\boldmath$0.1655$}} & {\Large \textcolor{blue}{\boldmath$0.9364$}} & {\Large \textcolor{red}{\boldmath$0.1962$}},~~{\Large \textcolor{red}{\boldmath$0.1515$}} & {\Large \textcolor{red}{\boldmath$0.1903$}} & {\Large \textcolor{red}{\boldmath$0.7678$}} \\
    \hline
    \end{tabular}}
    \par\vspace*{2pt}
    \begin{minipage}{0.97\textwidth}
    \footnotesize
    \item \textbf{Methods.} \textbf{A1}: Proprioceptive sensor-only localization and decoupled mapping using the inverse model, \textbf{A2}: Decoupled localization and decoupled mapping using the inverse model, \textbf{A3}: Coupled localization and decoupled mapping using the inverse model, \textbf{A4}: Decoupled localization and \textsf{T-BayesMap}, \textbf{A5}: Coupled localization and \textsf{T-BayesMap} (proposed).
    \item \textcolor{red}{Red} and \textcolor{blue}{blue} denote the best and second-best results, respectively.
    \end{minipage}
\end{center}
\normalsize
\end{table*}

\subsection{Ablation Study using Open-Source Maps}

\subsubsection{Experimental Setup}

The experiments are performed in a custom-built simulator using the KTH floorplan dataset~\cite{aydemir2012can}, which contains over 100 campus floorplans annotated with wall and door locations. All maps are downsampled to a resolution of 5 pixels per meter ($0.2\,\text{m}$ per pixel). From this dataset, we selected three floorplans categorized into small (55m × 160m), medium (90m × 170m), and large (130m × 170m) environments, as shown in Fig.~\ref{fig:10_open_source_map}. For each floorplan, we evaluate performance using 10 distinct initial locations and 10 independent random initializations from a fixed starting point, yielding a total of 60 experiments.

The robot is equipped with odometry as a proprioceptive sensor and a 2D LiDAR as an exteroceptive sensor. The LiDAR provides a full $360^\circ$ field of view with a maximum range of $10\,\text{m}$ at $10\,\text{Hz}$ (e.g., RPLIDAR A1M8). Because this sensing range is smaller than the scale of the environments (e.g., large rooms or long corridors), each LiDAR scan covers only a limited portion of the surroundings. To emulate real-world data acquisition, zero-mean Gaussian noise with standard deviation $0.01\,\text{m}$ is added to each beam measurement. Odometry inputs are also corrupted with Gaussian noise with standard deviations of $0.05\,\text{m}$ for translational motion and $0.02\,\text{rad}$ for rotation. To further evaluate robustness in both localization and mapping, orientation corrections are perturbed with an additional $0.01\,\text{rad}$ noise term. The simulation time step is $0.02\,\text{s}$.

As shown in Fig.~\ref{fig:framework}, the estimated map $\mathbf{m}_t$ is used to extract frontiers, from which candidate goal regions are generated. For each goal, the candidate control sequence $\mathbf{u}_{t+1:t'} \in \mathcal{U}_t$ is produced by a low-level planner guided by a global planner ($A^\star$). Following Algorithm~\ref{alg:ActBex}, coupled uncertainty along each candidate sequence is predicted using uncertainty-aware estimation, and the resulting coupled BIG is evaluated. The control sequence that maximizes BIG is then used as a reference (warm-start) for the sampling-based local planner—Model Predictive Path Integral (MPPI)—and subsequently fed back into the estimation framework. All \textsf{B-ActiveSEAL} parameters are fixed across experiments: $\gamma = 1.2$, $R_o = 0.39^2$, $R_u = 3.0^2$, and $N_{\max} = 3$.

\begin{figure}[!t]
    \centering
    \begin{minipage}[t]{0.42\textwidth}
        \centering
        \includegraphics[width=\textwidth]{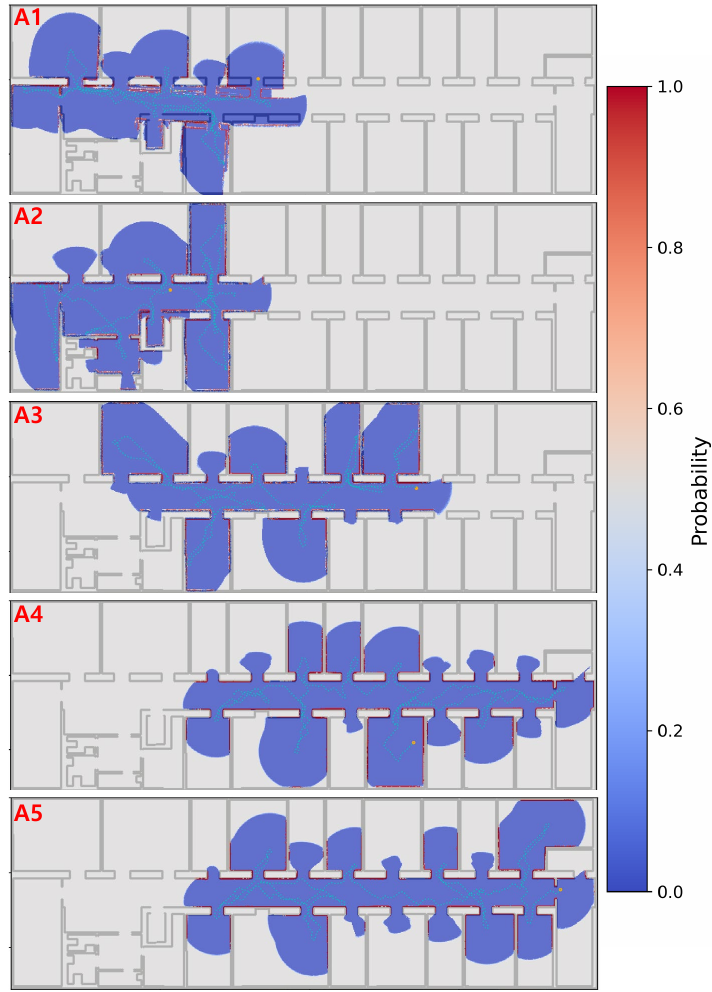}
    \end{minipage}
    \caption{{\small Qualitative comparisons of exploration and mapping results. A1 exhibits noticeable map drift due to the absence of LiDAR updates in localization. A2 and A3, which use decoupled mapping with the inverse model, produce jittered maps that degrade frontier detection—since frontiers may lie just beyond occupied cells—and consequently lead to poor exploration performance. In contrast, A4 and A5, which use \textsf{T-BayesMap}, produce stable, non-jittered maps that enable proper frontier detection and successful exploration. Additionally, A5 incorporates coupled localization, resulting in lower map error compared to A4.}}
    \label{fig::12_result}
\end{figure}

\smallskip
\subsubsection{Ablation Study for Localization–Mapping Frameworks}
In this study, we assess how \emph{coupled} uncertainty in localization–mapping affects both estimation accuracy and its closed-loop interaction with decision-making—that is, how estimation shapes exploratory actions, and how those actions, in turn, feed back into localization and mapping. To assess the contribution of each component within the localization–mapping framework, the decision-making process was fixed to~\eqref{eq::behavioral_entropy_cost} with $\alpha = 1$ (SE). Five ablation settings (A1–A5) were then evaluated, ranging from a proprioceptive sensor-only baseline (A1) to the full proposed framework (A5):
\begin{itemize}
    \item \textbf{A1}: Proprioceptive sensor-only localization and decoupled mapping using the inverse model,
    \item \textbf{A2}: Decoupled localization and decoupled mapping using the inverse model,
    \item \textbf{A3}: Coupled localization and decoupled mapping using the inverse model,
    \item \textbf{A4}: Decoupled localization and \textsf{T-BayesMap},
    \item \textbf{A5}: Coupled localization and \textsf{T-BayesMap} (proposed).
\end{itemize}

Using the quantitative results in Fig.~\ref{fig:11_result} and Table~\ref{Table_Ex1}, together with the qualitative results in Fig.~\ref{fig::12_result}, we observe that—although the same decision-making process is applied—the resulting exploration behaviors vary substantially depending on how coupled uncertainty is handled within the localization–mapping configuration. For A1, the map exhibits noticeable drift because no LiDAR updates are available to correct localization errors, leading to poor exploration behavior (i.e., slow reduction of map uncertainty), particularly as the map size increases. For A2 and A3, regardless of the localization strategy, the inverse model produces noisy (jittered) maps that degrade the decision-making process (e.g., unreliable frontier detection). As a result, the robot frequently becomes trapped in certain regions, yielding slow exploration.

In contrast, A4 and A5, which employ \textsf{T-BayesMap}, generate smooth (non-jittered) and consistent maps that support accurate frontier detection and effective exploration with progressively decreasing map uncertainty. A5—further incorporating coupled localization on top of A4—achieves the lowest translational and map errors overall, maintaining accurate frontier detection while minimizing map uncertainty throughout exploration. These results demonstrate that \emph{coupled} uncertainty affects not only estimation performance but also the overall quality of decision-making.

\smallskip
\subsubsection{Ablation Study for the Decision-Making Process}

We evaluate how \emph{coupled} uncertainty in decision-making shapes exploration behavior and estimation, and how the resulting estimation outcomes, in turn, influence subsequent decisions. To assess the contribution of each component within the decision-making process, the localization–mapping framework was fixed to the proposed configuration (A5). Three ablation settings (B1–B3) were then evaluated, ranging from a map-only setup with the inverse model (B1) to the proposed framework, \textsf{T-BayesMap} (B3):
\begin{itemize}
    \item \textbf{B1}: Map-only using the inverse model,
    \item \textbf{B2}: Map-only using the proposed model~\eqref{eq::likelihood_beam},
    \item \textbf{B3}: Map under coupled localization (\textsf{T-BayesMap}).
\end{itemize}

\begin{figure*}[!t]
    \centering
    \begin{minipage}[t]{0.25\textwidth}
        \centering
        \includegraphics[width=\textwidth]{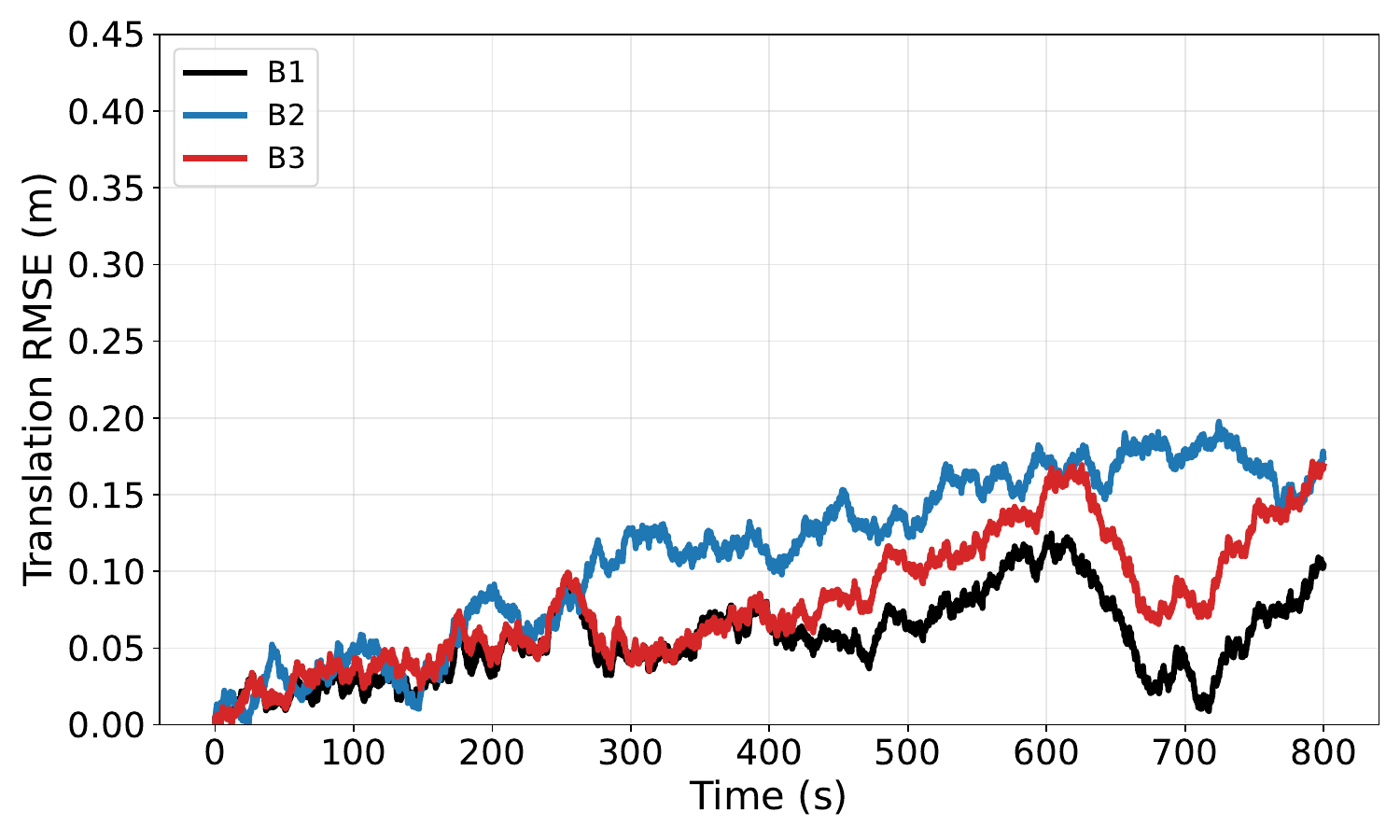}
    \end{minipage}
    \begin{minipage}[t]{0.25\textwidth}
        \centering
        \includegraphics[width=\textwidth]{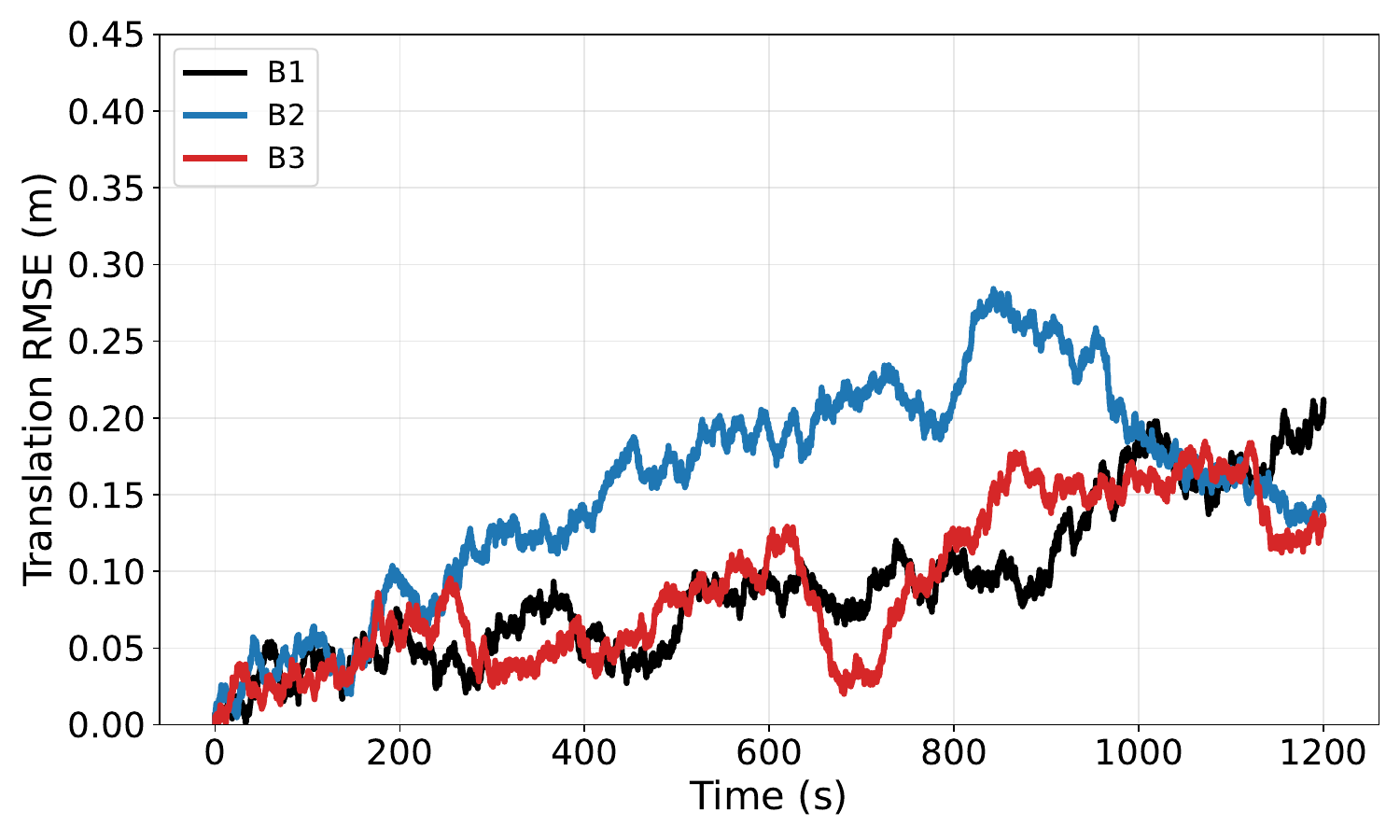}
    \end{minipage}
    \begin{minipage}[t]{0.25\textwidth}
        \centering
        \includegraphics[width=\textwidth]{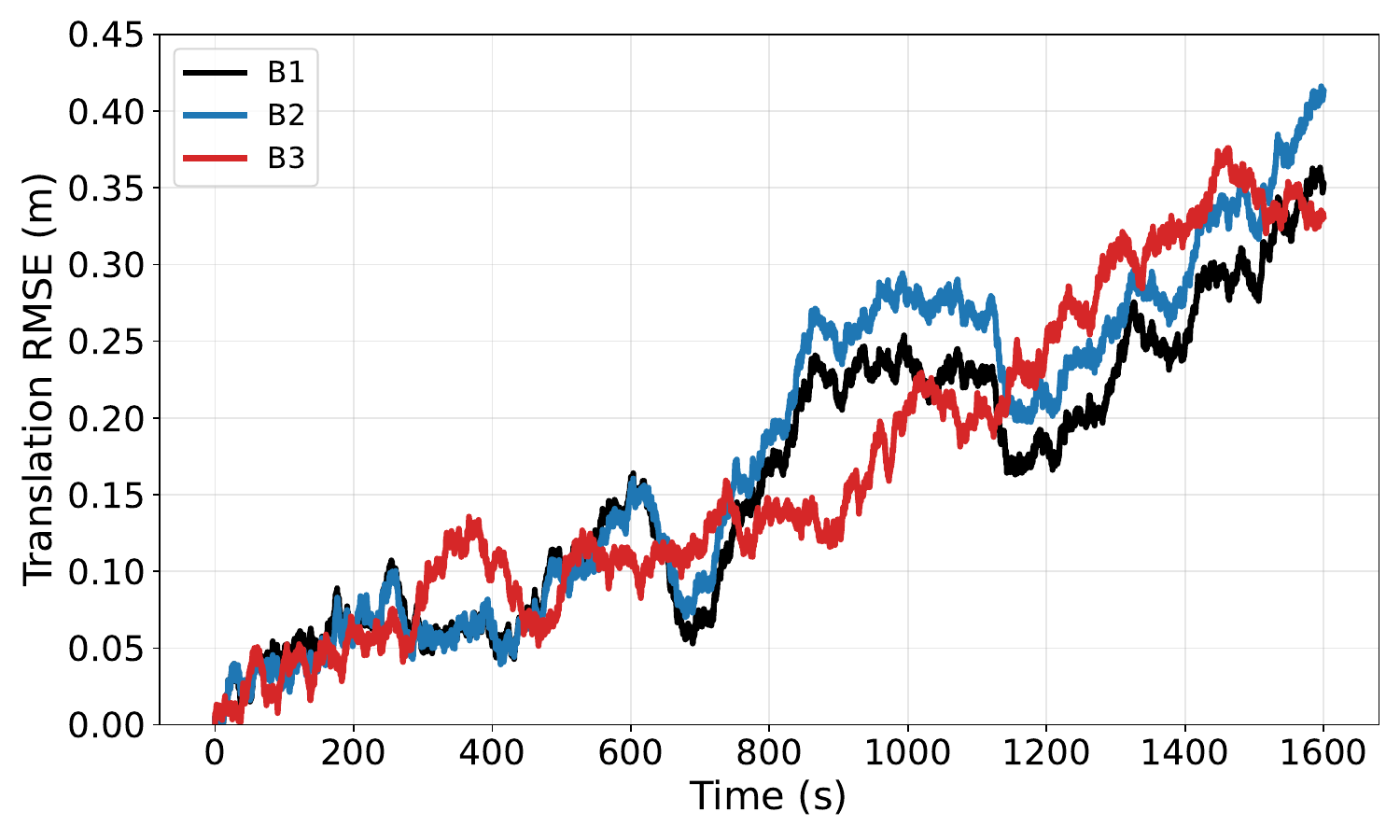}
    \end{minipage}
    \begin{minipage}[t]{0.25\textwidth}
        \centering
        \includegraphics[width=\textwidth]{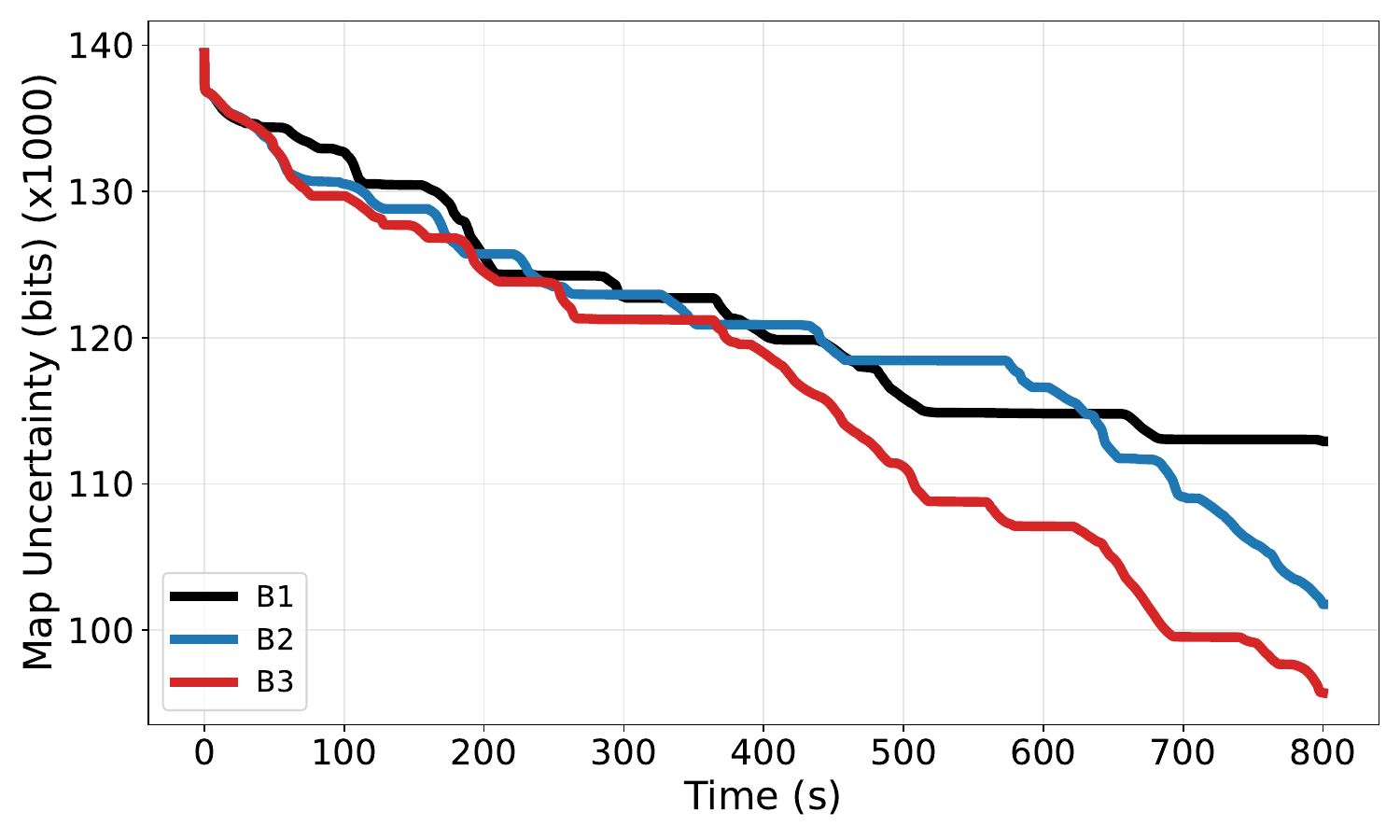}
    \end{minipage}
    \begin{minipage}[t]{0.25\textwidth}
        \centering
        \includegraphics[width=\textwidth]{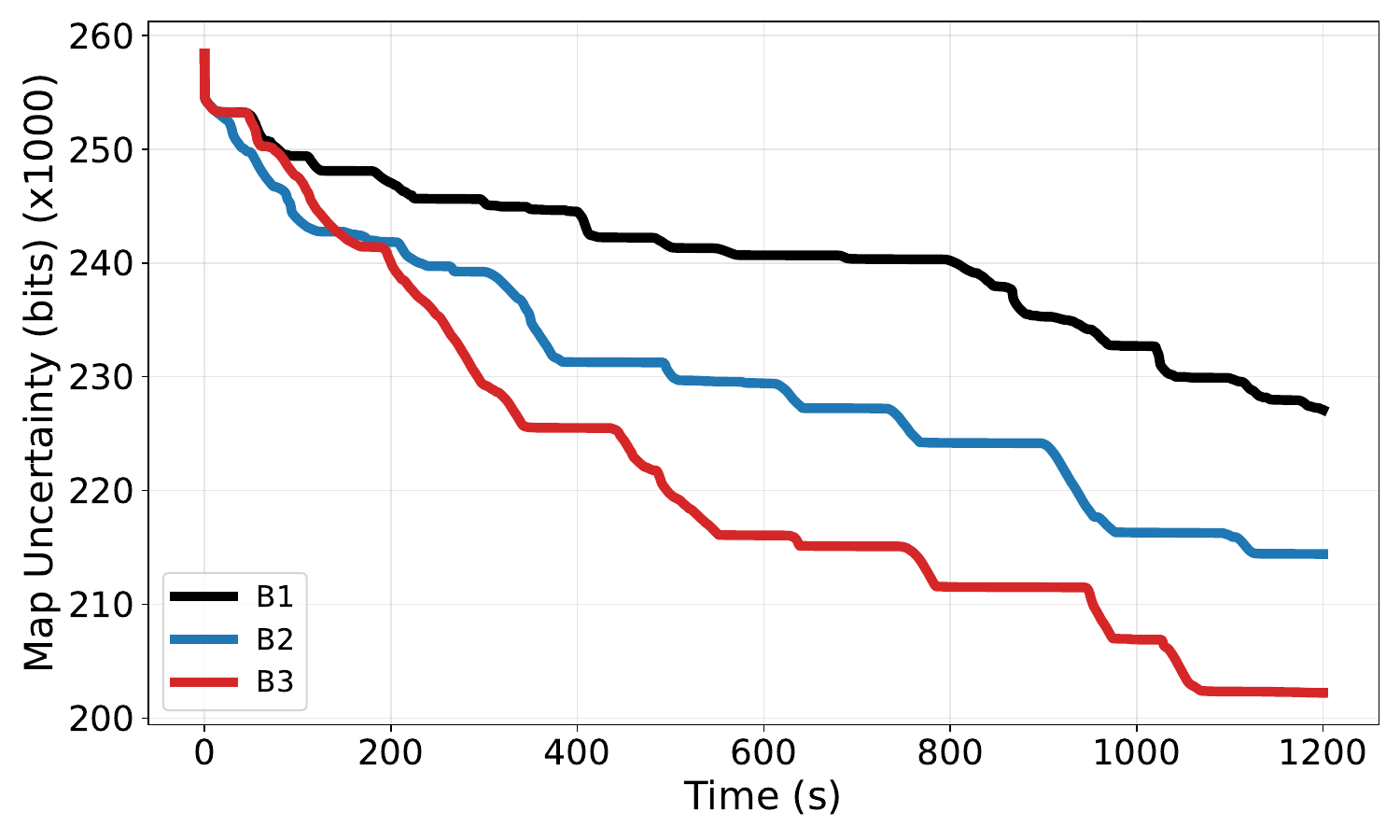}
    \end{minipage}
    \begin{minipage}[t]{0.25\textwidth}
        \centering
        \includegraphics[width=\textwidth]{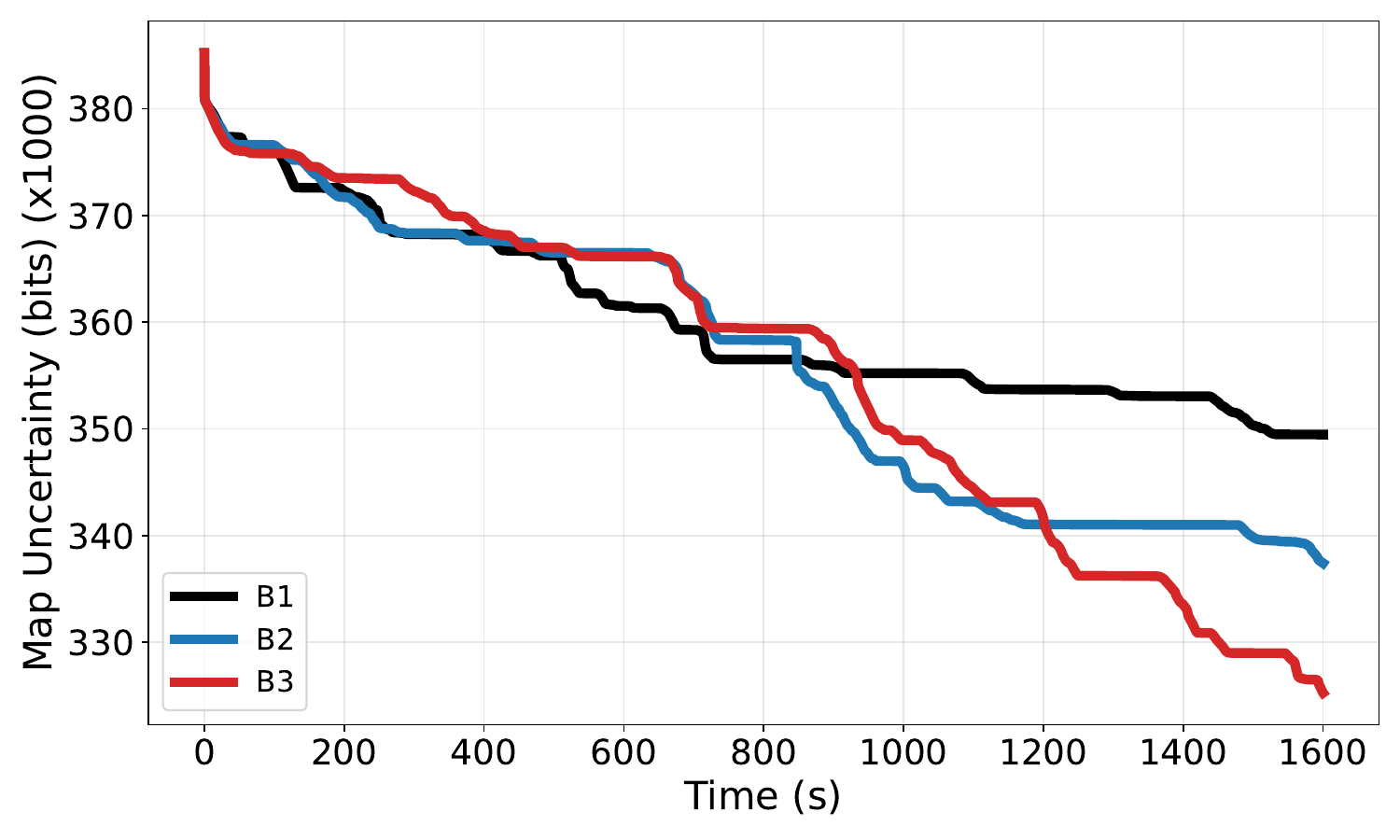}
    \end{minipage}
    \begin{minipage}[t]{0.251\textwidth}
        \centering
        \includegraphics[width=\textwidth]{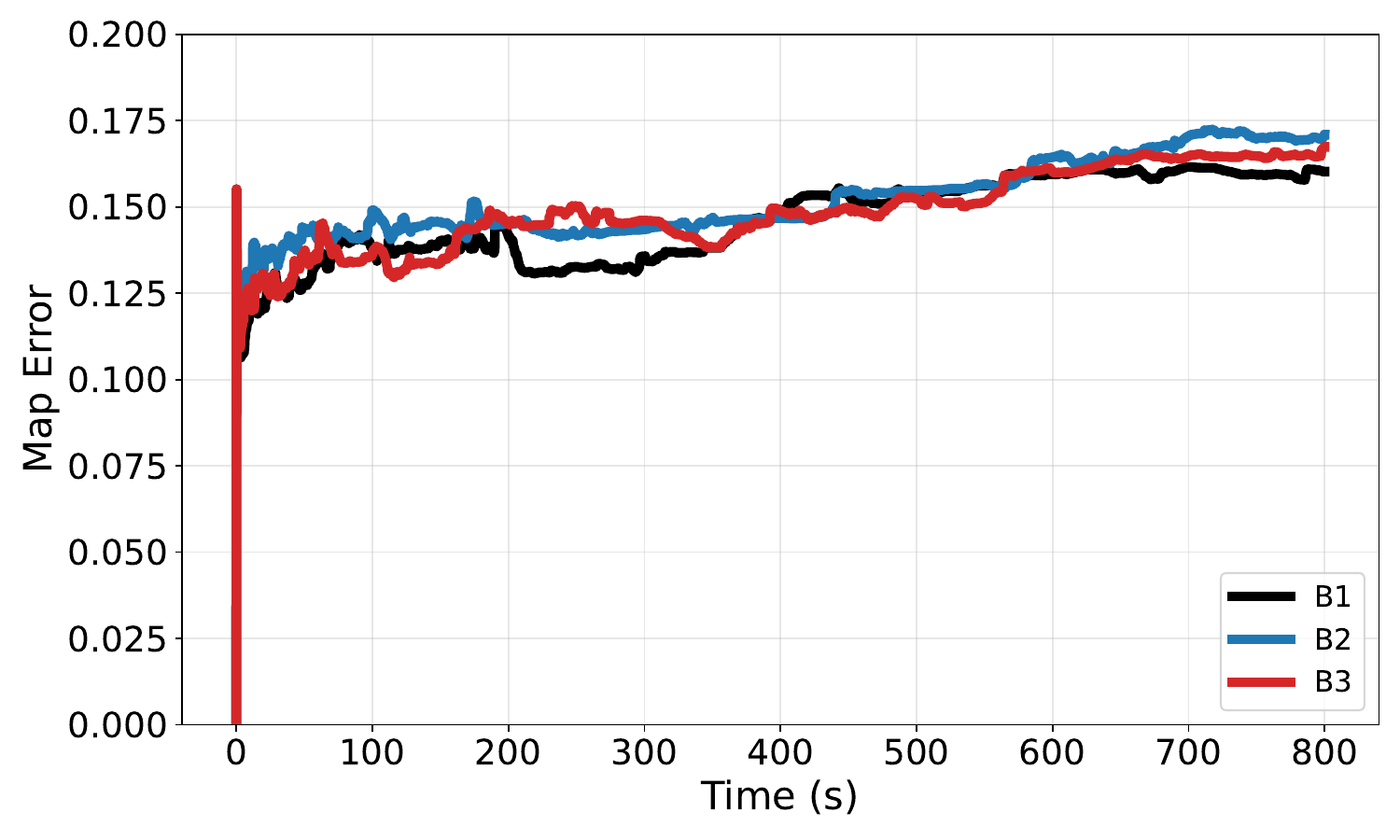}
        {{\scriptsize (a) Small Map (55m × 160m, 800s)}}
    \end{minipage}
    \begin{minipage}[t]{0.251\textwidth}
        \centering
        \includegraphics[width=\textwidth]{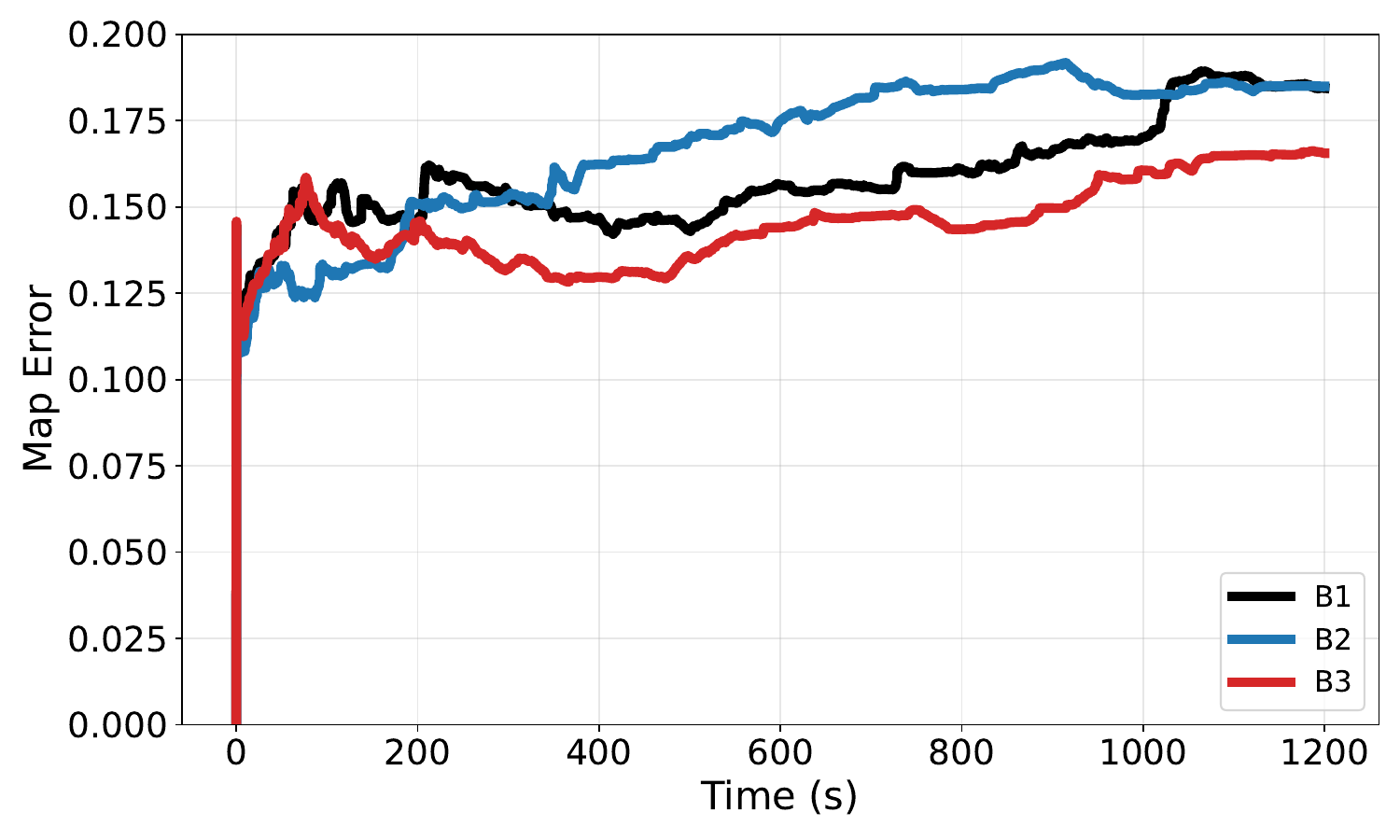}
        {{\scriptsize (b) Medium Map (90m × 170m, 1200s)}}
    \end{minipage}
    \begin{minipage}[t]{0.251\textwidth}
        \centering
        \includegraphics[width=\textwidth]{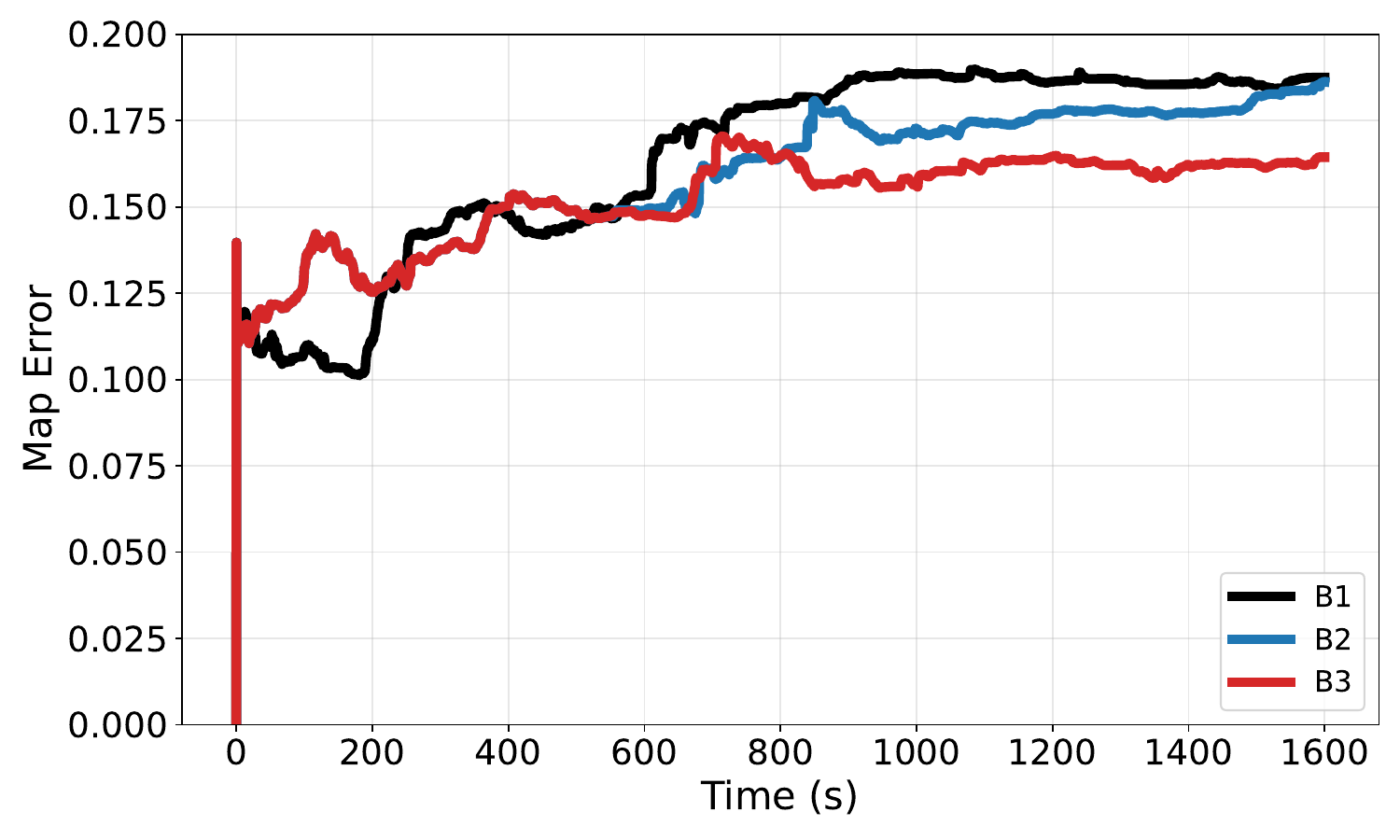}
        {{\scriptsize (c) Large Map (130m × 170m, 1600s)}}
    \end{minipage}
    \caption{{\small Comparisons of the mean results of the decision-making framework over time using 10 independent random initializations from a fixed initial location. B1 consistently exhibits poor exploration across all map sizes, which results in lower translational error compared to B2 and B3. B3, which employs \textsf{T-BayesMap}, achieves the highest exploration performance while maintaining low map error.}}
    \label{fig:13_result}
\end{figure*}

\begin{figure}[t]
    \centering
    \begin{minipage}[t]{0.48\textwidth}
        \centering
        \includegraphics[width=\textwidth]{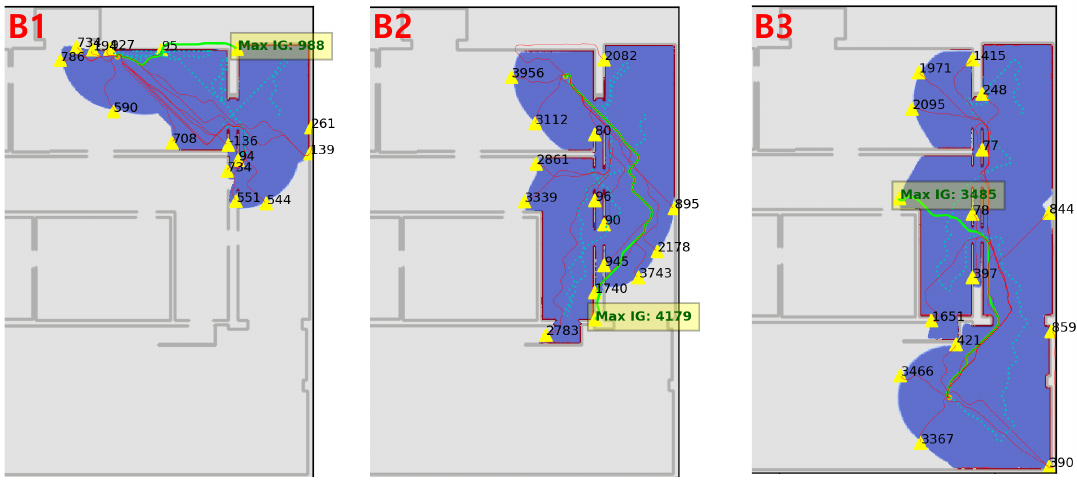}
    \end{minipage}
    \caption{{\small Qualitative illustration of the decision-making process. Yellow triangles indicate detected frontiers, and the green line marks the selected trajectory that maximizes BIG. B1 lacks localization-uncertainty awareness and generates an uninformative map, resulting in unreliable frontier selection and trajectory. B2 applies the proposed model, thus improving frontier detection, but still ignores localization uncertainty, causing the robot to move toward frontiers through unknown regions. B3 incorporates localization uncertainty (\textsf{T-BayesMap}), enabling reliable decisions and stable routing through well-known areas.}}
    \label{fig::12_result_dicision_making}
\end{figure}

Results in Fig.~\ref{fig:13_result} and Table~\ref{Table_Ex2} show that the mapping-only configuration with the inverse model (B1) appears to yield the smallest translational error. In fact, however, B1 neither incorporates localization uncertainty into decision-making nor produces an informative map. This combination leads to unreliable frontier selection (Fig.~\ref{fig::12_result_dicision_making}, Left), causing the robot to become stuck in certain regions and resulting in limited exploration. B2, which employs the proposed model for mapping, improves exploration by generating a more informative map that enables more reliable frontier detection than B1. However, like B1, its decision-making process ignores localization uncertainty, causing the robot to approach frontiers through unknown regions (Fig.~\ref{fig::12_result_dicision_making}, Middle) and thereby increasing translational error. B3, which incorporates localization uncertainty on top of B2 (\textsf{T-BayesMap}), achieves more extensive exploration while keeping localization uncertainty controlled. By deliberately routing through well-known areas before entering new regions (Fig.~\ref{fig::12_result_dicision_making}, Right), it enables consistent progress and maintains both reliable decision-making and low translational error.

\begin{table*}[t]
\begin{center}
\caption{Monte Carlo Ablation Study of the Decision-Making Process with 10 Distinct Initial Locations}
\label{Table_Ex2}
\renewcommand{\arraystretch}{1.25}
\large
\resizebox{0.98\textwidth}{!}{
    \begin{tabular}{c|c c c|c c c|c c c} 
    \hline
    & \multicolumn{9}{c}{\Large Map Size} \\
    \cline{2-10}
    & \multicolumn{3}{c}{\Large Small Map (55m × 160m, 800s)} & \multicolumn{3}{c}{\Large Medium Map (90m × 170m, 1200s)} & \multicolumn{3}{c}{\Large Large Map (130m × 170m, 1600s)} \\
    \cline{2-10}
    {\LARGE Methods} & Translation Error $\downarrow$ & Map Error $\downarrow$ & Map Uncertainty $\uparrow$ & Translation Error $\downarrow$ & Map Error $\downarrow$ & Map Uncertainty $\uparrow$ & Translation Error $\downarrow$ & Map Error $\downarrow$ & Map Uncertainty $\uparrow$ \\[-2pt]
    & RMSE ($m$), MAE ($m$) & RMSE & (bits/step) & RMSE ($m$), MAE ($m$) & RMSE & (bits/step) & RMSE ($m$), MAE ($m$) & RMSE & (bits/step) \\[3pt]
    \hline 
    \textbf{\Large B1} & {\Large \textcolor{red}{\boldmath$0.0617$}},~~{\Large \textcolor{red}{\boldmath$0.0486$}} & {\Large \textcolor{red}{\boldmath$0.1602$}} & {\Large $0.6639$} & {\Large \textcolor{red}{\boldmath$0.1029$}},~~{\Large \textcolor{red}{\boldmath$0.0813$}} & {\Large \textcolor{blue}{\boldmath$0.1844$}}& {\Large $0.5218$} & {\Large \textcolor{red}{\boldmath$0.1842$}},~~{\Large \textcolor{blue}{\boldmath$0.1528$}} & {\Large $0.2121$} & {\Large $0.5850$} \\
    \textbf{\Large B2} & {\Large $0.1246$},~~{\Large $0.0952$} & {\Large $0.1709$} & {\Large \textcolor{blue}{\boldmath$0.9424$}} & {\Large $0.1211$},~~{\Large $0.0957$} & {\Large $0.1849$} & {\Large \textcolor{blue}{\boldmath$0.7330$}} & {\Large $0.2096$},~~{\Large $0.1719$} & {\Large \textcolor{blue}{\boldmath$0.1977$}} & {\Large \textcolor{blue}{\boldmath$0.6447$}} \\
    \textbf{\Large B3} & {\Large \textcolor{blue}{\boldmath$0.0742$},~~\textcolor{blue}{\boldmath$0.0588$}} & {\Large \textcolor{blue}{\boldmath$0.1648$}} & {\Large \textcolor{red}{\boldmath$1.0937$}} & {\Large \textcolor{blue}{\boldmath$0.1032$}},~~{\Large \textcolor{blue}{\boldmath$0.0844$}} & {\Large \textcolor{red}{\boldmath$0.1655$}} & {\Large \textcolor{red}{\boldmath$0.9364$}} & {\Large \textcolor{blue}{\boldmath$0.1962$}},~~{\Large \textcolor{red}{\boldmath$0.1515$}} & {\Large \textcolor{red}{\boldmath$0.1903$}} & {\Large \textcolor{red}{\boldmath$0.7678$}}\\
    \hline
    \end{tabular}}
    \par\vspace*{2pt}
    \begin{minipage}{0.97\textwidth}
    \footnotesize
    \item \textbf{Methods.} \textbf{B1}: Map-only using the inverse model, \textbf{B2}: Map-only using the proposed model~\eqref{eq::likelihood_beam}, \textbf{B3}: Map under coupled localization (\textsf{T-BayesMap}). \\
    \textcolor{red}{Red} and \textcolor{blue}{blue} denote the best and second-best results, respectively.
    \end{minipage}
\end{center}
\normalsize
\end{table*}

\subsection{Behavioral Simulations in the ROS–Unity 3D Environment}

\subsubsection{Experimental Setup}

\begin{figure}[!t]
    \centering
    \begin{minipage}[t]{0.13\textwidth}
        \centering
        \includegraphics[width=\textwidth]{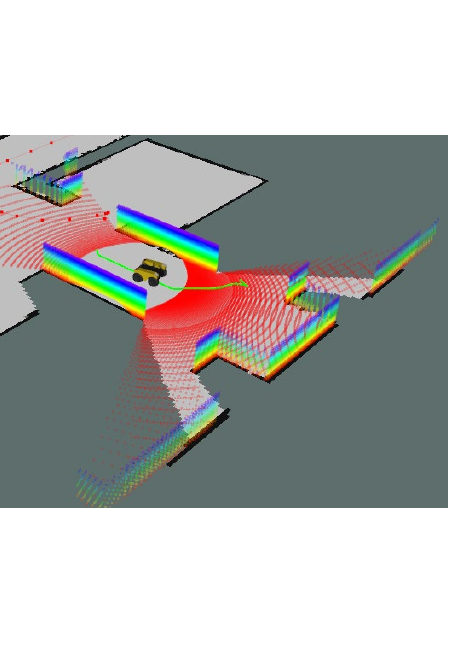}
        \scriptsize{(a) ROS-Unity 3D Sim}
    \end{minipage}
    \hskip\baselineskip
    \begin{minipage}[t]{0.12\textwidth}
        \centering
        \includegraphics[width=\textwidth]{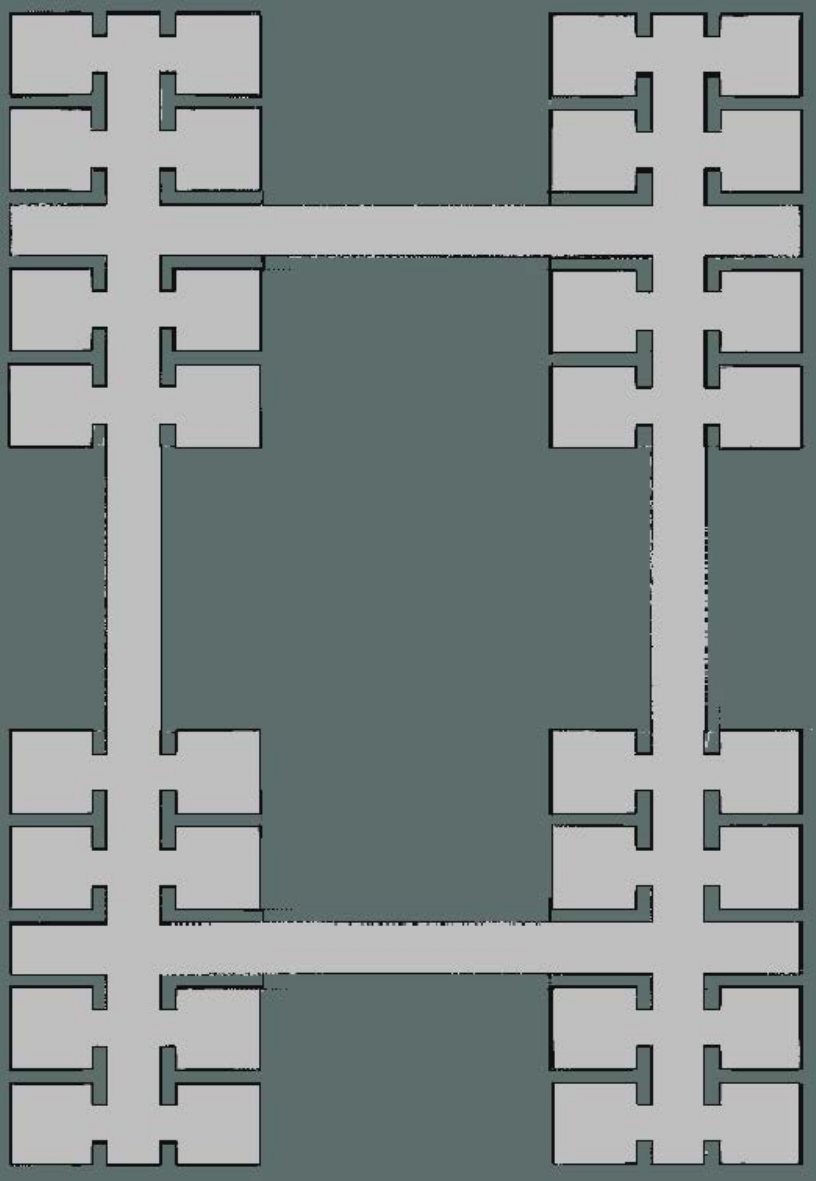}
        \scriptsize{(b) 32-room map}
    \end{minipage}
    \hskip\baselineskip
    \begin{minipage}[t]{0.124\textwidth}
        \centering
        \includegraphics[width=\textwidth]{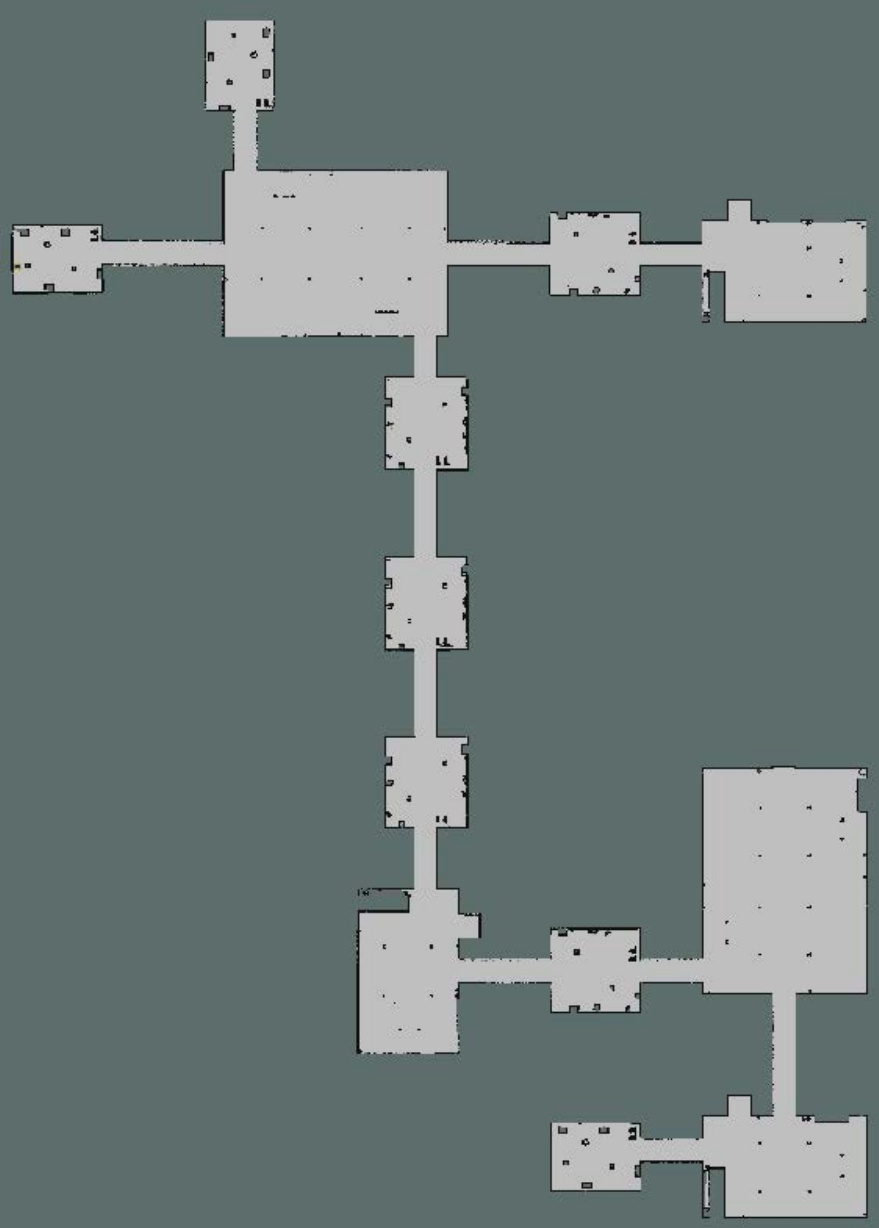}
        \scriptsize{(c) DARPA SubT map}
    \end{minipage}
    \caption{\small ROS–Unity 3D simulation environments with a 32-room map (125m × 169m) and a DARPA SubT map (220m × 291m).}
    \label{fig:7_ros_environment}
\end{figure}

We describe the simulation setup in the ROS–Unity 3D environments, as shown in Fig.~\ref{fig:7_ros_environment}. The first environment consists of four long corridors connected to 32 rooms, with no internal obstacles, and spans 125m × 169m. The second environment is based on the DARPA Subterranean Challenge Unity environment~\cite{rogers2021ISER}, with a map size of 220m × 291m. This environment includes a mix of large and small rooms, obstacles, and short corridors, making it a complex and challenging scenario for exploration. 

For the simulated hardware experiments, we use a Clearpath Warthog UGV\footnote{https://clearpathrobotics.com/warthog-unmanned-ground-vehicle-robot/} equipped with an Ouster LiDAR and odometry sensors. The LiDAR’s maximum range is set to $25$m to define the hit cell for each beam. All sensors and actuators are modeled according to real hardware specifications to ensure realistic simulation performance. The \textsf{B-ActiveSEAL} parameters match those used in the ablation studies across all experiments.

In this experiment, we study the $\alpha$-adjustable behavior of the proposed framework through Monte Carlo simulations. We evaluate three representative values of $\alpha$:
\begin{itemize}
    \item $\alpha=0.2$: capturing the regime $0 < \alpha <1$,
    \item $\alpha=1.0$: equivalent to Shannon entropy, and
    \item $\alpha=3.0$: representing $\alpha >1$.
\end{itemize}
Additionally, we evaluate exploration behavior using Rényi entropy (RE) with its corresponding $\alpha$, one of the widely used generalized entropy measures.

Finally, we compare our framework with RE-based active exploration with graph SLAM~\cite{carrillo2018autonomous}, which combines SE and RE to compute map entropy and localization uncertainty into a parameter $\alpha$ in RE, and with BE-based exploration with graph SLAM~\cite{suresh2024robotic}, which uses BE solely for evaluating map uncertainty.

\subsubsection{Results and Discussion}

\begin{figure}[t]
    \centering
    \begin{minipage}[t]{0.15\textwidth}
        \centering
        \includegraphics[width=\textwidth]{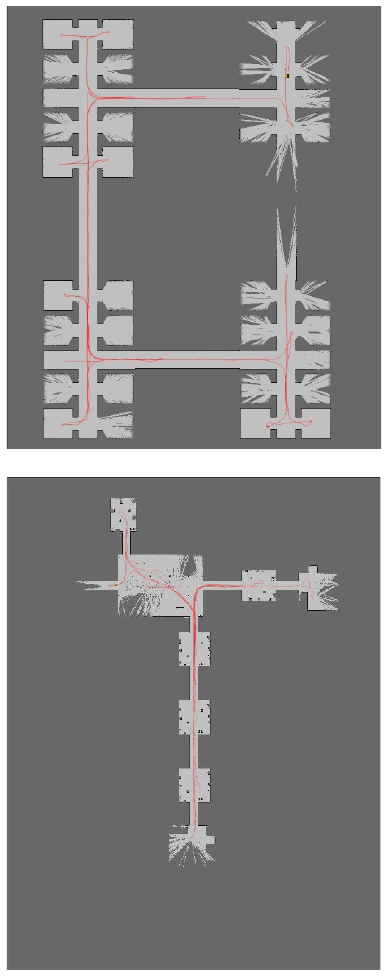}
        {{\scriptsize (a) $\alpha=0.2$}}
    \end{minipage}
    \hspace{-4pt}
    \begin{minipage}[t]{0.15\textwidth}
        \centering
        \includegraphics[width=\textwidth]{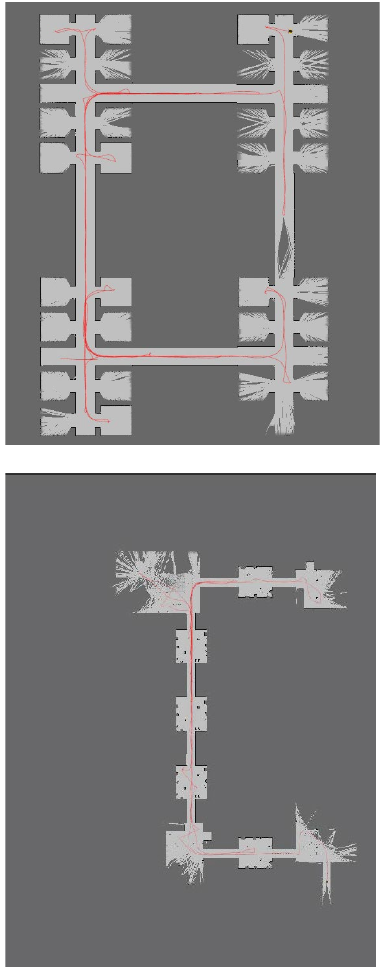}
        {{\scriptsize (b) $\alpha=1.0$}}
    \end{minipage}
    \hspace{-4pt}
    \begin{minipage}[t]{0.15\textwidth}
        \centering
        \includegraphics[width=\textwidth]{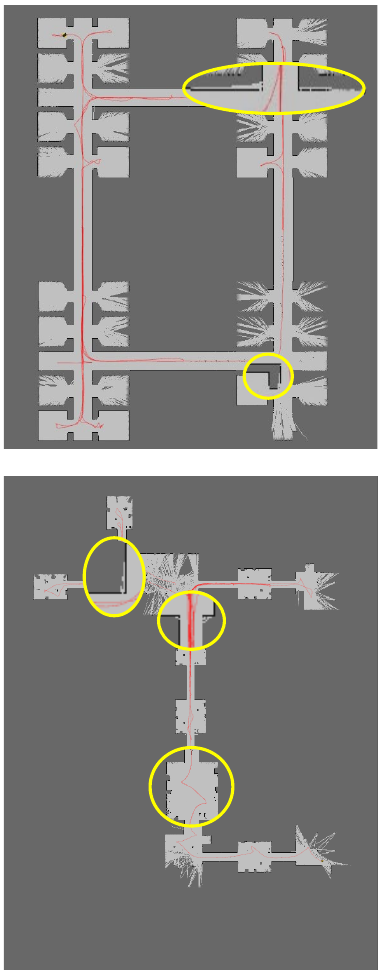}
        {{\scriptsize (c) $\alpha=3.0$}}
    \end{minipage}
    \caption{{\small Qualitative comparisons of \textsf{B-ActiveSEAL} (red line) and mapping results for different values of $\alpha$, evaluated at the same simulation time in the ROS–Unity 3D environment. For the 32-room map (top row), $\alpha = 0.2$ causes the robot to repeatedly enter rooms, yielding the slowest overall coverage. With $\alpha = 3.0$, the robot prioritizes long-corridor expansion and explores substantially faster, though with slightly increased map error (yellow circles). For the DARPA SubT map (bottom row), a similar pattern emerges: $\alpha = 0.2$ yields highly local exploration with repeated revisits to previously known areas, whereas $\alpha = 3.0$ drives corridor-based expansion into distant unexplored regions, again with slightly higher map error (yellow circles).}}
    \label{fig:15_ROS_results}
\end{figure}

\begin{figure}[!t]
\centering
\begin{minipage}[t]{0.45\textwidth}
\centering
\includegraphics[width=\textwidth]{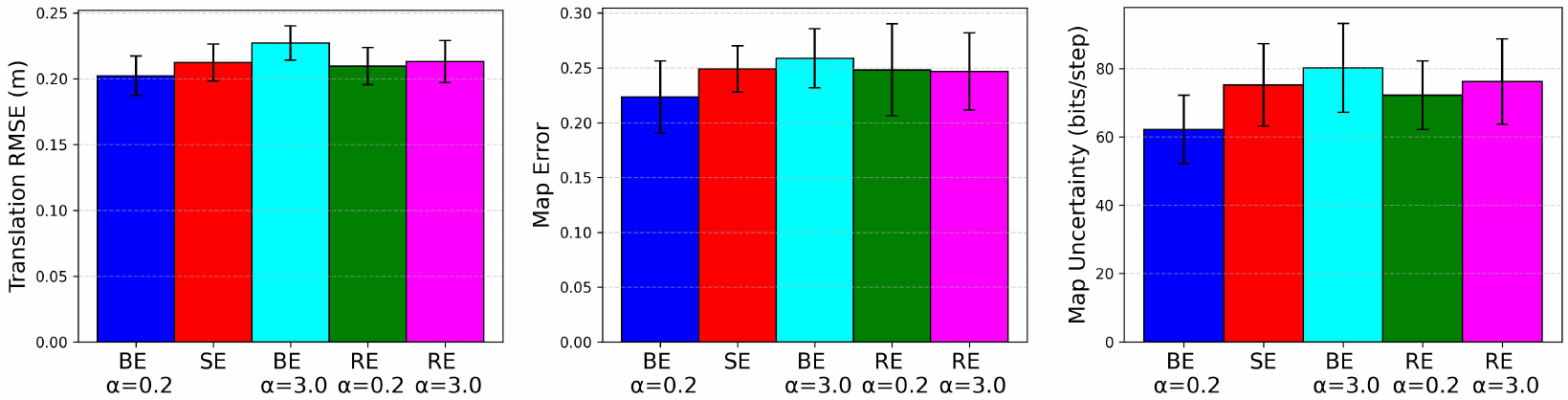}
{{\scriptsize (a) 32-room map}}
\end{minipage}
\vspace{1pt}
\begin{minipage}[t]{0.45\textwidth}
\centering
\includegraphics[width=\textwidth]{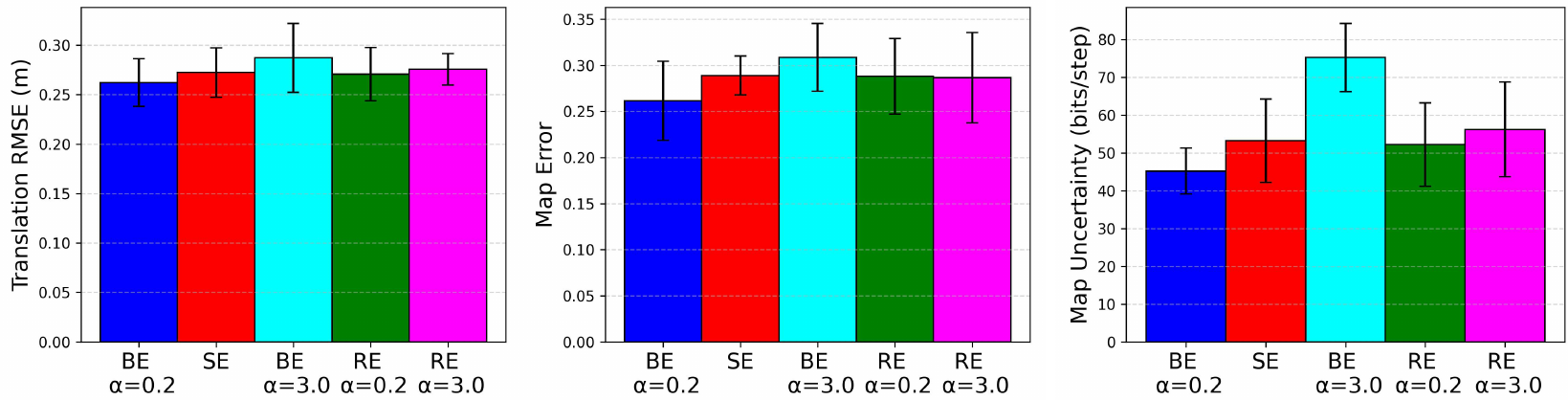}
{{\scriptsize (b) DARPA SubT map }}
\end{minipage}
\caption{{\small Quantitative comparison of Behavioral entropy (BE), Shannon entropy (SE), and Rényi entropy (RE)–based exploration across two ROS–Unity environments (32-room and DARPA SubT), evaluated over 10 Monte Carlo runs. Bars denote the mean with $1\sigma$ standard-deviation bounds. BE with different values of $\alpha$ produces clearly distinguishable exploratory behaviors, whereas RE—despite varying $\alpha$ from $0.2$ to $3.0$—shows minimal behavioral change, indicating limited expressiveness and weak controllability of exploration compared to BE. Notably, in the more complex DARPA SubT environment, the differences among BE-driven exploration behaviors across $\alpha$ values are more pronounced than in the regular, structured 32-room map.}}
\label{fig:16_ROS_stat_results}
\end{figure}

We evaluate exploration behavior under different values of $\alpha$ using a realistic hardware model in the real-world 3D environment. In the 32-room environment (Fig.~\ref{fig:15_ROS_results}, top row), which features long corridors and adjacent rooms, $\alpha = 0.2$ causes the robot to repeatedly enter rooms, yielding the slowest overall coverage. In contrast, $\alpha = 3.0$ prioritizes long corridor-driven expansion and achieves substantially faster exploration, albeit with slightly increased map error. A similar pattern appears in the DARPA SubT environment (Fig.~\ref{fig:15_ROS_results}, bottom row): $\alpha = 0.2$ results in repeated local exploration to unknown areas, whereas $\alpha = 3.0$ drives corridor-following behavior that reaches distant unexplored areas. As shown by the per-step reduction in map uncertainty in the statistical results (Fig.~\ref{fig:16_ROS_stat_results}), these distinctions are even more pronounced in the more complex SubT environment than in the structured 32-room map.

Furthermore, when comparing with RE-driven exploration, changing $\alpha$ in RE produces minimal behavioral variation—indicating limited expressiveness (see Fig.~\ref{fig::Generalized_Entropy}) and weak controllability compared to BE. Additional examples in Fig.~\ref{fig:17_Appen_1} demonstrate that $\alpha$ can be adjusted based on environment structure and sensor characteristics, enabling environment-adaptive exploration behavior.

Next, we compare our framework against existing active exploration approaches (Table~\ref{Table_Ex3}). In BE with graph SLAM, although loop closure is available, the decision-making process considers only map uncertainty. Consequently, loop closures respond passively to localization uncertainty rather than being actively promoted, resulting in faster exploration but noticeably larger translational and mapping errors.
In SE–RE with graph SLAM, localization uncertainty is incorporated heuristically through the parameter $\alpha$, while map uncertainty is computed by the difference between SE and RE. This coupling is limited in expressiveness and provides weak guidance for uncertainty-aware action selection. Even with active loop closure—where the robot revisits known regions to reduce localization uncertainty— translational and map errors persist compared to our approach. In contrast, our proposed method employs BE under coupled localization–mapping uncertainty, yielding a richer and more coherent representation of both uncertainty sources. This yields more reliable decisions and consistently lower localization and mapping errors than competing methods, with only a modest increase in exploration steps from more frequent routing through well-known areas. Overall, the results demonstrate that our approach naturally balances exploration and exploitation more effectively than prior methods.

\begin{table}[t]
\begin{center}
\caption{Quantitative Comparison of Exploration Results}
\label{Table_Ex3}
\scriptsize 
\resizebox{0.46\textwidth}{!}{
    \begin{tabular}{c|c c c} 
    \hline
    & \multicolumn{3}{c}{32-room map}  \\
    \cline{2-4}
    Methods & Translation Error $\downarrow$ & Map Error $\downarrow$ & $90\%$ Completion Steps $\downarrow$ \\
    & RMSE(m) & RMSE & step \\
    \hline
    BE with graph SLAM ($\alpha=3.0$) & 0.3154 & 0.3202 & 18592 \\
    SE-RE with graph SLAM & 0.2586 & 0.2886 & 24894 \\
    Ours, \textsf{B-ActiveSEAL} ($\alpha=3.0$) & 0.2499 & 0.2715 & 28832 \\
    \hline
    & \multicolumn{3}{c}{DARPA SubT map}  \\
    \cline{2-4}
    Methods & Translation Error $\downarrow$ & Map Error $\downarrow$ & $90\%$ Completion Steps $\downarrow$ \\
    & RMSE(m) & RMSE & step \\
    \hline
    BE with graph SLAM ($\alpha=3.0$) & 0.3865 & 0.3775 & 51253\\
    SE-RE with graph SLAM & 0.2912 & 0.3385 & 61821\\
    Ours, \textsf{B-ActiveSEAL} ($\alpha=3.0$) & 0.2861 & 0.3225 & 69542\\
    \hline
    \end{tabular}}
\end{center}
\normalsize
\end{table}

\section{Conclusions} \label{sec::con}

In this work, we propose an uncertainty-aware active exploration framework that tightly couples localization and mapping uncertainties within a unified, entropy-driven decision-making formulation. By breaking the classical separation between localization and mapping in the decision-making process, the framework provides a principled alternative to heuristic approaches and naturally balances the trade-off between exploration and exploitation.

Motivated by human perceptual decision theory, we integrated \emph{Behavioral entropy} (BE) as a generalized entropy measure within the proposed active exploration framework (\textsf{B-ActiveSEAL}). BE enables adjustable, uncertainty-aware action selection, allowing the robot to adapt its exploration strategy to diverse environments and sensing conditions while naturally balancing exploration and exploitation.

We validated the proposed framework through theoretical analysis and extensive ablation studies that quantify how coupled uncertainty influences both estimation and decision-making. Additional evaluations across multiple simulated and real-world–scale 3D environments further showed that the method consistently improves exploration robustness, frontier detection and selection reliability, and overall mapping quality. Although implemented within a filtering-based architecture, the framework achieves stable and reliable performance by explicitly managing coupled uncertainty throughout the entire estimation–decision-making pipeline.

Overall, this work establishes coupled uncertainty and Behavioral entropy as principled mechanisms that support adaptive and reliable active exploration.


\appendix
\subsection{Proof of Lemma~\ref{lem::4_1}} \label{sec::appen_A}
The posterior of localization can be represented via marginalization over the predicted map as
\begin{align}
    &p(\mathbf{x}_t|\mathbf{u}_{1:t},\mathbf{z}_{1:t}) \nonumber \\
    &\propto p(\mathbf{z}_t|\mathbf{x}_t) p(\mathbf{x}_t|\mathbf{u}_{1:t},\mathbf{z}_{1:t-1})
    \nonumber \\
    &= \prod\nolimits_{\{ k|\mathcal{I}^{h_k}_t\neq \emptyset\}} p(z^k_t|\mathbf{x}_t)^{\frac{\gamma}{N^h_t}} p(\mathbf{x}_t|\mathbf{u}_{1:t},\mathbf{z}_{1:t-1}) \nonumber \\
    &= \!\!\!\!\!\!\! \prod_{\{ k|\mathcal{I}^{h_k}_t\neq \emptyset\}} \!\!\!\!\!\!\!\! \prod\nolimits_{i \in \mathcal{I}^{h_k}_t} \!\! \biggl(\sum_{m^i_t \in \{0, 1\}} \!\!\!\!\! p(z^k_t| \mathbf{x}_t, m^i_t) p(m^i_t|\mathbf{u}_{1:t-1},\mathbf{z}_{1:t-1}) \! \biggl)^{\frac{\gamma}{N^h_t}}  \nonumber \\
    &\quad~ \times \underbrace{p(\mathbf{x}_t|\mathbf{u}_{1:t},\mathbf{z}_{1:t-1})}_{\text{$\mathcal{N}(\mathbf{x}_t;\bar{\mu}_t, \bar{\Sigma}_t)$}}, \nonumber
\end{align}
where 
\begin{align}
    p(m^i_t|\mathbf{u}_{1:t-1},\mathbf{z}_{1:t-1}) = p(m^i_{t-1}|\mathbf{u}_{1:t-1},\mathbf{z}_{1:t-1}) \nonumber
\end{align}
is derived from the assumption of a static environment. Here, the marginalization term is computed efficiently in the moment form.

Since $m^i_t \in \{0, 1\}$ is the binary random variable, the marginalization term can be represented as a weighted sum of two Gaussian distributions with the same mean but different covariances.

We employ a Gaussian approximation for the marginalization term. Then, the posterior can be approximated as
\begin{align}
    &\approx \!\!\!\!\!\!\! \prod_{\{ k|\mathcal{I}^{h_k}_t\neq \emptyset\}} \!\!\!\!\!\!\!\! \prod\nolimits_{i \in \mathcal{I}^{h_k}_t} \mathcal{N}(z^k_t; h(\mathbf{x}_t, m^i_t), \, R^i_{\mathbf{x}})^{\frac{\gamma}{N^h_t}}~\mathcal{N}(\mathbf{x}_t;\bar{\mu}_t, \bar{\Sigma}_t). \nonumber
\end{align}
Thus, $p(\mathbf{x}_t|\mathbf{u}_{1:t},\mathbf{z}_{1:t}) \approx \mathcal{N}(\mathbf{x}_t; \mu_t, \Sigma_t)$ with~\eqref{eq::update_localization}. The approximation arises from the Jacobian matrix $H_{\mathbf{x}}^i$. 

\subsection{Proof of Lemma~\ref{lem::4_2}} \label{sec::appen_B}
The posterior of $i$-th map $m^i_t$, where $i \in \mathcal{I}^k_t$, can be represented via marginalization over the updated localization as
\begin{align}
    &p(m^i_t|\mathbf{u}_{1:t},\mathbf{z}_{1:t}) \nonumber \\
    &\quad=\eta_{m^i_t} p(\mathbf{z}_t|m^i_t) p(m^i_t|\mathbf{u}_{1:t-1},\mathbf{z}_{1:t-1})  \nonumber \\
    &\quad= \eta_{m^i_t} \biggr(\int_{\mathbf{x}_t} p(z^k_t| \mathbf{x}_t, m^i_t)~p(\mathbf{x}_t|\mathbf{u}_{1:t},\mathbf{z}_{1:t}) \text{d}\mathbf{x}_t\biggr)^{w^i_t} \nonumber \\ 
    &\quad~~~\times p(m^i_t|\mathbf{u}_{1:t-1},\mathbf{z}_{1:t-1}), \nonumber
\end{align}
where the marginalization term can be solved analytically with $H_{\mathbf{x}}^i$ as
\begin{align}
     \int_{\mathbf{x}_t} \!\!\!\!\! \underbrace{p(z^k_t| \mathbf{x}_t, m^i_t)}_{\text{$\mathcal{N}(z^k_t; h(\mathbf{x}_t, m^i_t), \, R)$}} \underbrace{p(\mathbf{x}_t|\mathbf{u}_{1:t},\mathbf{z}_{1:t})}_{\mathcal{N}(\mathbf{x}_t; \mu_t, \Sigma_t)} \text{d}\mathbf{x}_t \approx \mathcal{N}(z^k_t; \hat{\mu}_t, \hat{\Sigma}^i_t), \nonumber
\end{align}
and the normalization term $\eta_{m^i_t}$ is computed in closed form:
\begin{align}
    \eta_{m^i_t} = \sum_{m^i_t \in \{0, 1\}} \mathcal{N}(z^k_t; \hat{\mu}_t, \hat{\Sigma}^i_t)^{w^i_t}p(m^i_t|\mathbf{u}_{1:t-1},\mathbf{z}_{1:t-1}). \nonumber
\end{align}
Thus, $p(m^i_t|\mathbf{u}_{1:t},\mathbf{z}_{1:t})$ become~\eqref{eq::update_map}.

\subsection{Proof of Theorem~\ref{thm::5_1}} \label{sec::appen_C}
We show that the predicted map BE decreases under lower localization uncertainty, and consequently, BIG increases. We focus on the $i$-th cell $m^i$, since $\mathbf{m}$ is the summation over all $m^i$ under Assumption~\ref{assum::indep_cell}.

For any $\alpha > 0$, the predicted map BE for the $i$-th cell at time $t+1$ is given by $\mathbb{H}^B_{\alpha}\bigl(p(m^i_{t+1}|\mathbf{u}_{t+1},z^k_{t+1})\bigl)$. The change in the predicted map BE is proportional to the change in the log-posterior of the $i$-th cell, $\ln{p(m^i_{t+1}|\mathbf{u}_{t+1},z^k_{t+1})}$, which in turn is proportional to the log-likelihood function of the $i$-th cell: 
\begin{align}
    &\Big|\mathbb{H}^B_{\alpha}\bigl(p(m^i_{t+1}|\mathbf{u}_{t+1},z^k_{t+1})\bigl)\Big| \propto \\
    &\qquad\qquad~~~ \Big|w^i_{t+1}(z^k_{t+1} - \hat{\mu}_{t+1})^\top (\hat{\Sigma}^i_{t+1})^{-1}(z^k_{t+1} - \hat{\mu}_{t+1})\Big|, \nonumber
\end{align}
where $\hat{\Sigma}^i_{t+1} = \mathbf{H}^i_{\mathbf{x}} \Sigma_{t+1} (\mathbf{H}^i_{\mathbf{x}})^{\top} \!+ R$ and $R$ is either $R_o$ or $R_u$. Given $w^i_{t+1}$, $z^k_{t+1}$, and $\hat{\mu}_{t+1}$, the likelihood change is maximized when $\mathbf{H}^i_{\mathbf{x}} \Sigma_{t+1} (\mathbf{H}^i_{\mathbf{x}})^{\top}$ is minimized.

Now consider two different covariances, $\Sigma_{t+1}$ and $\bar{\Sigma}_{t+1}$, where $\Sigma_{t+1} \leq \bar{\Sigma}_{t+1}$. Since the BE with $\beta = e^{(1-\alpha)\ln(\ln L)}$, $L=2$, is a monotonically decreasing function, it follows that
\begin{align}
    &\mathbb{H}^B_{\alpha}\bigl(p(m^i_{t+1}|\mathbf{u}_{t+1},z^k_{t+1})\bigl) \\ 
    &\qquad\qquad\qquad\quad \leq~ \mathbb{H}^B_{\alpha}\bigl(p(\bar{m}^i_{t+1}|\mathbf{u}_{t+1},z^k_{t+1})\bigl). \nonumber
\end{align}
By Assumption~\ref{assum::indep_cell}, summing over all $i$ yields~\eqref{eq::theorem_BE}. Since the maps at time $t$ are identical, the corresponding current BEs are also equal:
\begin{align} \label{eq::proof_BE}
    \mathbb{H}^B_{\alpha}\bigl(p(\mathbf{m}_t| \mathbf{u}_t,\mathbf{z}_t)\bigl) ~=~ \mathbb{H}^B_{\alpha}\bigl(p(\mathbf{\bar{m}}_t|\mathbf{u}_t,\mathbf{z}_t)\bigl).
\end{align}

Given~\eqref{eq::proof_BE}, the BE inequality in~\eqref{eq::theorem_BE} leads to the BIG inequality in~\eqref{eq::theorem_BIG}, thereby completing the proof.

\subsection{Proof of Theorem~\ref{thm::5_2}} \label{sec::appen_D}

Based on Theorem 3 in~\cite{suresh2024robotic}, the Behavioral entropy satisfies $\mathbb{H}^B_{\alpha > 1} \leq \mathbb{H}^B_{\alpha = 1} \leq \mathbb{H}^B_{0 < \alpha < 1}$, with equality only when $p(m^i_t) = 0.5$. Therefore, when the robot explores unknown areas (i.e., $p(m^i_t) = 0.5$), the rate of change in entropy is greater for $\alpha > 1$ compared to $\alpha = 1$ as  $p(m^i_t)$ deviates from $0.5$: $\left| \frac{\partial \mathbb{H}^B_{\alpha > 1}}{\partial p(m^i_t)} \right| > \left| \frac{\partial \mathbb{H}^B_{\alpha = 1}}{\partial p(m^i_t)} \right| \quad \text{for} \quad p(m^i_t) \neq 0.5$. In contrast, the rate of change is less for $0 < \alpha < 1$ compared to $\alpha = 1$: $\left| \frac{\partial \mathbb{H}^B_{0<\alpha < 1}}{\partial p(m^i_t)} \right| < \left| \frac{\partial \mathbb{H}^B_{\alpha = 1}}{\partial p(m^i_t)} \right| \quad \text{for} \quad p(m^i_t) \neq 0.5$. Finally, by Theorem~\ref{thm::5_1}, changes in localization uncertainty $\Sigma_t$ induce corresponding changes in the predicted occupancy probability $p(m^i_t)$. Thus, the sensitivity of BE to localization uncertainty $\Sigma_t$ directly follows from its sensitivity to $p(m^i_t)$. Hence, the theorem is established.

\subsection{Qualitative Results for \texorpdfstring{$\alpha$}{alpha}-Behavioral Exploration Across Different Environments} \label{sec::appen_E}

We additionally present qualitative simulation results demonstrating the behavior of \textsf{B-ActiveSEAL} under different values of $\alpha$ across multiple environments.

\begin{figure}[ht]
\centering
\begin{minipage}[t]{0.31\textwidth}
\centering
\includegraphics[width=\textwidth]{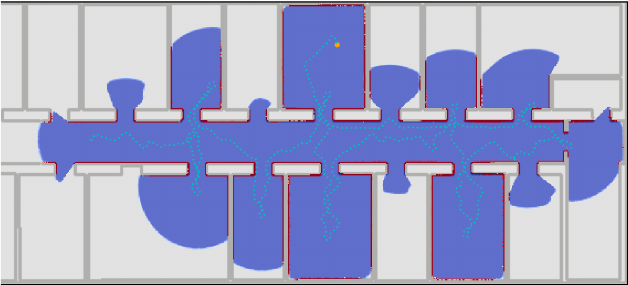}
{{\scriptsize (a) $\alpha = 0.2$ }}
\end{minipage}
\vspace{2pt}
\begin{minipage}[t]{0.31\textwidth}
\centering
\includegraphics[width=\textwidth]{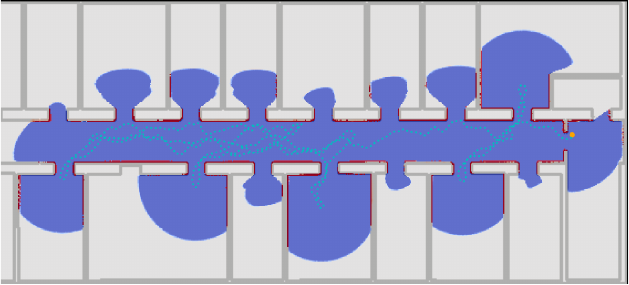}
{{\scriptsize (b) $\alpha = 3.0$}}
\end{minipage}
\vspace{2pt}
\begin{minipage}[t]{0.18\textwidth}
\centering
\includegraphics[width=\textwidth]{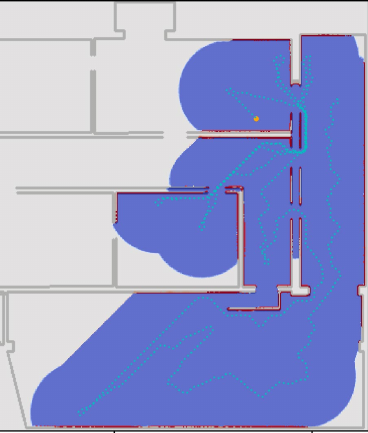}
{{\scriptsize (c) $\alpha = 0.2$}}
\end{minipage}
\hspace{2pt}
\begin{minipage}[t]{0.18\textwidth}
\centering
\includegraphics[width=\textwidth]{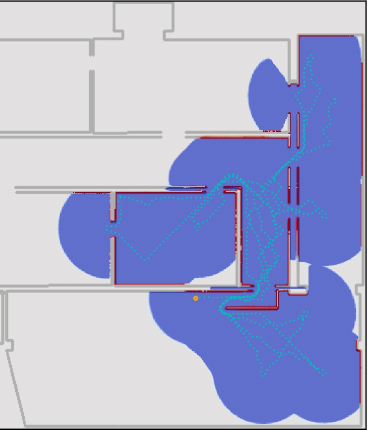}
{{\scriptsize (d) $\alpha = 3.0$}}
\end{minipage}
\begin{minipage}[t]{0.23\textwidth}
\centering
\includegraphics[width=\textwidth]{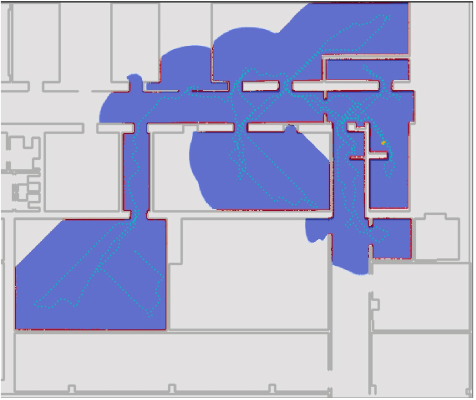}
{{\scriptsize (e) $\alpha = 0.2$}}
\end{minipage}
\hspace{2pt}
\begin{minipage}[t]{0.23\textwidth}
\centering
\includegraphics[width=\textwidth]{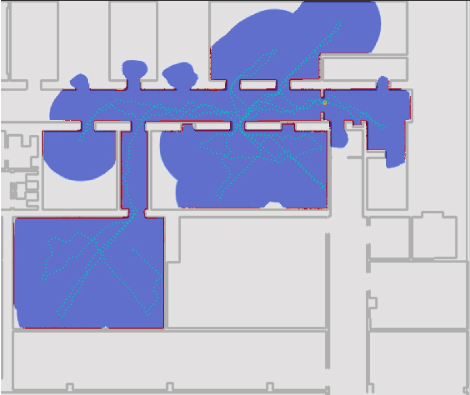}
{{\scriptsize (f) $\alpha = 3.0$}}
\end{minipage}
\caption{{\small Qualitative \textsf{B-ActiveSEAL} results under different values of $\alpha$ across diverse environments. In the long-corridor map ((a), (b)), $\alpha=0.2$ yields room-focused exploration, whereas $\alpha=3.0$ induces corridor-driven expansion. In the room-cluster map ((c), (d)), where the environment size exceeds the LiDAR sensing range, $\alpha=0.2$ leads to cautious wall-following, while $\alpha=3.0$ leverages well-known areas to push into new regions—though limited sensing ultimately results in less overall coverage than $\alpha=0.2$. In the large-scale office-like environment ((e), (f)), $\alpha=0.2$ produces cautious navigation through narrow corridors and small rooms, whereas $\alpha=3.0$ drives exploration of the large open area and expansion into unseen~corridors.}}
\label{fig:17_Appen_1}
\end{figure}



\bibliographystyle{IEEEtran}
\bibliography{bib/alias, bib/references.bib}


 






\end{document}